\documentclass[lettersize,journal]{IEEEtran}
\usepackage{amsmath,amsfonts}
\usepackage{algorithmic}
\usepackage{algorithm}
\usepackage{array}
\usepackage[caption=false,font=normalsize,labelfont=sf,textfont=sf]{subfig}
\usepackage{textcomp}
\usepackage{stfloats}
\usepackage{url}
\usepackage{verbatim}
\usepackage{graphicx}
\usepackage{cite}

\usepackage{hyperref}       % hyperlinks
\usepackage{url}            % simple URL typesetting
\usepackage{booktabs}       % professional-quality tables
\usepackage{amsfonts}       % blackboard math symbols
\usepackage{nicefrac}       % compact symbols for 1/2, etc.
\usepackage{microtype}      % microtypography
\usepackage{xcolor}         % colors
\usepackage{amssymb}
\usepackage{multirow}
\usepackage{arydshln}
\usepackage{wrapfig}
\usepackage{colortbl}
\usepackage{tabularx}
\usepackage{adjustbox}
\usepackage{tcolorbox}
\usepackage{makecell}
\usepackage{color}
\usepackage{pgfplots}
\usepackage{circledtext}
\usepackage{enumitem}
\usepackage{bm} 

\usetikzlibrary{shapes}
\usetikzlibrary{arrows,decorations.pathmorphing,backgrounds,positioning,fit,petri}
\usetikzlibrary{arrows.meta,fit,shapes.arrows}
\usetikzlibrary{positioning,shadows,patterns}

\definecolor{tiffanyblue}{RGB}{129,216,208}
\definecolor{bangdiblue}{RGB}{0,149,182}
\definecolor{kleinblue}{RGB}{0,47,167}

\definecolor{myBrown}{RGB}{140, 50, 50}
\definecolor{myPurple}{RGB}{128, 0, 200}

\usepackage{tikz}
\usepackage{subfig}
\usepackage{tkz-kiviat,pgfplots}
\pgfplotsset{compat=newest}

\usepackage{listings}

\newcommand\name{$\mathcal{X}$Transplant}

\definecolor{given}{RGB}{197,217,197}
\definecolor{response}{RGB}{176,224,230}

\newtcolorbox{promptbox}[2][]{
	width=\linewidth,
	colback = gray!8, 
	colframe = gray!8, 
	boxsep=0pt,left=5pt,right=5pt,top=2pt,bottom=2pt,
	fontupper=\linespread{1.2}\selectfont,
	title=#2,#1,
  fontupper=\small}

\begin{document}

\title{Exploring Cross-lingual Latent Transplantation: Mutual Opportunities and Open Challenges}

\author{Yangfan Ye, Xiaocheng Feng, Xiachong Feng, Libo Qin, Yichong Huang, Lei Huang, Weitao Ma, Qichen Hong, Zhirui Zhang, Yunfei Lu, Xiaohui Yan, Duyu Tang, Dandan Tu, and Bing Qin
        % <-this % stops a space
\thanks{Yangfan Ye, Xiaocheng Feng, Yichong Huang, Lei Huang, Weitao Ma, Bing Qin are with the Faculty of Computing, Harbin Institute of Technology, Harbin, China. Email: \{yfye, xcfeng, ychuang, lhuang, wtma, qinb\}@ir.hit.edu.cn. Xiachong Feng are with the Department of Computer Science, The University of Hong Kong, Hong Kong, China. Email: fengxc@hku.hk. Libo Qin are with the School of Computer Science and Engineering, Central South University, Changsha, China. Email: lbqin@csu.edu.cn. Qichen Hong, Zhirui Zhang, Yunfei Lu, Xiaohui Yan, Duyu Tang, Dandan Tu are with Huawei Technologies Co., Ltd, Beijing, China. Email: hongqichen@huawei.com, zrustc11@gmail.com, \{luyunfei6, yanxiaohui2, tangduyu, tudandan\}@huawei.com.}% <-this % stops a space
\thanks{Corresponding Author: Xiaocheng Feng.}}

% The paper headers
\markboth{Journal of \LaTeX\ Class Files,~Vol.~14, No.~8, August~2021}%
{Shell \MakeLowercase{\textit{et al.}}: A Sample Article Using IEEEtran.cls for IEEE Journals}

% \IEEEpubid{0000--0000/00\$00.00~\copyright~2021 IEEE}
% % Remember, if you use this you must call \IEEEpubidadjcol in the second
% % column for its text to clear the IEEEpubid mark.

\maketitle

%%% 0_abstact_start
\begin{abstract}
Current large language models (LLMs) often exhibit imbalances in multilingual capabilities and cultural adaptability, largely attributed to their English-centric pretraining data. 
In this paper, we introduce and investigate cross-lingual latent transplantation ({\name}), a probing framework which aims to further exploit the model's internalized multilingual knowledge during inference and examine its effects on the multilingual capability and cultural adaptability of LLMs.
{\name} framework enables models to harness the complementary strengths of both English and non-English resources by transplanting latent activations across languages.
Through extensive analysis, we empirically demonstrate that {\name}, a form of cross-lingual interaction, has mutually beneficial effects on the multilingual capability and cultural adaptability of LLMs, particularly for low-resource languages and cultures.
We further reveal that attention modules play a pivotal role in supporting multilingual understanding, while feed-forward modules are more adept at capturing culture-specific knowledge.
In addition, we conduct in-depth analysis of {\name}'s stability, effectiveness, and generalizability. By probing the upper bound performance of {\name}, we expose the considerable underutilization of current LLMs' multilingual potential—a challenge that remains open.
We hope our analysis offers a new lens for advancing cross-lingual interactions and better leveraging models' internalized multilingual knowledge.
\end{abstract}
%%% 0_abstact_end


%%%% 1_intro_start
\section{Introduction}\label{sec:intro}
\IEEEPARstart{I}{n} recent years, large language models (LLMs) have showcased their remarkable versatility across a wide range of downstream tasks~\cite{zhao2023survey, liu2023pre, dong-etal-2024-survey, wei2022emergent, wei2022chain, shanahan2024talking}, as well as their evident generalizability and adaptability in multilingual scenarios.
However, the significant imbalances in their multilingual capabilities and cultural adaptability still remain challenges that researchers are striving to resolve~\cite{ye2023language, li2024culturellm, shi2024culturebank, qin2024multilingual, ye-etal-2024-globesumm}. These issues primarily stem from their unbalanced training corpora, which is predominantly in English, leading to these models being termed \textit{English-centric} LLMs~\cite{brown2020language, zhang2022opt, touvron2023llama, biderman2023pythia}. 

Existing effective and widely adopted methods towards these challenges primarily focus on \textit{Multilingual Pretraining}, \textit{Multilingual Fine-tuning} and \textit{Language-specific Fine-tuning}. Multilingual pretraining involves initially or continuously training models on diverse multilingual corpus to develop an overall improvement of their basic multilingual capabilities~\cite{lin-etal-2022-shot, scao2022bloom, gao2024multilingual, li2024prealign}. While multilingual fine-tuning leverages supervised multilingual datasets to further enhance models' performance on downstream multilingual tasks~\cite{conneau-etal-2020-unsupervised, chen-etal-2024-monolingual, indurthi-etal-2024-improving, kew-etal-2024-turning}. And different from the above joint multilingual learning techniques, language-specific fine-tuning focuses on fine-tuning a model on a specific language or language group, often to improve performance in that particular linguistic context~\cite{reid-artetxe-2023-role, cahyawijaya-etal-2023-instructalign, ye2023language, khurana2024cross}.

However, these training-based methods have shown potential limitations. Joint multilingual pre-training and fine-tuning methods face the challenge of ``curse of multilinguality''~\cite{conneau-etal-2020-unsupervised, wang-etal-2020-negative}, a form of negative interference, where expanding too many languages during joint training can lead to a noticeable performance decline. In addition, while language-specific fine-tuning techniques are effective for individual languages, they still face generalization issues in multilingual scenarios.

In light of these challenges, this work investigates an alternative perspective: further unlocking the multilingual potential of LLMs at inference time.
% In light of these challenges, this paper aims to explore the potential for further unlocking the multilingual capabilities of LLMs at inference stage.
Recent work about ``language-specific neurons''~\cite{tang2024language, kojima2024multilingual, zhao2024large}, which refers to a subset of neurons exhibiting heightened activity for specific languages, inspired our work. These language-specific neurons significantly impacts the model's performance in corresponding languages, with minimal overlap between the English-specific and non-English-specific neurons—indicating a high degree of isolation between them~\cite{zhao2024large}. This situation leads to a dilemma for current English-centric LLMs: Given a certain question, (1) posing it in English may overlook language-specific neurons activated only by non-English inputs, potentially leading to incomplete or inaccurate responses. On the other hand, (2) posing the question in a non-English language may fail to leverage the model's strong general capabilities in English, thereby compromising overall performance.
This naturally leads to a key consideration:

\emph{Can LLMs leverage both their robust capabilities (in English) and their multilingual knowledge (in non-English languages) during inference to further unlock their multilingual potential?}

\begin{figure*}[t]
    \centering
    \includegraphics[width=0.95\textwidth]{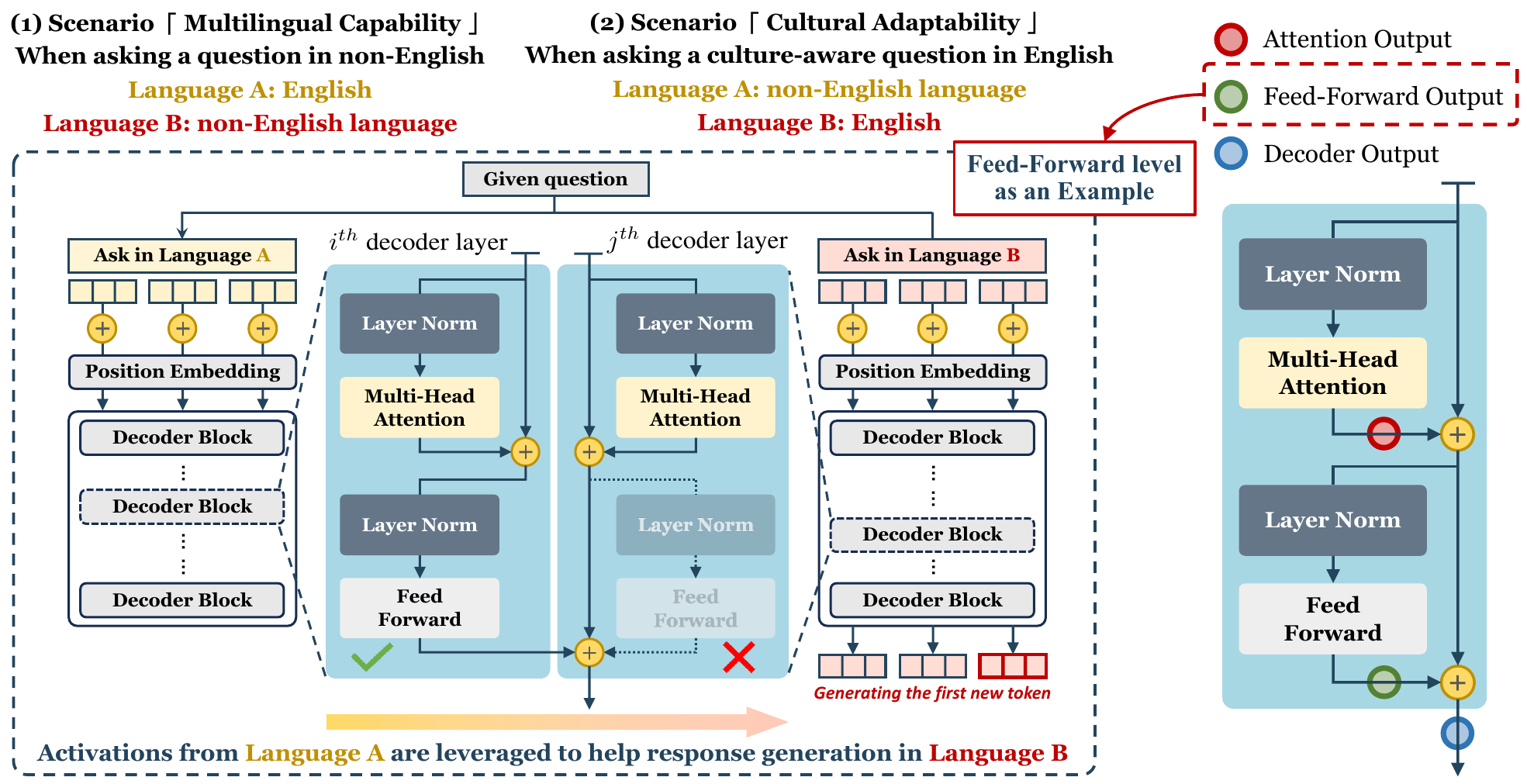}
    % \vspace{-1\baselineskip}
    \caption{Overview of the {\name} framework (feed-forward level). {\name} leverages the latent activations from the $i$-th layer of prompting in language A to replace the activations at the $j$-th layer when prompting in language B, thereby influencing the forward propagation in language B.}
    \label{fig:method}
    % \vspace{-1\baselineskip}
\end{figure*}

In response, we introduce \textbf{Cross-lingual Latent Transplantation} ({\name}), a probing framework designed to analyze how latent activations from one language influence decision making in another and aims to enable the model to benefit from latent activations derived from both English and non-English resources.
As illustrated in Figure~\ref{fig:method}, during the inference stage, {\name} extracts the activations from a specific module (e.g., feed-forward module) of the $i$-th decoder layer corresponding to one language's input and transplants them into the corresponding module of the $j$-th decoder layer for another language's input, replacing the original activations ($i$ and $j$ can be different layers). Forward propagation then continues using the transplanted activations.
Through this probe, our study delves into two distinct avenues related to two different capabilities of LLMs: 

\begin{itemize}[leftmargin=*]
\setlength{\parsep}{0pt}
\setlength{\parskip}{0pt}
\item \textbf{Multilingual Capability} reflects the model's overall performance in non-English scenarios, where we prompt the model with various non-English languages. In this context, we aim to leverage latent activations from English to enhance performance on non-English languages.
\item \textbf{Cultural Adaptability} reflects the model's ability to handle questions related to different cultural backgrounds, where we ask the model culture-related questions in English. In this context, we aim to leverage latent activations from culture-specific languages to improve performance on corresponding culture-related English questions.
\end{itemize}

Our extensive empirical study spans multiple models, tasks, languages/cultures, and levels of granularity. The key contributions of this work are as follows:

(1) We introduce {\name}, a cross-lingual latent transplantation framework that enables LLMs to benefit from both English and non-English resources, thereby unlocking greater multilingual potential, particularly for low-resource languages and cultures. We further analyze the stability, effectiveness, and generalizability of {\name}. The mutual transplantation attempts in {\name} reveal the complementary strengths of English and non-English latent activations in enhancing multilingual capability and cultural adaptability.

(2) Through empirical analysis across three levels of granularity—attention, feed-forward, and whole decoder block—we uncover structural insights: self-attention modules are pivotal for multilingual understanding, whereas feed-forward modules capture more nuanced, culture-related knowledge.

(3) Our empirical analysis show that the effectiveness of {\name} is sensitive to the choice of source and target layers during transplantation. To this end, we investigate layer selection strategies from both static and dynamic perspectives, achieving improvements for low-resource languages and cultures. In addition, by probing the upper bound performance of {\name}, we also highlight the considerable underutilization of current LLMs' multilingual potential, an important challenge that remains open.
%%%% 1_intro_end

%%%% 2_overview_start
\section{Overview}
\subsection{Framework}\label{sec:method}
For a model $M$ with $N$ decoder layers, given an input $x_s$ in source language $S$, the $x_s$ undergoes a forward propagation through all decoder layers to predict the next token. Let the input of these $N$ decoders be denoted as $H_s=\{h_s^k\}_{k=1}^{N}$ and output be denoted as $O_s=\{o_s^k\}_{k=1}^{N}$, where each $o_s^k$ is obtained by combining $h_s^k$, the attention activations $a_s^k$ and the feed-forward activations $f_s^k$ through residual connections.
Similarly, for translated version of $x_s$ in target language $T$, denoted as $x_t$, we also have $O_t=\{o_t^k\}_{k=1}^{N}$ with corresponding $H_s=\{h_t^k\}_{k=1}^{N}$, $\{f_t^k\}_{k=1}^{N}$ and $\{a_t^k\}_{k=1}^{N}$. With vanilla forward propagation process,they would predict the first new token $\hat{y}_{s}$ and $\hat{y}_{t}$ with the unembed matrix $W_{unembed}$ as follows:
\begin{equation}
\hat{y}_{s} = \text{softmax}(W_{unembed} \cdot ((h_s^N+a_s^N)+f_s^N))
\end{equation}
\begin{equation}
\hat{y}_{t} = \text{softmax}(W_{unembed} \cdot ((h_t^N+a_t^N)+f_t^N))
\end{equation}
{\name} framework refines the process by transplanting the latent activations from the $i^{th}$ decoder layer with input $x_s$ to the $j^{th}$ decoder layer with input $x_t$. Taking the transplantation of \textbf{feed-forward activations as an example} (Figure~\ref{fig:method}), formally, $f_t^j$ is replaced with $f_s^i$ and the forward propagation of $x_t$ then continues with this modification. Consequently, the original $\{o_t^k\}_{k=j}^{N}$ will be altered into $\{\tilde{o}_t^k\}_{k=j}^{N}$ due to the update in $f_t^j$, leading to new prediction outcomes $\hat{y}_{t}^{\text{(modified)}}$ as follows:
\begin{equation}
\hat{y}_{t}^{\text{(modified)}}  = \text{softmax}(W_{unembed} \cdot \tilde{o}_t^N)
\end{equation}
Notably, {\name} performs transplantation at the position of the last input token to introduce the intervention. Following tokens are then generated autoregressively without additional transplantation.

\subsection{Granularity of Latent Transplantation}
Modern transformer-based LLMs do not have a priori guarantees about how different functional components within models partition multilingual capacity and cultural knowledge. 
Given this uncertainty, in this paper, we explore the effects of activation transplantation at varying granularities. 
Specifically, we investigate latent transplantation at the level of the \textbf{attention module}, \textbf{feed-forward module}, and the \textbf{whole decoder layer}.
This analysis allows us to assess the distinct contribution of each component to cross-lingual latent interactions and their impact on multilingual and culturally adaptive behavior.

\subsection{Mutual Transplantation}
\S\ref{sec:method} details how {\name} facilitates the transfer from language $S$ to language $T$.
Notably, {\name} naturally supports transplantation in two directions, aligning with the two scenarios of \textbf{Multilingual Capability} and \textbf{Cultural Adaptability} discussed in \S\ref{sec:intro}. 
Specifically, our analysis explore the mutual attempt of {\name}: $\texttt{En} \rightarrow \texttt{non-En}$ to enhance multilingual performance, and $\texttt{non-En} \rightarrow \texttt{En}$ to enrich cultural understanding in English prompting context.

\subsection{$N^2$ Configurations}
For a model $M$ with $N$ decoder layers, both the source layer and target layer selections in {\name} offer $N$ possible choices, resulting in $N \times N = N^2$ potential transplantation combinations. For the evaluation datasets involved in our investigation, we conducted {\name} for every single sample across all $N^2$ configurations. Though $N^2$ enumeration is time-consuming, our goal is to comprehensively explore all possibilities to reveal the underlying mechanisms and assess the full potential of {\name} framework.

%%%% 2_overview_end

%%%% 3_setup_start
\section{Analysis Setup}\label{sec:analysis}

\subsection{Models.}
We selected 4 representative LLMs in this study:
(1) \textit{LLaMA-2-7B-Chat}~\cite{touvron2023llama} ($N=32$), (2) \textit{Mistral-7B-Instruct-v0.3}~\cite{jiang2023mistral7b} ($N=32$), and (3) \textit{Qwen2-7B-Instruct}~\cite{yang2024qwen2technicalreport} ($N=28$), three English-centric models for main analysis. Besides, we also utilize (4) \textit{Chinese-Alpaca-2-7B}~\cite{cui2023efficient} ($N=32$), a \textit{LLaMA-2-7B} based Chinese-centric model, for targeted further analysis in Section~\ref{sec:chinese}.  ($N$: the number of decoder layers)

\subsection{Datasets and Scope.}
% We conduct our analysis on data samples from 3 benchmarks, which can be categorized into:
We conduct experiments on data samples from 3 benchmarks, covering three representative tasks that reflect the multilingual capability and cultural adaptability of the  evaluated model, which can be categorized into: (Besides, our experiments in Section~\ref{sec:attempts} include additional 3 out-of-distribution benchmarks as introduced in Table~\ref{tab:attempts_setting}.)

\begin{itemize}[leftmargin=*]
\setlength{\parsep}{0pt}
\setlength{\parskip}{0pt}
\item \textbf{Multilingual Capability (understanding \& generation):} (1) \textit{XNLI}~\cite{conneau2018xnli}, a natural language inference dataset, (2) \textit{XQuAD}~\cite{artetxe-etal-2020-cross}, a question answering dataset. These datasets consist of linguistically parallel questions to assess the model's ability across languages. For questions in these non-English languages, we apply $\texttt{En} \rightarrow \texttt{non-En}$ {\name}.
\item \textbf{Cultural Adaptability:} \textit{GlobalOpinionQA}~\cite{durmus2023towards} contains QAs from cross-national surveys designed to capture diverse opinions on global issues across different countries, and all the questions are in English. For these questions in English, we apply $\texttt{non-En} \rightarrow \texttt{En}$ {\name}.
% , hoping the model to leverage latent activations from non-English languages to better capture cultural nuances.
\end{itemize}

Due to the extensive scale of our experiments\footnote{We perform inference across all $N^2$ configurations for every single instance. Such as for model \textit{LLaMA-2-7B-Chat} with layer number $N=32$, $N^2=1024$ times inference are conducted for each instance.}, we did not use the full version of each dataset. Instead, we constructed \textit{PilotSets} through linguistically and culturally balanced sampling, covering questions from 16 languages and 20 cultural backgrounds.
Notably, for the \textit{GlobalOpinionQA} dataset, we retain only samples where the maximum option probability exceeds 0.8 to ensure answer reliability.
For better reproducibility, these \textit{PilotSets} are publicly available along with our code. The detailed information of these \textit{PilotSets} is as follows:
\begin{promptbox}

\textbf{Involved Languages / Cultures}
    \vspace{0.5\baselineskip} \\
    XNLI (14): ar, bg, de, el, en, es, fr, hi, ru, sw, tr, ur, vi, zh \\
    XQuAD (12): ar, de, el, en, es, hi, ro, ru, th, tr, vi, zh \\
    GlobalOpinionQA (20): am, ar, de, el, en, es, fr, hi, id, ja, pt, ru, sv, sw, tl, tr, uk, ur, vi, zh-CN
\end{promptbox}

\begin{promptbox}

\textbf{Data Size}
    \vspace{0.5\baselineskip} \\
    \text{XNLI: $700$}, \text{XQuAD: $600$}, \text{GlobalOpinionQA: $1000$}
\end{promptbox}

\subsection{Evaluations.}
The prompts we used for each \textit{PilotSet} are listed in Supplementary Materials~A Table~X. For each model involved, we apply greedy decoding strategy and set the max new tokens generated by the model to 20. This token limit was determined based on empirical observations, ensuring that it sufficiently covers the typical response length across involved datasets. We used Accuracy as our evaluation metric, and for different task types, we applied the following rules:

\begin{itemize}[leftmargin=*]
\setlength{\parsep}{0pt}
\setlength{\parskip}{0pt}
\item \textbf{For Multiple-choice Tasks (Classification):} \textit{XNLI} and \textit{GlobalOpinionQA} all belong to the multiple-choice category. For these tasks, a model's response is considered correct only if it contains the correct option and excludes all other options.
\item \textbf{For Question-Answering Tasks (Generation):} For the generative task \textit{XQuAD}, the model's answer is deemed correct if the gold answer appears in the model's response.
\end{itemize}

%%%% 3_setup_end

%%%% 4_granularity_start
\section{Transplantation Across Granularities}

\begin{table}
  \centering 
  \caption{The averaged perplexity and input-output language consistency results across all $N^2$ configurations of {\name} on LLaMA-2-7B-Chat, compared with the performance of vanilla model. $\texttt{non-En}$ and $\texttt{En}$ represent the input-output language required by corresponding tasks.}
  % \vspace{0.5\baselineskip}
  \resizebox{0.48\textwidth}{!}{%
      \begin{tabular}{lccc}
          \toprule
          \multirow{1}{*}{\bf {Perplexity}} & \textbf{XNLI} & \textbf{XQuAD}  & \textbf{GlobalOpinionQA} \\
          \midrule
          Vanilla (w/o {\name}) & 64.35 & 127.46 & 54.45 \\
          \noalign{\vskip 0.2ex}\cdashline{1-4}\noalign{\vskip 0.4ex}
          {\name} (Self-Attention) & 56.50 & 143.10 & 39.76 \\
          {\name} (Feed-Forward) & 81.48 & 135.95 & 40.57 \\
          {\name} (Whole Block) & 89.80 & 140.30 & 43.01 \\
          % \bottomrule
          \toprule
          \multirow{1}{*}{\bf {Language}} & \textbf{XNLI} & \textbf{XQuAD} & \textbf{GlobalOpinionQA} \\
          \multirow{1}{*}{\bf Consistency} (\%) & ($\texttt{non-En}$) & ($\texttt{non-En}$) & ($\texttt{En}$) \\
          \midrule
          Vanilla (w/o {\name}) & {95.20} & {83.00} & {99.83} \\
          \noalign{\vskip 0.2ex}\cdashline{1-4}\noalign{\vskip 0.4ex}
          {\name} (Self-Attention) & 92.50 & 77.23 & 99.17 \\
          {\name} (Feed-Forward) & 94.12 & 89.03 & 99.74 \\
          {\name} (Whole Block) & 89.67 & 72.12 & 98.69 \\
          \bottomrule
      \end{tabular}}
      \label{tab:modeling}
\end{table}

In this section, we explore latent transplantation at different granularities of \textit{Attn-level}, \textit{FFN-level}, and \textit{Block-level}, to understand their potential individual contributions and effectiveness.

\subsection{Impact on Language Modeling Ability}

Directly modifying activations during inference is a delicate operation and may disrupt the model's outputs. To ensure the reliability of {\name}, we first investigate two aspects before applying it to downstream tasks. (1) \textbf{Response Perplexity} serves as an indicator of whether {\name} distorts the model's fundamental language modeling ability. (2) \textbf{Language Consistency} analysis evaluates whether {\name} introduces unintended interference where models fail to response in the appropriate language given the context. 

The perplexity results in Table~\ref{tab:modeling} indicate that, on average, {\name} framework does not noticeably impair the model's fundamental language modeling ability. But whole \textit{Block-level} {\name} may pose slightly higher potential risks compared to \textit{Attn-level} and \textit{FFN-level}.

The language consistency results\footnote{Languages are identified using the \textit{lid.176.bin} model from \textit{fastText}, which recognizes 176 languages.} across all $N^2$ transplantation settings for both \textit{Attn-level} and \textit{FFN-level} {\name} aligns well with that of the vanilla setting (except for \textit{Attn}-level on \textit{XQuAD}). This indicates that \textit{Attn-level} and \textit{FFN-level} {\name} rarely affect the language consistency between input and output language. 
In contrast, \textit{Block-level} {\name} consistently yields noticeably lower language consistency across all datasets, suggesting a higher risk of introducing language inconsistency.

Together, above analysis verify that {\name} preserves the model's core fluency while avoiding language inconsistency, providing a stable foundation for subsequent task evaluations.

\textbf{Finding-1:} \textit{Transplanting latent activations at the Attn- or FFN-level appears to be a relatively safe strategy, and is worth further exploration. In contrast, block-level transplantation tends to introduce greater instability, which can impair the model's fundamental language modeling capabilities.}

(\textbf{Note:} Given Finding-1, we focus subsequent analysis on {\name} at the \textit{Attn-} and \textit{FFN-level}.)

\definecolor{my-red}{RGB}{239, 150, 190}
\definecolor{my-yellow}{RGB}{149, 224, 196}
\definecolor{my-green}{RGB}{154,154,222}

\begin{figure}[t]
    \pgfplotsset{width=0.5\linewidth, height=1\linewidth, compat=1.15}
    \centering
    \begin{tikzpicture}[]
        \node[font=\footnotesize] at (3.3,2.3) {(1) \textit{LLaMA-2-7B-Chat}};
        % XTransplant-attention
        \begin{axis}[
            clip=true, % 让边框浮于柱子之上
            at={(0em, 0em)},
            xbar stacked,
            bar width=10pt,
            width=4.7cm,
            height=3.2cm,
            xmin=0, xmax=100,
            xlabel=\scriptsize,
            ytick=data,
            yticklabels={
                XNLI,
                XQuAD,
                Global\\OpinionQA
            },
            y dir=reverse,
            yticklabel style={align=center, font=\footnotesize, text width=1.5cm, xshift=-1pt}, % 设置 text width 和居中
            title={{\name} (\textit{Attn-level})},
            title style={yshift=-0.5em,font=\footnotesize},
            tick label style={font=\scriptsize},
            enlarge y limits=0.3, % 通过扩大x轴边界来增加条形之间的间隔
        ]
        
        \addplot+[xbar, fill=my-red, draw=none] coordinates {(95.1,0) (21.8,1) (6.9,2)};
        \addplot+[xbar, fill=my-yellow, draw=none]  coordinates {(1.6,0) (2.4,1) (0.4,2)};
        \addplot+[xbar, fill=my-green, draw=none]  coordinates {(3.4,0)  (75.8,1) (92.7,2)};

        \end{axis}
        
        % XTransplant-feed forward
        \begin{axis}[
            at={(3.7cm,0cm)},
            xbar stacked,
            bar width=10pt,
            width=4.7cm,
            height=3.2cm,
            xmin=0, xmax=100,
            xlabel=\scriptsize,
            yticklabels=\empty,
            ytick=data,
            y dir=reverse,
            title={{\name} (\textit{FFN-level})},
            title style={yshift=-0.5em,font=\footnotesize},
            tick label style={font=\scriptsize},
            enlarge y limits=0.3, % 通过扩大x轴边界来增加条形之间的间隔
        ]
        \addplot+[xbar, fill=my-red, draw=none] coordinates {(77.8,0) (22.5,1) (14.5,2)};
        \addplot+[xbar, fill=my-yellow, draw=none]  coordinates {(3.1,0) (5.7,1) (0.9,2)};
        \addplot+[xbar, fill=my-green, draw=none]  coordinates {(19.0,0) (71.8,1) (84.6,2)};

        \end{axis}

        \node[font=\footnotesize] at (3.3,-0.7) {(2) \textit{Mistral-7B-Instruct-v0.3}};
        % XTransplant-attention
        \begin{axis}[
            clip=true, % 让边框浮于柱子之上
            at={(0em, -8.5em)},
            xbar stacked,
            bar width=10pt,
            width=4.7cm,
            height=3.2cm,
            xmin=0, xmax=100,
            xlabel=\scriptsize,
            ytick=data,
            yticklabels={
                XNLI,
                XQuAD,
                Global\\OpinionQA
            },
            y dir=reverse,
            yticklabel style={align=center, font=\footnotesize, text width=1.5cm, xshift=-1pt}, % 设置 text width 和居中
            title={{\name} (\textit{Attn-level})},
            title style={yshift=-0.5em,font=\footnotesize},
            tick label style={font=\scriptsize},
            enlarge y limits=0.3, % 通过扩大x轴边界来增加条形之间的间隔
        ]
        
        \addplot+[xbar, fill=my-red, draw=none] coordinates {(85.0,0) (15.4,1) (12.9,2)};
        \addplot+[xbar, fill=my-yellow, draw=none]  coordinates {(2.8,0) (6.7,1) (4.2,2)};
        \addplot+[xbar, fill=my-green, draw=none]  coordinates {(12.2,0) (77.9,1) (82.9,2)};

        \end{axis}
        
        % XTransplant-feed forward
        \begin{axis}[
            % axis line style={thick, black}, % 设置边框加粗
            at={(3.7cm,-8.5em)},
            xbar stacked,
            bar width=10pt,
            width=4.7cm,
            height=3.2cm,
            xmin=0, xmax=100,
            xlabel=\scriptsize,
            yticklabels=\empty,
            ytick=data,
            y dir=reverse,
            title={{\name} (\textit{FFN-level})},
            title style={yshift=-0.5em,font=\footnotesize},
            tick label style={font=\scriptsize},
            enlarge y limits=0.3, % 通过扩大x轴边界来增加条形之间的间隔
            % enlarge x limits=0.003, % 增加 X 轴范围，避免柱子被挡住
        ]
        \addplot+[xbar, fill=my-red, draw=none] coordinates {(70.1,0) (40.4,1) (27.6,2)};
        \addplot+[xbar, fill=my-yellow, draw=none]  coordinates {(4.6,0) (7.0,1) (5.3,2)};
        \addplot+[xbar, fill=my-green, draw=none]  coordinates {(25.3,0) (52.6,1) (67.1,2)};

        \end{axis}

        \node[font=\footnotesize] at (3.3,-3.7) {(3) \textit{Qwen2-7B-Instruct}};
        % XTransplant-attention
        \begin{axis}[
            % axis line style={thick, black}, % 设置边框加粗
            clip=true, % 让边框浮于柱子之上
            at={(0em,-17em)},
            xbar stacked,
            bar width=10pt,
            width=4.7cm,
            height=3.2cm,
            xmin=0, xmax=100,
            xlabel=\scriptsize,
            ytick=data,
            yticklabels={
                XNLI,
                % XCOPA,
                XQuAD,
                Global\\OpinionQA
            },
            y dir=reverse,
            yticklabel style={align=center, font=\footnotesize, text width=1.5cm, xshift=-1pt}, % 设置 text width 和居中
            title={{\name} (\textit{Attn-level})},
            title style={yshift=-0.5em,font=\footnotesize},
            tick label style={font=\scriptsize},
            enlarge y limits=0.3, % 通过扩大x轴边界来增加条形之间的间隔
            % enlarge x limits=0.005, % 增加 X 轴范围，避免柱子被挡住
        ]
        
        \addplot+[xbar, fill=my-red, draw=none] coordinates {(48.2,0) (24.0,1) (27.7,2)};
        \addplot+[xbar, fill=my-yellow, draw=none]  coordinates {(7.7,0) (2.0,1) (5.2,2)};
        \addplot+[xbar, fill=my-green, draw=none]  coordinates {(44.1,0) (74.0,1) (67.1,2)};

        \end{axis}
        
        % XTransplant-feed forward
        \begin{axis}[
            % axis line style={thick, black}, % 设置边框加粗
            at={(3.7cm,-17em)},
            xbar stacked,
            bar width=10pt,
            width=4.7cm,
            height=3.2cm,
            xmin=0, xmax=100,
            xlabel=\scriptsize,
            yticklabels=\empty,
            ytick=data,
            y dir=reverse,
            title={{\name} (\textit{FFN-level})},
            title style={yshift=-0.5em,font=\footnotesize},
            tick label style={font=\scriptsize},
            enlarge y limits=0.3, % 通过扩大x轴边界来增加条形之间的间隔
            % enlarge x limits=0.003, % 增加 X 轴范围，避免柱子被挡住
        ]
        \addplot+[xbar, fill=my-red, draw=none] coordinates {(31.1,0) (15.7,1) (27.0,2)};
        \addplot+[xbar, fill=my-yellow, draw=none]  coordinates {(8.7,0) (2.5,1) (12.9,2)};
        \addplot+[xbar, fill=my-green, draw=none]  coordinates {(60.2,0) (81.8,1) (60.1,2)};

        \end{axis}

        % 图例（共用）
        \begin{scope}[shift={(2.3,2.9)}]
        \path (0,-0.17) node[rectangle, draw=none] (leg) {
            \begin{axis}[
                axis line style={thick, black}, % 设置边框加粗
                hide axis,
                xmin=0, xmax=1,
                ymin=0, ymax=1,
                legend style={
                    legend columns=3,
                    column sep=1ex,
                    at={(0.5,1.0)}, anchor=south,
                    font=\scriptsize
                }
            ]
            \addlegendimage{area legend, fill=my-red, draw=none}
            \addlegendentry{Win Rate}
            \addlegendimage{area legend, fill=my-yellow, draw=none}
            \addlegendentry{Tie Rate}
            \addlegendimage{area legend, fill=my-green, draw=none}
            \addlegendentry{Lose Rate}
            \end{axis}
        };
        \end{scope}
    \end{tikzpicture}
    \caption{Overall ``Win, Tie, Lose'' rates across all $N^2$ configurations at \textsc{Attn-} and \textsc{FFN-Level} against the vanilla performance. 
    (a)	\textbf{Win Rate}: the percentage of $N^2$ evaluation cases in which the averaged performance across all involved languages under a given transplantation configuration is higher than the performance of corresponding backbone model.
    (b)	\textbf{Tie Rate}: the percentage of $N^2$ evaluation cases in which the averaged performance across all involved languages under a given transplantation configuration is equal to the performance of corresponding backbone model.
    (c)	\textbf{Lose Rate}: the percentage of $N^2$ evaluation cases in which the averaged performance across all involved languages under a given transplantation configuration is lower than the performance of corresponding backbone model.
    }
    \label{fig:win}
\end{figure}

\begin{table}
  \centering 
  \caption{Average Performance Of {\name} Across All $N^2$ Configurations At Attn- And FFN-Level Against Vanilla Model.}
  \resizebox{0.48\textwidth}{!}{%
      \begin{tabular}{lccc}
          \toprule
          \multirow{1}{*}{\bf {Datasets}} & Vanilla Performance & \textit{Attn-level} $N^2$ Average & \textit{FFN-level} $N^2$ Average\\
          \midrule
          \multicolumn{4}{c}{\textit{LLaMA-2-7B-Chat}} \\
          \noalign{\vskip 0.2ex}\cdashline{1-4}\noalign{\vskip 0.4ex}
          XNLI & 30.1 & 33.2 & 30.5 \\
          XQuAD & 33.5 & 31.3 & 31.9 \\
          GlobalOpinionQA & 32.1 & 25.8 & 29.0 \\
          \midrule
          \multicolumn{4}{c}{\textit{Mistral-7B-Instruct-v0.3}} \\
          \noalign{\vskip 0.2ex}\cdashline{1-4}\noalign{\vskip 0.4ex}
          XNLI & 37.7 & 40.3 & 38.0 \\
          XQuAD & 39.8 & 35.9 & 39.9 \\
          GlobalOpinionQA & 68.3 & 66.4 & 66.5 \\
          \midrule
          \multicolumn{4}{c}{\textit{Qwen2-7B-Instruct}} \\
          \noalign{\vskip 0.2ex}\cdashline{1-4}\noalign{\vskip 0.4ex}
          XNLI & 55.2 & 55.2 & 54.4 \\
          XQuAD & 47.3 & 44.2 & 45.5 \\
          GlobalOpinionQA & 64.2 & 62.5 & 62.4 \\
          \bottomrule
      \end{tabular}}
      \label{tab:avg}
\end{table}

\subsection{Effectiveness Across $N^2$ Transplantation Configurations}\label{sec:N2con}

Here we present the overall ``Win, Tie, Lose'' rates and average performance of {\name} across all $N^2$ configurations at different granularities against the vanilla performance, highlighting the extent to which {\name} yields improvements or incurs declines under varying transplantation granularity.

From the effectiveness results in Figure~\ref{fig:win} and Table~\ref{tab:avg}, we observe that on multilingual understanding task \textit{XNLI}, many of the $N^2$ configurations of \textit{Attn-level} and \textit{FFN-level} {\name} can outperform the vanilla model. Besides, \textit{Attn-level} achieves better performance than \textit{FFN-level} in terms of both win rate and average performance. However, on multilingual generation task \textit{XQuAD} and culture-aware task \textit{GlobalOpinionQA}, the advantage of {\name} becomes less pronounced, with only a limited number of configurations outperforming the vanilla model. In these two tasks, \textit{FFN-level} {\name} slightly outperforms \textit{Attn-level} {\name}.

\textbf{Finding-2:} \textit{The effectiveness varies across different granularities. Specifically, the self-attention modules within models play a crucial role in the model's multilingual understanding capability, while the feed-forward modules capture more nuanced, culture-related knowledge.}

\textbf{Finding-3:} \textit{The effectiveness of {\name} greatly depends on the selection of source and target layers during transplantation, which highlights the importance of layer-specific considerations to ensure that {\name} remains effective.}

\begin{figure*}[ht]
  \centering
  \includegraphics[width=0.95\textwidth]{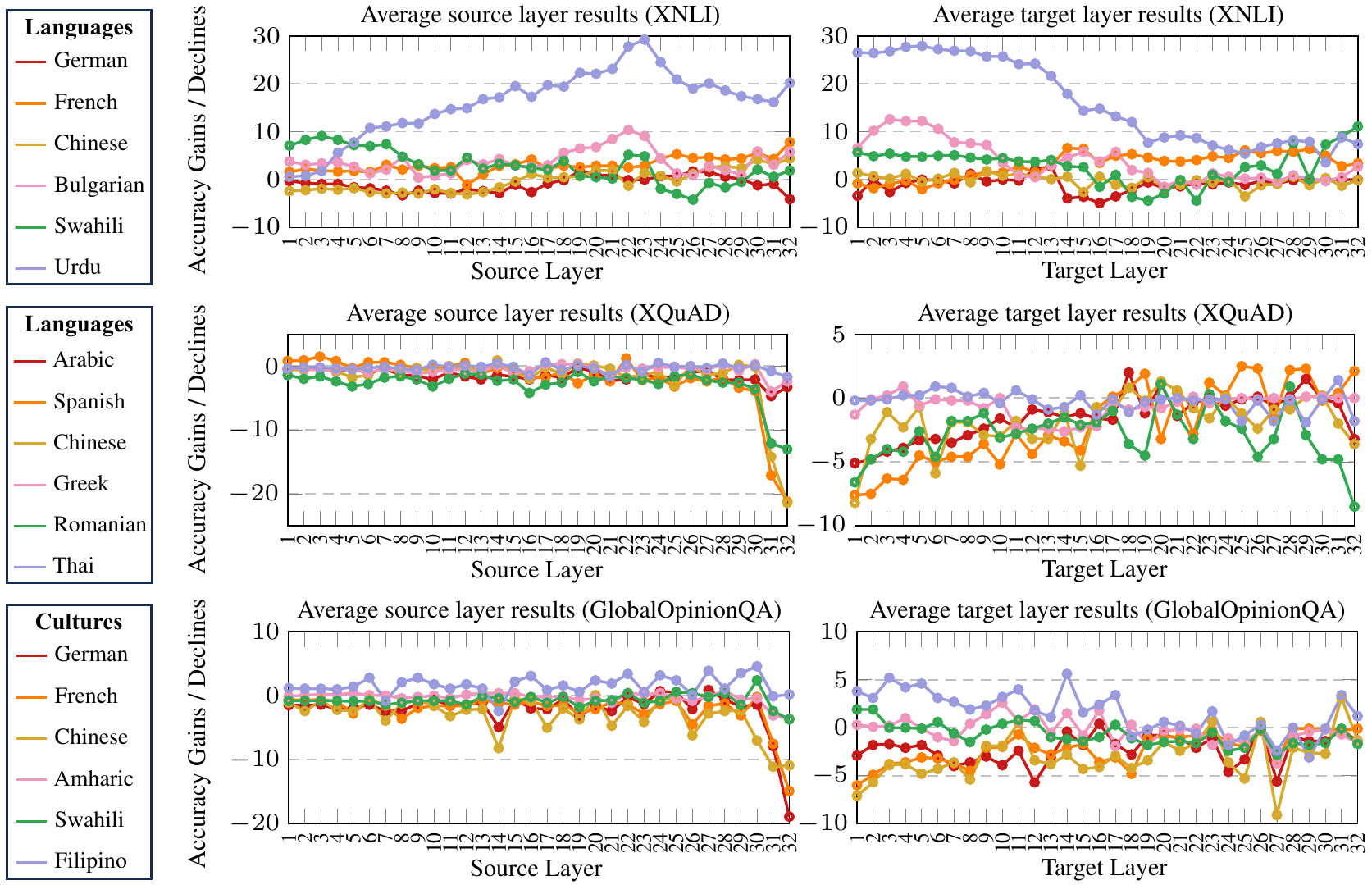}
  \caption{The average layer-wise performance gains or declines of {\name} on \textit{LLaMA-2-7B-Chat}, under different source or target layer configurations (results on more models are provided in Supplementary Materials B Figure 6,7).
  }
  \label{fig:layer_lang_llama}
\end{figure*}

\subsection{Layer-wise Effectiveness Across Languages and Cultures}\label{sec:layer}

Finding-3 highlights the importance of source and target layer selection in {\name} framework. In this section, we explore the layer-wise patterns that govern the effectiveness of {\name}.
Building on Finding-2, we apply \textit{attn-level} {\name} to \textit{XNLI} dataset, while \textit{ffn-level} {\name} is used for \textit{XQuAD} and culture-aware task \textit{GlobalOpinionQA}.
(To ensure readability, only parts of the languages / cultures are exhibited.)

From the average layer-wise performance gains or declines of {\name} in Figure~\ref{fig:layer_lang_llama}, we observe that in both multilingual and culture-aware scenarios, {\name} provides notably greater improvements for low-resource languages (e.g., Swahili, Urdu, Filipino) compared to high-resource languages (e.g., German, French, Chinese). Besides, despite some differences in detail, the overall trend of layer-wise effectiveness of {\name} exhibits a similar pattern across the 3 LLMs involved (Figure~\ref{fig:layer_lang_llama},~7,~8), suggesting the potential existence of universal underlying mechanisms in LLMs for handling multilingual and culture-aware inputs. (Figure~7,~8 are placed in Supplementary Materials B.)

\textbf{Finding-4:} \textit{{\name} demonstrates greater effectiveness in low-resource language scenarios and in addressing culture-specific questions related to low-resource languages.}

\textbf{Finding-5:} \textit{The layer-wise effectiveness shows a similar pattern across involved LLMs, demonstrating great generalization across models.}

%%% 4_granularity_end

(\textbf{Note:} The following experiments in Section~\ref{sec:attempts} and Section~\ref{sec:discuss} do not require $N^2$ inferences, the extensive $N^2$-inference analysis in Section~\ref{sec:analysis} is conducted to provide a comprehensive overview of the feasibility of {\name}.)

%%% 5_static_start

\section{Attempts With Coarse-grained Layer Selection}

\begin{table*}[ht]
  \small
    \centering
    \caption{The Settings And The Evaluation Datasets Used In Our Coarse-grained Attempts.}
    \resizebox{0.85\linewidth}{!}{
      \begin{tabular}{ccccc}
      \toprule
      \multirow{2.5}{*}{\bf {Task Type}} & \multirow{2.5}{*}{\bf {{\name} Granularity}} & \multirow{2.5}{*}{\textbf{Layer Selection ( Source, Target )}} & \multicolumn{2}{c}{\bf {Unseen Evaluation Datasets}} \\
      \cmidrule(lr){4-5}
       & & & \textbf{ID Data} & \textbf{OOD Data} \\
      \midrule
      Multilingual Understanding & Attn-level & ( Last layer, First layer ) & XNLI & XCOPA \\
      Multilingual Generation & FFN-level & ( First layer, First layer ) & XQuAD & MKQA\\
      Cultural Adaptability & FFN-level & ( First layer, First layer ) & GlobalOpinionQA & CulturalBench \\
      \bottomrule
      \end{tabular}
    }
    \label{tab:attempts_setting}
  \end{table*}

  \begin{table*}[ht]
    \centering
    \scriptsize
    \renewcommand{\arraystretch}{0.8} % 设置行间距为默认的 1.5 倍
    \setlength{\tabcolsep}{1pt}
    \setlength{\dashlinedash}{3pt} % 设置虚线段长度
    \setlength{\dashlinegap}{1pt}  % 设置虚线段之间的间隔
    \caption{Performance Comparisons Between Vanilla Model And Aftering Applying Coarse-grained {\name} On \textit{LLaMA-2-7B-Chat} (Results On More Models Are In Supplementary Materials C-B Figure~XII, XIII). \textbf{High}, \textbf{Mid}, And \textbf{Low} Refer To High-, Medium-, And Low-resource Languages Or Cultures\protect\footnotemark.}
    \begin{tabularx}{\textwidth}{>{\centering\arraybackslash}p{1.2cm}*{7}{>{\centering\arraybackslash}X}*{5}{>{\centering\arraybackslash}X}*{5}{>{\centering\arraybackslash}X}}
    \toprule
    \textbf{Datasets} & \multicolumn{17}{c}{\bf Unseen In-Distribution Data} \\
    % \cmidrule(lr){1-19}
    \midrule
    \multirow{3}{*}{\bf \shortstack{XNLI}} & \multicolumn{7}{c}{\bf High} & \multicolumn{5}{c}{\bf Mid} & \multicolumn{5}{c}{\bf Low} \\
    \cmidrule(lr){2-8}
    \cmidrule(lr){9-13}
    \cmidrule(lr){14-18}
     & ar & de & en & es & fr & zh & \textbf{Avg.} & hi & ru & tr & vi & \textbf{Avg.} & bg & el & sw & ur & \textbf{Avg.} \\
    \midrule
    Vanilla & 34.8 & 35.3 & 47.3 & 35.6 & 39.7 & 33.3 & 37.7 & 8.4 & 34.4 & 34.7 & 32.8 & 27.6 & 32.6 & 34.1 & 18.6 & 0.5 & 21.5 \\
    \noalign{\vskip 0.2ex}
    Coarse & 35.4 & 34.2 & 48.6 & 35.8 & 37.6 & 37.0 & 38.1 & 26.6 & 32.3 & 31.5 & 31.8 & 30.6 & 34.5 & 33.7 & 32.1 & 29.9 & 32.6 \\
    \end{tabularx}
    
    \begin{tabularx}{\textwidth}{>{\centering\arraybackslash}p{1.2cm}*{7}{>{\centering\arraybackslash}X}*{4}{>{\centering\arraybackslash}X}*{4}{>{\centering\arraybackslash}X}}
    \toprule
    \multirow{3}{*}{\bf \shortstack{XQuAD}} & \multicolumn{7}{c}{\bf High} & \multicolumn{4}{c}{\bf Mid} & \multicolumn{4}{c}{\bf Low} \\
    \cmidrule(lr){2-7}
    \cmidrule(lr){8-12}
    \cmidrule(lr){13-16}
     & ar & de & en & es & zh & \textbf{Avg.} & hi & ru & tr & vi & \textbf{Avg.} & el & ro & th & \textbf{Avg.} \\
    \cmidrule(lr){1-7}
    \cmidrule(lr){8-12}
    \cmidrule(lr){13-16}
    Vanilla & 7.4 & 51.3 & 70.8 & 56.1 & 46.4 & 46.4 & 8.3 & 39.7 & 27.0 & 31.4 & 26.6 & 6.0 & 48.2 & 4.3 & 19.5 \\
    \noalign{\vskip 0.2ex}
    Coarse & 7.1 & 51.2 & 70.8 & 56.5 & 46.3 & 46.4 & 8.7 & 39.3 & 27.1 & 32.6 & 26.9 & 6.1 & 54.0 & 4.4 & 21.5 \\
    \end{tabularx}

    \begin{tabularx}{\textwidth}{>{\centering\arraybackslash}p{1.2cm}*{8}{>{\centering\arraybackslash}X}*{7}{>{\centering\arraybackslash}X}*{8}{>{\centering\arraybackslash}X}}
    \toprule
    \multirow{3}{*}{\bf \shortstack{Global\\OpinionQA}} & \multicolumn{8}{c}{\bf High} & \multicolumn{7}{c}{\bf Mid} & \multicolumn{8}{c}{\bf Low} \\
    \cmidrule(lr){2-9}
    \cmidrule(lr){10-16}
    \cmidrule(lr){17-24}
     & ar & de & en & es & fr & ja & zh & \textbf{Avg.} & hi & pt & ru & sv & tr & vi & \textbf{Avg.} & am & el & id & sw & tl & uk & ur & \textbf{Avg.} \\
    \cmidrule(lr){1-9}
    \cmidrule(lr){10-16}
    \cmidrule(lr){17-24}
    Vanilla & 32.7 & 41.0 & 29.6 & 38.6 & 32.4 & 26.8 & 30.3 & 33.1 & 25.0 & 38.2 & 39.5 & 53.8 & 32.8 & 14.3 & 33.9 & 21.1 & 30.0 & 30.3 & 28.6 & 22.2 & 33.3 & 30.5 & 28.0 \\
    \noalign{\vskip 0.2ex}
    Coarse & 26.3 & 40.0 & 33.3 & 38.9 & 31.9 & 26.8 & 28.8 & 32.3 & 33.3 & 40.4 & 39.5 & 53.8 & 32.8 & 14.3 & 35.7 & 29.6 & 44.3 & 31.8 & 25.7 & 33.3 & 33.3 & 32.2 & 32.9 \\
    \end{tabularx}
    
    \begin{tabularx}{\textwidth}{>{\centering\arraybackslash}p{1.2cm}*{7}{>{\centering\arraybackslash}X}*{5}{>{\centering\arraybackslash}X}*{6}{>{\centering\arraybackslash}X}}
    \toprule
    {\bf {Datasets}} & \multicolumn{12}{c}{\bf Unseen Out-of-distribution Data} \\
    \midrule
    \multirow{3}{*}{\bf \shortstack{XCOPA}} & \multicolumn{3}{c}{\bf High} & \multicolumn{4}{c}{\bf Mid} & \multicolumn{5}{c}{\bf Low} \\
    \cmidrule(lr){2-4}
    \cmidrule(lr){5-8}
    \cmidrule(lr){9-13}
     & en & zh & \textbf{Avg.} & it & tr & vi & \textbf{Avg.} & ht & id & sw & ta & \textbf{Avg.} \\
    \cmidrule(lr){1-4}
    \cmidrule(lr){5-8}
    \cmidrule(lr){9-13}
    Vanilla & 53.6 & 56.4 & 55.0 & 29.6 & 47.6 & 50.2 & 42.4 & 12.2 & 53.1 & 0.0 & 0.0 & 16.3 \\
    \noalign{\vskip 0.2ex}
    Coarse & 56.7 & 49.6 & 53.1 & 49.1 & 53.3 & 46.4 & 49.6 & 15.6 & 49.3 & 43.6 & 48.4 & 39.2 \\
    \end{tabularx}
    
    \begin{tabularx}{\textwidth}{>{\centering\arraybackslash}p{1.2cm}*{8}{>{\centering\arraybackslash}X}*{6}{>{\centering\arraybackslash}X}*{6}{>{\centering\arraybackslash}X}}
    \toprule
    \multirow{3}{*}{\bf \shortstack{MKQA}} & \multicolumn{8}{c}{\bf High} & \multicolumn{6}{c}{\bf Mid} & \multicolumn{6}{c}{\bf Low} \\
    \cmidrule(lr){2-9}
    \cmidrule(lr){10-15}
    \cmidrule(lr){16-21}
     & ar & de & en & es & fr & ja & zh & \textbf{Avg.} & it & ko & pt & tr & vi & \textbf{Avg.} & fi & he & hu & nl & no & \textbf{Avg.} \\
    \cmidrule(lr){1-9}
    \cmidrule(lr){10-15}
    \cmidrule(lr){16-21}
    Vanilla & 0.9 & 9.9 & 19.7 & 9.3 & 7.0 & 4.0 & 4.0 & 7.8 & 7.9 & 3.3 & 10.7 & 9.2 & 12.4 & 8.7 & 7.0 & 0.8 & 11.7 & 12.7 & 15.7 & 9.6 \\
    \noalign{\vskip 0.2ex}
    Coarse & 0.9 & 9.8 & 19.7 & 11.4 & 7.6 & 4.7 & 4.2 & 8.3 & 8.2 & 3.1 & 10.9 & 9.7 & 12.1 & 8.8 & 7.6 & 0.9 & 11.8 & 12.8 & 16.0 & 9.8 \\
    \end{tabularx}
    
    \begin{tabularx}{\textwidth}{>{\centering\arraybackslash}p{1.2cm}*{8}{>{\centering\arraybackslash}X}*{6}{>{\centering\arraybackslash}X}*{5}{>{\centering\arraybackslash}X}}
    \toprule
    \multirow{3}{*}{\bf \shortstack{Cultural\\Bench}} & \multicolumn{8}{c}{\bf High} & \multicolumn{6}{c}{\bf Mid} & \multicolumn{5}{c}{\bf Low} \\
    \cmidrule(lr){2-9}
    \cmidrule(lr){10-15}
    \cmidrule(lr){16-20}
     & ar & de & en & es & fr & ja & zh & \textbf{Avg.} & hi & ko & ru & tr & vi & \textbf{Avg.} & he & ms & pl & tl & \textbf{Avg.} \\
    % \cmidrule(lr){1-1}
    \cmidrule(lr){1-9}
    \cmidrule(lr){10-15}
    \cmidrule(lr){16-20}
    Vanilla & 52.9 & 46.9 & 84.4 & 67.5 & 64.3 & 67.9 & 54.2 & 62.6 & 54.3 & 34.1 & 60.0 & 63.2 & 63.0 & 54.9 & 46.2 & 45.5 & 58.3 & 66.7 & 54.2 \\
    \noalign{\vskip 0.2ex}
    Coarse & 52.9 & 46.9 & 84.4 & 67.5 & 64.3 & 64.2 & 52.5 & 61.8 & 54.3 & 34.1 & 60.0 & 63.2 & 63.0 & 54.9 & 53.8 & 45.5 & 62.5 & 66.7 & 57.1 \\
    \toprule
    \end{tabularx}
    
    \label{tab:attempts_results_llama}
  \end{table*}
  
\subsection{Coarse-grained {\name}}\label{sec:attempts}

Sections~\ref{sec:N2con} and~\ref{sec:layer} reveal the complementary benefits of English and non-English resources in enhancing (low-resource) multilingual capability and cultural adaptability under {\name}, but its effectiveness specifically depends on the choice of source and target layers during transplantation.

Based on these empirical results across all $N^2$ configurations from Section~\ref{sec:analysis}, we notice that a coarse-grained static layer strategy yields improvements for low-resource languages/cultures where we apply \textit{Attn-level Last-to-First} {\name} for understanding tasks and \textit{FFN-level First-to-First} {\name} for generation and culture-related tasks, as summarized in Table~\ref{tab:attempts_setting} (detailed layer-specific results under all $N^2$ configurations are in Supplementary Materials~C-A). 

To verify the generalizability of {\name} and the layer selection strategy, we conduct evaluation on two types of data: (1) \textbf{unseen in-distribution (ID) data}, referring to the data points from the same datasets as the \textit{PilotSets} but not included in the \textit{PilotSets}, and (2) \textbf{unseen out-of-distribution (OOD) data}. Settings are in Table~\ref{tab:attempts_setting} and results are in Table~\ref{tab:attempts_results_llama},~XIII,~XIV. (Table~XIII,~XIV are placed in Supplementary Materials~C-B.)

\footnotetext{The categorization follows the taxonomy in \url{https://microsoft.github.io/linguisticdiversity/assets/lang2tax.txt}, where languages rated as 5 are classified as high-resource, those rated as 4 as medium-resource, and all others as low-resource.}

\textbf{(1)} {\name} with coarse-grained layer selection strategy improves multilingual capabilities in low-resource languages and cultural adaptability in low-resource cultural backgrounds. 
\textbf{(2)} However, the coarse-grained benefits for high-resource languages and high-resource cultural backgrounds are less consistent, suggesting the need for a finer-grained layer selection strategy, as further discussed in \S\ref{sec:discuss}. 
\textbf{(3)} Besides, coarse-grained {\name} demonstrates relatively stronger effectiveness on multilingual understanding task (\textit{XNLI}, \textit{XCOPA}) compared to the other two tasks.

We also observe that \textit{Qwen2-7B-Instruct} often fails to follow instructions (Supplementary Materials~C-D, even with English inputs in the vanilla setting).
We hypothesize that this instability may account for the limited effectiveness of coarse-grained {\name} on \textit{Qwen2-7B-Instruct} observed in Table~XIV.
To verify this, we conduct additional instruction-tuning (implementations are in Supplementary Materials~D) to enhance it's instruction-following ability and re-evaluate coarse-grained {\name} on OOD data. As shown by the $\Delta$ values in Table~XII (Supplementary Materials~C-D), \textit{Qwen2-7B-Instruct} with further tuning can significantly benefit more from cross-lingual transplantation under the {\name} framework. This further indicates the effectiveness of {\name} is closely tied to the model's fundamental instruction-following capability.

\textbf{Finding-6:} \textit{{\name} demonstrates great generalizability on both unseen in-distribution data and out-of-distribution data within the same task category.}

\textbf{Finding-7:} \textit{Models with stronger instruction-following capability benefit more from {\name} framework.}

\subsection{Post-Transplant Prediction Distribution}\label{sec:post}

To better understand the behavioral effects of {\name} beyond aggregate accuracy improvements, we analyze the distribution of post-transplant responses generated in \S\ref{sec:attempts} relative to the original source- and target-language predictions. Specifically, for each instance, we categorize the post-transplant output into three types: \textbf{Target-preserved}, where the prediction remains identical to the original target-language output; \textbf{Source-aligned}, where the prediction shifts to match the source-language prediction; and \textbf{Divergent}, where the response differs from both. The distribution is shown in Table~\ref{tab:post_transplant_distribution}.

Across all models, the proportion of Source-aligned responses remains consistently low. On XNLI, most post-transplant outputs are either Target-preserved or Divergent, with the latter even exceeding Source-aligned. This pattern contradicts the answer bypass hypothesis—the assumption that transplantation merely copies or directly transfers the source-language answer to the target language.
For GlobalOpinionQA, English targets are relatively more stable, yielding more Target-preserved responses. Nevertheless, the proportion of Divergent responses still surpasses Source-aligned ones.
Overall, these results suggest that {\name} does not transfer final answers directly. Instead, it modifies intermediate representations while allowing the target-language forward pass to produce its own decision, indicating that the framework influences internal reasoning rather than performing answer-level copying. Besides, we also report the distribution of a Context-Augmentation baseline in Supplementary Materials~C-F Table~XVII, where the source-language response is explicitly inserted into the target-language prompt in natural language. This comparison further clarifies the nature of the {\name} framework, showing that it functions as a form of decision-level fusion rather than a simple transfer of answers across languages.

\textbf{Finding-8:} \textit{{\name} achieves decision-level integration via transplantation between intermediate representations of the source and target languages, rather than relying on trivial answer-level copying or shortcut-based bypass.}

\begin{table}
  \centering 
  \caption{Distribution of post-transplant responses relative to the original source- and target-language predictions.}
  \resizebox{0.48\textwidth}{!}{%
      \begin{tabular}{lccc}
        \toprule
        {Dataset: XNLI} & Target-preserved & Source-aligned & Divergent \\
        \midrule
        \multicolumn{4}{c}{\textit{(Source: English, Target: Non-English)}} \\
        \midrule
        Llama-2-7b-chat & 45.49\% & 15.93\% & 38.58\% \\
        Mistral-7B-Instruct-v0.3 & 73.07\% & 12.91\% & 14.02\% \\
        Qwen2-7B-Instruct & 90.77\% & 4.19\% & 5.04\% \\
        \midrule
        {Dataset: GlobalOpinionQA} & Target-preserved & Source-aligned & Divergent \\
        \midrule
        \multicolumn{4}{c}{\textit{(Source: Non-English, Target: English)}} \\
        \midrule
        Llama-2-7b-chat & 94.31\% & 2.49\% & 3.20\% \\
        Mistral-7B-Instruct-v0.3 & 96.05\% & 0.77\% & 3.18\% \\
        Qwen2-7B-Instruct & 97.14\% & 1.34\% & 1.52\% \\
        \bottomrule
      \end{tabular}}
      \label{tab:post_transplant_distribution}
\end{table}

%%% 5_static_end

%%% 6_dynamic_start

\section{Discussions Towards Finer-grained Instance-aware Upper Bound}\label{sec:discuss}

Though the coarse-grained static layer selection strategy discussed in \S\ref{sec:attempts} yields partial improvements, the effectiveness of {\name} remains limited by the fixed choice of source and target layers across all instances.
This suggests that a finer-grained, instance-aware transplantation strategy—tailoring layer selection dynamically based on the specific input—may be important to further unlock the model's multilingual potential.

To estimate the upper bound of {\name}'s potential, we adapt an instance-aware, layer-wise evaluation. Specifically, for a model $M$ with $N$ decoder layers, {\name} supports $N$ choices for both source and target layers, resulting in $N$ candidate responses when one of the two layers is fixed. We consider an instance as correctly answered if at least one of these $N$ responses matches the golden answer. 
We denote \(M_{S_{i} \rightarrow T_{j}}(x)\) as the output of model $M$ towards question $x$ from dataset $D$ after applying {\name} from $i^{\text{th}}$ layer of language $S$ to the $j^{\text{th}}$ layer of language $T$. Let $y_{true}$ represents the gold answer of question $x$ and $\mathbb{I}(\cdot)$ is a indicator function that equals 1 if the condition is true, 0 otherwise.
The layer-wise instance-aware upper bound is formulated as follows:

\begin{equation}
  \small
  \begin{aligned}
      & \operatorname{Source-wise UpperBound}(M, D, a) =  \\
      & \frac{1}{|D|} \sum_{x \in D} \max_{\substack{i=a, j \in \{1, \dots, N\}}} \mathbb{I}({M_{S_{i} \rightarrow T_{j}}}(x) = y_{true})
      \label{eq:upper_source}
  \end{aligned}
\end{equation}

\begin{equation}
  \small
  \begin{aligned}
      & \operatorname{Target-wise UpperBound}(M, D, b) =  \\
      & \frac{1}{|D|} \sum_{x \in D} \max_{\substack{i \in \{1, \dots, N\}, j=b}} \mathbb{I}({M_{S_{i} \rightarrow T_{j}}}(x) = y_{true})
      \label{eq:upper_target}
  \end{aligned}
\end{equation}

where $a$ and $b$ represent the fixed source and target layers.

The layer-wise upper bound results in Figure~\ref{fig:layer-wise} reveal a clear trend: \textit{{\name} achieves the greatest upper bound when using the last-layer as the source or the first-layer as the target, with the latter yielding slightly greater gains.} 

Based on above observation and the findings in Section~\ref{sec:attempts}, we further investigate and propose an dynamic instance-aware layer selection strategy. Specifically, we fix the target layer to the first layer and simplify the source layer selection as a binary classification task: selecting either the first or the last layer.
To enable dynamic selection, we adapt an entropy-based strategy guided by the model's response confidence. Specifically, we follow the rule that a lower entropy in the predicted probability distribution indicates higher confidence in the response. Therefore, our selection strategy compares the entropies of the model outputs under the two candidate source layers (first or the last layer) and selects the one yielding the lower entropy. Besides, we further experiment with (1) Chain-of-Thought (CoT) prompting, which inspires the model's potential through step-by-step reasoning, and (2) multilingual supervised fine-tuning (ML-SFT), which strengthens the model's multilingual capability with additional training data (detailed implementations are in Supplementary Materials-D). To further validate the cross-model generalizability, we conduct additional experiments on \textit{Llama-3.1-8B-Instruct}, \textit{Qwen2.5-7B-Instruct} and reasoning models \textit{Qwen3-4B}, \textit{Qwen3-8B}. The results are shown in Supplementary Materials~E Table~XV,~XIV.

The comparisons in Table~\ref{tab:compare} demonstrate that (1) the upper bound within $N$ candidate responses in {\name} significantly surpasses the model's vanilla performance, underscoring both the promise of finer-grained transplantation and the extent to which current LLMs' multilingual potential remain underutilized.  
(2) And the poor performance of CoT suggests that the model's multilingual potential does not seem to be further unlocked simply through step-by-step thinking. 
(3) While both ML-SFT and our entropy-based strategy yield great improvements over vanilla performance, a considerable gap remains compared to the upper bound established by {\name} (SourceLast or TargetFirst).

\definecolor{myGreen}{RGB}{50, 168, 82}
\definecolor{myYellow}{RGB}{214, 169, 45}
\definecolor{myRed}{RGB}{201, 24, 24}

\begin{figure*}[t]
  \centering
  \begin{tikzpicture}[]
      \scriptsize{
      \begin{axis}[
      title={UpperBound Results with ``Fixed Source Layer''},
      title style={yshift=-0.6em},
	    at={(-31em,0em)}, 
      anchor=north west,  
      ymajorgrids,
      grid style=densely dashed,
      width=.46\textwidth,
      height=.28\textwidth,
      legend style={at={(-0.5,0)}, anchor=south west, text width=1.05cm, align=left, row sep=0.1cm},
      xlabel={\scriptsize{Source Layer}},
      ylabel={\scriptsize{Avg. Accuracy}},
      ylabel style={yshift=0.25em},xlabel style={yshift=0.05em},
      ymin=35,ymax=80, ytick={30,40,50,60,70,80},
      xmin=1,xmax=32, xtick={1,2,...,32},
      xticklabels={1,2,...,32},
      xticklabel style={rotate=90, font=\tiny},
      ]

      \addplot[myRed, thick] coordinates {(0,0)};
      \addlegendentry{LLaMA.}
    
      \addplot[myGreen, thick] coordinates {(0,0)};
      \addlegendentry{Mistral.}

      \addplot[myYellow, thick] coordinates {(0,0)};
      \addlegendentry{Qwen.}

      \addplot[white, thick] coordinates {(0,0)};
      \addlegendentry{} % 空行，不显示内容

      \addplot[black, thick, mark=diamond*, only marks, mark options={fill=white}] coordinates {(0,0)};
      \addlegendentry{XNLI}
    
      \addplot[black, thick, mark=*, only marks, mark options={fill=white}] coordinates {(0,0)};
      \addlegendentry{XQuAD}

      \addplot[black, thick, mark=square, only marks, mark options={fill=white}] coordinates {(0,0)};
      \addlegendentry{Global\\OpinionQA}

      \addplot[white, thick] coordinates {(0,0)};
      \addlegendentry{} % 空行，不显示内容

      % llama + XNLI
      \addplot[myRed,mark=diamond*,mark size=1.8pt,thick,mark options={fill=white,draw=myRed,line width=1.0pt}] coordinates {(1, 47.9) (2, 46.6) (3, 48.9) (4, 49.4) (5, 52.0) (6, 51.1) (7, 52.3) (8, 52.3) (9, 51.6) (10, 52.4) (11, 51.4) (12, 54.6) (13, 52.7) (14, 53.1) (15, 57.4) (16, 58.9) (17, 55.9) (18, 54.4) (19, 60.3) (20, 56.1) (21, 54.0) (22, 52.0) (23, 52.0) (24, 53.3) (25, 54.9) (26, 54.3) (27, 55.1) (28, 57.1) (29, 56.7) (30, 57.4) (31, 57.3) (32, 62.1)};

      % llama + XQuAD
      \addplot[myRed,mark=*,mark size=1.2pt,thick,mark options={fill=white,draw=myRed,line width=1.0pt}] coordinates {(1, 42.7) (2, 41.7) (3, 43.0) (4, 43.2) (5, 44.5) (6, 45.7) (7, 46.0) (8, 44.7) (9, 44.7) (10, 45.2) (11, 45.3) (12, 45.3) (13, 45.7) (14, 45.7) (15, 46.2) (16, 44.7) (17, 47.3) (18, 45.7) (19, 44.7) (20, 46.2) (21, 43.3) (22, 46.7) (23, 45.8) (24, 46.2) (25, 44.3) (26, 47.2) (27, 45.8) (28, 48.5) (29, 46.7) (30, 48.8) (31, 51.3) (32, 48.3)};

      % llama + GlobalOpinionQA
      \addplot[myRed,mark=square,mark size=1pt,thick,mark options={fill=white,draw=myRed,line width=1.0pt}] coordinates {(1, 39.6) (2, 39.5) (3, 40.0) (4, 40.1) (5, 41.7) (6, 40.4) (7, 43.7) (8, 44.1) (9, 43.1) (10, 43.1) (11, 42.6) (12, 44.2) (13, 43.2) (14, 42.1) (15, 41.8) (16, 40.3) (17, 37.5) (18, 35.6) (19, 35.6) (20, 38.1) (21, 35.9) (22, 37.6) (23, 40.2) (24, 37.5) (25, 40.2) (26, 42.4) (27, 39.1) (28, 38.5) (29, 41.9) (30, 38.9) (31, 47.3) (32, 48.0)};

      % mistral + XNLI
      \addplot[myGreen,mark=diamond*,mark size=1.8pt,thick,mark options={fill=white,draw=myGreen,line width=1.0pt}] coordinates {(1, 51.4) (2, 51.6) (3, 52.1) (4, 52.4) (5, 52.4) (6, 52.3) (7, 52.0) (8, 52.7) (9, 53.7) (10, 53.7) (11, 53.9) (12, 52.1) (13, 52.3) (14, 52.4) (15, 53.1) (16, 53.3) (17, 55.6) (18, 55.0) (19, 55.6) (20, 56.9) (21, 54.6) (22, 55.7) (23, 56.3) (24, 56.3) (25, 56.9) (26, 55.6) (27, 55.4) (28, 56.0) (29, 57.9) (30, 58.9) (31, 55.4) (32, 56.7)};

      % mistral + XQuAD
      \addplot[myGreen,mark=*,mark size=1.2pt,thick,mark options={fill=white,draw=myGreen,line width=1.0pt}] coordinates {(1, 50.2) (2, 49.8) (3, 49.7) (4, 49.0) (5, 50.0) (6, 49.7) (7, 50.8) (8, 51.3) (9, 49.8) (10, 52.2) (11, 50.7) (12, 52.0) (13, 49.8) (14, 52.7) (15, 53.2) (16, 51.5) (17, 51.3) (18, 50.5) (19, 51.7) (20, 55.0) (21, 51.0) (22, 51.2) (23, 51.7) (24, 55.7) (25, 53.8) (26, 54.2) (27, 55.2) (28, 55.5) (29, 57.0) (30, 55.3) (31, 56.8) (32, 58.0)};

      % mistral + GlobalOpinionQA
      \addplot[myGreen,mark=square,mark size=1pt,thick,mark options={fill=white,draw=myGreen,line width=1.0pt}] coordinates {(1, 71.1) (2, 71.0) (3, 70.9) (4, 71.8) (5, 71.7) (6, 71.6) (7, 72.1) (8, 71.9) (9, 72.8) (10, 72.2) (11, 72.4) (12, 72.5) (13, 72.9) (14, 72.7) (15, 72.2) (16, 71.9) (17, 72.5) (18, 72.4) (19, 72.2) (20, 72.7) (21, 72.4) (22, 72.2) (23, 72.9) (24, 72.6) (25, 73.0) (26, 72.7) (27, 73.0) (28, 73.1) (29, 73.1) (30, 74.7) (31, 75.4) (32, 75.5)};

      % qwen + XNLI
      \addplot[myYellow,mark=diamond*,mark size=1.8pt,thick,mark options={fill=white,draw=myYellow,line width=1.0pt}] coordinates {(1, 61.4) (2, 62.1) (3, 61.6) (4, 61.6) (5, 61.9) (6, 61.6) (7, 61.4) (8, 61.4) (9, 61.9) (10, 62.4) (11, 62.3) (12, 61.7) (13, 62.3) (14, 62.4) (15, 62.6) (16, 62.1) (17, 62.1) (18, 64.1) (19, 63.3) (20, 62.9) (21, 63.7) (22, 63.1) (23, 63.4) (24, 62.0) (25, 60.7) (26, 60.4) (27, 59.9) (28, 60.1)};

      % qwen + XQuAD
      \addplot[myYellow,mark=*,mark size=1.2pt,thick,mark options={fill=white,draw=myYellow,line width=1.0pt}] coordinates {(1, 57.2) (2, 56.7) (3, 56.8) (4, 56.3) (5, 57.3) (6, 57.3) (7, 58.2) (8, 59.0) (9, 63.3) (10, 62.3) (11, 58.0) (12, 62.3) (13, 62.5) (14, 60.0) (15, 61.7) (16, 59.8) (17, 60.0) (18, 60.2) (19, 59.8) (20, 61.0) (21, 63.5) (22, 63.8) (23, 61.8) (24, 65.0) (25, 63.7) (26, 64.7) (27, 70.3) (28, 67.7)};

      % qwen + GlobalOpinionQA
      \addplot[myYellow,mark=square,mark size=1pt,thick,mark options={fill=white,draw=myYellow,line width=1.0pt}] coordinates {(1, 65.5) (2, 65.8) (3, 65.7) (4, 66.1) (5, 65.9) (6, 66.4) (7, 65.9) (8, 66.9) (9, 66.3) (10, 66.4) (11, 69.0) (12, 67.3) (13, 67.6) (14, 67.6) (15, 66.3) (16, 67.0) (17, 66.3) (18, 67.3) (19, 67.9) (20, 68.5) (21, 68.6) (22, 68.5) (23, 68.0) (24, 67.0) (25, 67.0) (26, 68.4) (27, 75.7) (28, 69.1)};

      \end{axis}
     }

     \scriptsize{
      \begin{axis}[
      title={UpperBound Results with ``Fixed Target Layer''},
      title style={yshift=-0.6em},
      at={(0em,0em)},  % 这里定位在左上角
      anchor=north west,  % 左上角对齐
      ymajorgrids,
      grid style=densely dashed,
      width=.46\textwidth,
      height=.28\textwidth,
      xlabel={\scriptsize{Target Layer}},
      ylabel style={yshift=0.25em},xlabel style={yshift=0.05em},
      ymin=32,ymax=80, ytick={30,40,50,60,70,80},
      xmin=1,xmax=32,xtick={1,2,...,32},
      xticklabels={1,2,...,32},
      xticklabel style={rotate=90, font=\tiny},
      legend style={yshift=-6pt,xshift=-1em, legend plot pos=right,font={\footnotesize},cells={anchor=west}},
      ]
      
      \addplot[myRed,mark=diamond*,mark size=1.8pt,thick,mark options={fill=white,draw=myRed,line width=1.0pt}] coordinates {(1, 67.6) (2, 67.4) (3, 64.3) (4, 65.9) (5, 65.4) (6, 63.6) (7, 63.4) (8, 61.6) (9, 61.4) (10, 63.9) (11, 60.3) (12, 58.7) (13, 59.0) (14, 59.6) (15, 57.4) (16, 54.7) (17, 53.3) (18, 50.1) (19, 48.1) (20, 46.4) (21, 46.9) (22, 46.4) (23, 44.7) (24, 45.6) (25, 42.6) (26, 43.9) (27, 43.3) (28, 45.1) (29, 44.6) (30, 43.6) (31, 43.0) (32, 42.3)};
      
      \addplot[myRed,mark=*,mark size=1.2pt,thick,mark options={fill=white,draw=myRed,line width=1.0pt}] coordinates {(1, 53.2) (2, 52.2) (3, 50.3) (4, 48.5) (5, 49.2) (6, 47.8) (7, 47.3) (8, 45.7) (9, 45.2) (10, 46.0) (11, 46.5) (12, 45.8) (13, 46.8) (14, 45.3) (15, 45.3) (16, 45.8) (17, 45.0) (18, 46.5) (19, 45.0) (20, 43.5) (21, 43.7) (22, 43.3) (23, 42.2) (24, 41.0) (25, 41.7) (26, 39.8) (27, 39.7) (28, 39.3) (29, 40.5) (30, 38.5) (31, 38.2) (32, 37.5)};
      
      \addplot[myRed,mark=square,mark size=1pt,thick,mark options={fill=white,draw=myRed,line width=1.0pt}] coordinates 
      {(1, 53.8) (2, 51.4) (3, 49.8) (4, 50.1) (5, 49.0) (6, 46.6) (7, 45.5) (8, 44.2) (9, 43.4) (10, 44.1) (11, 41.2) (12, 41.9) (13, 42.1) (14, 42.1) (15, 41.9) (16, 42.4) (17, 40.5) (18, 39.2) (19, 38.3) (20, 37.0) (21, 37.4) (22, 37.4) (23, 35.6) (24, 35.5) (25, 36.7) (26, 36.6) (27, 35.4) (28, 37.0) (29, 34.4) (30, 34.7) (31, 35.3) (32, 33.5)};
      
      \addplot[myGreen,mark=diamond*,mark size=1.8pt,thick,mark options={fill=white,draw=myGreen,line width=1.0pt}] coordinates 
      {(1, 62.7) (2, 61.9) (3, 62.0) (4, 61.9) (5, 61.7) (6, 62.4) (7, 62.3) (8, 61.7) (9, 61.6) (10, 61.9) (11, 61.1) (12, 61.0) (13, 60.7) (14, 60.4) (15, 59.3) (16, 58.9) (17, 58.6) (18, 56.3) (19, 56.6) (20, 55.7) (21, 55.4) (22, 54.0) (23, 53.6) (24, 53.6) (25, 53.0) (26, 56.0) (27, 53.7) (28, 52.9) (29, 51.0) (30, 50.0) (31, 45.4) (32, 45.0)};
      
      \addplot[myGreen,mark=*,mark size=1.2pt,thick,mark options={fill=white,draw=myGreen,line width=1.0pt}] coordinates 
      {(1, 60.7) (2, 58.8) (3, 58.0) (4, 56.7) (5, 57.5) (6, 57.5) (7, 55.2) (8, 54.8) (9, 55.8) (10, 55.7) (11, 56.0) (12, 55.3) (13, 53.7) (14, 53.0) (15, 53.5) (16, 55.0) (17, 51.7) (18, 51.7) (19, 50.8) (20, 48.3) (21, 46.7) (22, 45.7) (23, 45.3) (24, 44.7) (25, 45.5) (26, 44.8) (27, 45.0) (28, 44.8) (29, 44.8) (30, 45.7) (31, 46.3) (32, 41.2)};
      
      \addplot[myGreen,mark=square,mark size=1pt,thick,mark options={fill=white,draw=myGreen,line width=1.0pt}] coordinates 
      {(1, 74.6) (2, 75.5) (3, 75.3) (4, 75.1) (5, 75.0) (6, 74.2) (7, 73.7) (8, 74.0) (9, 73.0) (10, 73.0) (11, 73.0) (12, 72.8) (13, 73.3) (14, 72.7) (15, 73.5) (16, 73.8) (17, 73.6) (18, 72.4) (19, 72.6) (20, 71.7) (21, 71.7) (22, 71.5) (23, 71.0) (24, 71.0) (25, 71.5) (26, 71.0) (27, 70.7) (28, 70.9) (29, 70.1) (30, 68.8) (31, 70.4) (32, 69.4)};
      
      \addplot[myYellow,mark=diamond*,mark size=1.8pt,thick,mark options={fill=white,draw=myYellow,line width=1.0pt}] coordinates 
      {(1, 61.6) (2, 61.1) (3, 60.3) (4, 63.9) (5, 61.6) (6, 60.7) (7, 60.1) (8, 59.9) (9, 60.0) (10, 60.1) (11, 59.6) (12, 60.3) (13, 60.3) (14, 61.1) (15, 60.7) (16, 61.1) (17, 60.6) (18, 60.4) (19, 59.9) (20, 59.7) (21, 60.9) (22, 58.7) (23, 58.9) (24, 58.7) (25, 58.1) (26, 58.1) (27, 57.1) (28, 59.7)};
      
      \addplot[myYellow,mark=*,mark size=1.2pt,thick,mark options={fill=white,draw=myYellow,line width=1.0pt}] coordinates 
      {(1, 71.5) (2, 64.8) (3, 63.2) (4, 61.8) (5, 61.0) (6, 61.2) (7, 62.5) (8, 62.0) (9, 63.0) (10, 62.7) (11, 62.0) (12, 62.3) (13, 62.0) (14, 62.8) (15, 62.7) (16, 62.3) (17, 61.7) (18, 60.2) (19, 61.7) (20, 61.8) (21, 58.2) (22, 57.2) (23, 54.7) (24, 54.5) (25, 54.2) (26, 54.7) (27, 55.3) (28, 48.2)};
      
      \addplot[myYellow,mark=square,mark size=1pt,thick,mark options={fill=white,draw=myYellow,line width=1.0pt}] coordinates 
      {(1, 73.9) (2, 69.0) (3, 69.5) (4, 69.0) (5, 68.8) (6, 68.9) (7, 68.2) (8, 68.2) (9, 68.6) (10, 69.2) (11, 69.2) (12, 69.4) (13, 69.3) (14, 68.4) (15, 69.1) (16, 68.9) (17, 69.0) (18, 69.3) (19, 69.4) (20, 69.1) (21, 68.0) (22, 67.2) (23, 67.7) (24, 68.2) (25, 66.3) (26, 65.3) (27, 63.8) (28, 63.7)};
      \end{axis}
     }
    \end{tikzpicture}
  \caption{The layer-wise instance-aware upper bound results across different LLMs and \textit{PilotSets}. The left represents the source-wise upper bound and the right represents the target-wise upper bound. LLaMA., Mistral. and Qwen. respectively represent \textit{LLaMA-2-7B-Chat}, \textit{Mistral-7B-Instruct-v0.3} and \textit{Qwen2-7B-Instruct}.} 
  \label{fig:layer-wise}
\end{figure*}

\begin{table}
  \centering 
  \caption{Performance Comparisons Between Vanilla Model, CoT, Multilingual SFT And Upper Bound Results Under Fixed Source (Last-Layer) Or Target (First-Layer) Settings On \textit{PilotSets}.}
  \resizebox{0.48\textwidth}{!}{%
      \begin{tabular}{lccccccccc}
          \toprule
          \multirow{4}{*}{\textbf{Settings}} & \multicolumn{9}{c}{\bf \textit{PilotSets}} \\ 
          \cmidrule(lr){2-10}
          & \multicolumn{3}{c}{\bf XNLI} & \multicolumn{3}{c}{\bf XQuAD} & \multicolumn{3}{c}{\bf GlobalOpinionQA} \\
          \cmidrule(lr){2-4}
          \cmidrule(lr){5-7}
          \cmidrule(lr){8-10}
          & High & Mid & Low & High & Mid & Low & High & Mid & Low \\
          \midrule
          \multicolumn{10}{c}{\textbf{\textit{LLaMA-2-7B-Chat}}} \\
          \midrule
          Vanilla & 44.3 & 29.0 & 17.5 & 45.6 & 28.5 & 20.0 & 34.9 & 35.7 & 23.7\\
          CoT & 25.3 & 17.5 & 13.5 & 42.4 & 24.5 & 16.7 & 17.1 & 17.3 & 16.6 \\
          ML-SFT & 34.3 & 36.0 & 28.5 & 46.4 & 35.5 & 47.3 & 39.1 & 40.0 & 30.6 \\
          \midrule
          \multicolumn{10}{c}{ Finer-grained Attempt under {\name} Framework} \\
          \midrule
          Entropy-based & 46.0 & 34.5 & 31.0 & 46.0 & 29.0 & 19.3 & 35.7 & 35.7 & 25.1 \\
          \midrule
          \multicolumn{10}{c}{ Instance-aware Upper Bound} \\
          \midrule
          SourceLast & 62.3 & 60.0 & 64.0 & 63.2 & 42.0 & 32.0 & 48.3 & 56.0 & 40.9 \\
          TargetFirst & 66.0 & 69.0 & 68.5 & 66.4 & 50.5 & 34.7 & 54.9 & 58.7 & 48.6 \\
          \midrule
          \midrule

          \multicolumn{10}{c}{\textbf{\textit{Mistral-7B-Instruct-v0.3}}} \\
          \midrule
          Vanilla & 43.3 & 41.0 & 27.5 & 47.2 & 38.5 & 29.3 & 70.6 & 68.3 & 62.9 \\
          \midrule
          \multicolumn{10}{c}{Methods for Comparison} \\
          \midrule
          CoT & 40.7 & 22.5 & 17.5 & 52.0 & 24.5 & 22.0 & 41.1 & 41.3 & 29.1 \\
          ML-SFT & 40.7 & 41.5 & 33.5 & 46.0 & 39.0 & 48.7 & 66.9 & 64.3 & 57.1 \\
          Entropy-based & 44.0 & 49.5 & 31.0 & 46.8 & 38.0 & 30.7 & 69.4 & 68.7 & 62.9 \\
          \midrule
          \multicolumn{10}{c}{Instance-aware Upper Bound} \\
          \midrule
          SourceLast & 57.0 & 62.5 & 50.5 & 67.2 & 55.0 & 46.7 & 79.4 & 78.3 & 69.1 \\
          TargetFirst & 62.3 & 69.5 & 56.5 & 68.0 & 56.0 & 54.7 & 77.4 & 76.0 & 70.6 \\
          \midrule
          \midrule
          
          \multicolumn{10}{c}{\textbf{\textit{Qwen2-7B-Instruct}}} \\
          \midrule
          Vanilla & 65.3 & 56.0 & 41.0 & 54.0 & 44.0 & 32.0 & 67.4 & 63.7 & 57.1 \\
          CoT & 59.3 & 43.0 & 19.0 & 60.0 & 51.5 & 33.3 & 41.1 & 43.7 & 37.1 \\
          ML-SFT & 52.7 & 40.0 & 43.0 & 62.0 & 57.5 & 62.0 & 62.9 & 60.7 & 52.6 \\
          \midrule
          \multicolumn{10}{c}{Finer-grained Attempt under {\name} Framework} \\
          \midrule
          Entropy-based & 63.7 & 51.0 & 51.0 & 52.8 & 45.0 & 32.0 & 67.4 & 64.3 & 57.1 \\
          \midrule
          \multicolumn{10}{c}{Instance-aware Upper Bound} \\
          \midrule
          SourceLast & 67.7 & 62.5 & 46.5 & 77.6 & 67.0 & 52.0 & 71.4 & 70.3 & 65.7 \\
          TargetFirst & 68.0 & 65.5 & 48.0 & 79.6 & 69.0 & 61.3 & 77.4 & 75.3 & 69.1 \\
          \bottomrule
      \end{tabular}}

      \label{tab:compare}
\end{table}

\begin{table}[ht]
  \centering
  \scriptsize
  \renewcommand{\arraystretch}{1.05} 
  \setlength{\tabcolsep}{3pt}
  \caption{Performance comparison of applying entropy-based layer selection strategy on \textit{LLaMA-2-7B-Chat} and \textit{Chinese-Alpaca-2-7B}, using activations sourced from English and Chinese inputs.}
  \resizebox{0.48\textwidth}{!}{%
  \begin{tabular}{lcccccccccccccccc}
  \toprule
  {\bf \shortstack{XNLI}} & en & ar & bg & de & el & es & fr & hi & ru & sw & tr & ur & vi & zh & \textbf{Avg.} \\
  \midrule
  \multicolumn{16}{c}{\bf \textit{LLaMA-2-7B-Chat}} \\
  \midrule
  Entropy-based (English) & 60.0 & 34.0 & 42.0 & 52.0 & 30.0 & 36.0 & 54.0 & 24.0 & 40.0 & 28.0 & 38.0 & 24.0 & 36.0 & 40.0 & \textbf{38.4} \\
  \noalign{\vskip 0.2ex}
  Entropy-based (Chinese) & 62.0 & 34.0 & 30.0 & 50.0 & 34.0 & 38.0 & 52.0 & 16.0 & 46.0 & 22.0 & 32.0 & 6.0 & 34.0 & 32.0 & \textbf{34.9} \\
  \midrule
  \multicolumn{16}{c}{\bf \textit{Chinese-Alpaca-2-7B}} \\
  \midrule
  Entropy-based (English) & 38.0 & 8.0 & 0.0 & 18.0 & 30.0 & 34.0 & 4.0 & 2.0 & 28.0 & 6.0 & 2.0 & 10.0 & 22.0 & 28.0 & \textbf{16.4} \\
  \noalign{\vskip 0.2ex}
  Entropy-based (Chinese) & 40.0 & 10.0 & 10.0 & 28.0 & 20.0 & 34.0 & 12.0 & 2.0 & 24.0 & 10.0 & 18.0 & 0.0 & 24.0 & 34.0 & \textbf{19.0} \\
  
  \end{tabular}}
  \resizebox{0.48\textwidth}{!}{%
  \begin{tabular}{lcccccccccccccc}
  \toprule
  {\bf \shortstack{XQuAD}} & en & ar & de & el & es & hi & ro & ru & th & tr & vi & zh & \textbf{Avg.} \\
  \midrule
  \multicolumn{14}{c}{\bf \textit{LLaMA-2-7B-Chat}} \\
  \midrule
  Entropy-based (English) & 64.0 & 8.0 & 58.0 & 12.0 & 60.0 & 8.0 & 40.0 & 42.0 & 6.0 & 24.0 & 42.0 & 40.0 & \textbf{33.7} \\
  \noalign{\vskip 0.2ex}
  Entropy-based (Chinese) & 62.0 & 10.0 & 56.0 & 12.0 & 64.0 & 6.0 & 42.0 & 42.0 & 6.0 & 22.0 & 38.0 & 40.0 & \textbf{33.3} \\
  
  \midrule
  \multicolumn{14}{c}{\bf \textit{Chinese-Alpaca-2-7B}} \\
  \midrule
  Entropy-based (English) & 62.0 & 0.0 & 40.0 & 4.0 & 38.0 & 0.0 & 30.0 & 36.0 & 2.0 & 12.0 & 34.0 & 48.0 & \textbf{25.5} \\
  \noalign{\vskip 0.2ex}
  Entropy-based (Chinese) & 60.0 & 0.0 & 40.0 & 6.0 & 44.0 & 0.0 & 32.0 & 30.0 & 2.0 & 12.0 & 34.0 & 56.0 & \textbf{26.3} \\
  \bottomrule
  
  \end{tabular}}
  
  \label{tab:chinese}
\end{table}

\textbf{Finding-9:} \textit{Cross-lingual latent transplantation demonstrates substantial potential within a constrained candidate space of size $N$ (e.g., using either source-last or target-first), exceeding the model's vanilla performance by a significant margin.}

\textbf{Open Challenges:} \textit{Though {\name} taps into latent-level cross-lingual interactions and delivers modest gains in our coarse-grained and entropy-based attempts, closing the gap to its instance-wise upper bound remains an open problem. Even large-scale multilingual SFT falls short of this ceiling, suggesting that the multilingual potential of current models is both immense and deeply elusive. Within {\name} framework, future work can explore more dynamic, instance-aware layer selection strategies—potentially via lightweight meta-learning or learned controllers—that can adapt layer selection on a per-instance basis. Moreover, understanding which linguistic or cultural features drive optimal layer choices will be critical for designing hybrid approaches that integrate latent manipulation with minimal fine-tuning, striking the right balance between data efficiency and maximal utilization of the model's internalized multilingual knowledge.}

\textbf{Discussions Towards Layer-wise Functions:} \textit{Recent layer-level analysis converge on a shared but incomplete picture of non-trivial layer-wise organization of multilingual representations in LLMs. One line of research emphasizes \textbf{latent dominance and anchoring}~\cite{wendler-etal-2024-llamas, alabi-etal-2024-hidden}, showing that intermediate representations are often English-centric and the prediction trajectories tend to evolve primarily in the source language across most layers, with the target language becoming prominent only in the final layers. These findings reveal an asymmetry in how languages propagate through depth. Another line highlights \textbf{structural modularity and specialization}~\cite{choenni-etal-2024-examining, bandarkar2025layer, wu2025semantic}, identifying language-specialized subnetworks and demonstrating that swapping top or bottom layers with language experts substantially alters cross-lingual transfer. Meanwhile, certain layers appear to function as shared semantic hubs that align representations across languages. The substantial variation in our layer-wise upper bound analysis supports the existence of partial modularity and depth asymmetry. In particular, the superior upper bound derived from \emph{source-last} and \emph{target-first} configurations suggests that abstractions are comparatively transferable, while boundary layers remain more language-sensitive entry and exit points of computation. In this sense, {\name} does not merely analyze layer structure—it exposes a routing bottleneck: multilingual knowledge is present, stratified, and partially modular, yet insufficiently integrated during standard forward propagation.}

%%% 6_dynamic_end

%%% 7_further_start

\section{Further Analysis}

\subsection{Experiments on Chinese-centric LLM}\label{sec:chinese}

Our main experiments mainly focus on several representative English-centric LLMs. In this section, we further explore the generalizability of {\name} on non-English dominant LLM. We compare the effects of entropy-based layer selection on \textit{LLaMA-2-7B-Chat} and \textit{Chinese-Alpaca-2-7B}\footnote{A model based on \textit{LLaMA-2-7B}, secondarily pre-trained and further instruction-tuned on Chinese data.}, examining how activations from each model's dominant language (English vs. Chinese) contribute to performance. As shown in Table~\ref{tab:chinese}, for \textit{LLaMA-2-7B-Chat}, activations from English yields better performance than from Chinese. In contrast, for \textit{Chinese-Alpaca-2-7B}, Chinese activations lead to better results. These findings suggest a form of native preference, where latent activations from the model's dominant language produce greater improvements, likely due to closer alignment with the internalized knowledge of the model.

\begin{table}
  \centering 
  \renewcommand{\arraystretch}{1.1}
  \setlength\tabcolsep{4pt}
  \caption{The Averaged Results of {\name} Attempts on Long-form Answering.}
  \resizebox{0.48\textwidth}{!}{%
      \begin{tabular}{lcccccccccccc}
          \toprule
          {\bf {Accuracy on MGSM}} & en & bn & de & es & fr & ja & ru & sw & te & th & zh &\bf Avg.\\
          \midrule
          \multicolumn{13}{c}{\textbf{\makecell{Llama-2-7b-Chat}}} \\
          \midrule
          Backbone Model & 26.0 & 0.0 & 22.4 & 22.8 & 20.4 & 15.2 & 14.0 & 2.0 & 0.0 & 0.0 & 16.0 &\bf 12.6 \\
          \noalign{\vskip 0.2ex}\cdashline{1-13}\noalign{\vskip 0.4ex}
          {\name} \\
          -- (Attn-level, Last-to-First) & 27.2 & 0.0 & 24.4 & 23.6 & 24.0 & 17.6 & 16.4 & 4.0 & 0.0 & 2.0 & 18.4 &\bf 14.3 \\
          -- (FFN-level, First-to-First) & 26.0 & 0.0 & 22.4 & 22.4 & 20.8 & 17.6 & 14.0 & 2.0 & 0.0 & 1.2 & 18.0 &\bf 13.1 \\
          \midrule
          \multicolumn{13}{c}{\textbf{\makecell{Mistral-7B-Instruct-v0.3}}} \\
          \midrule
          Backbone Model & 52.0 & 10.4 & 41.2 & 43.6 & 5.2 & 25.6 & 38.4 & 10.4 & 2.0 & 18.0 & 37.2 &\bf 25.8 \\
          \noalign{\vskip 0.2ex}\cdashline{1-13}\noalign{\vskip 0.4ex}
          {\name} \\
          -- (Attn-level, Last-to-First) & 50.0 & 14.4 & 41.2 & 41.2 & 7.6 & 24.8 & 40.0 & 10.0 & 2.4 & 18.0 & 36.8 &\bf 26.0 \\
          -- (FFN-level, First-to-First) & 52.0 & 11.2 & 41.2 & 44.0 & 5.6 & 25.2 & 38.0 & 10.8 & 2.4 & 18.0 & 37.6 &\bf 26.0 \\
          \midrule
          \multicolumn{13}{c}{\textbf{\makecell{Qwen2-7B-Instruct}}} \\
          \midrule
          Backbone Model & 73.6 & 23.6 & 67.2 & 68.8 & 67.6 & 51.6 & 70.8 & 20.0 & 9.2 & 51.2 & 68.8 &\bf 52.0\\
          \noalign{\vskip 0.2ex}\cdashline{1-13}\noalign{\vskip 0.4ex}
          {\name} \\
          -- (Attn-level, Last-to-First) & 76.4 & 37.6 & 68.0 & 69.6 & 69.6 & 53.2 & 64.0 & 18.4 & 12.0 & 54.8 & 60.8 &\bf 53.1 \\
          -- (FFN-level, First-to-First) & 73.6 & 23.2 & 67.6 & 69.6 & 66.8 & 51.2 & 72.0 & 19.6 & 9.2 & 50.8 & 70.4 &\bf 52.2\\
          \bottomrule
      \end{tabular}}
      \label{tab:multi_step}
\end{table}

\subsection{{\name} Attempts on Long-form Answering}

As introduced in \S\ref{sec:method}, {\name} applies transplantation only at the first newly generated token. Although this single-step intervention can influence subsequent decoding in autoregressive generation, it remains unclear whether its effect diminishes during long-form generation.
To investigate this, we conduct additional experiments on \textit{MGSM}~\cite{shi2023language}, a multilingual math reasoning benchmark that requires multi-step, long-form generation (e.g., step-by-step reasoning), and the results are shown in Table~\ref{tab:multi_step}. Beyond our default setting (a single transplantation at the first new token following the setup in Table~\ref{tab:attempts_setting}), we further explore a multi-step strategy that injects English activations every 10 generated tokens during decoding, and the results are shown in Supplementary Materials~E Table~XVIII.

The results demonstrate that {\name} provides positive gains in tasks requiring long-form multi-step reasoning, suggesting its effectiveness beyond short-form generation. However, applying repeated interventions every 10 tokens causes a substantial performance drop, likely due to semantic misalignment between tokens across languages. This finding is supported by the observation of model outputs, where transplanting mid-generation often leads to semantic drift and broken sentence structure as shown by the cases in Supplementary Materials~E Figure~9. These examples suggest that repeatedly transplanting activations without fine-grained token-level semantic alignment may disrupt generation.

\begin{figure}[t]
  \begin{center}
  \includegraphics[width=0.45\textwidth]{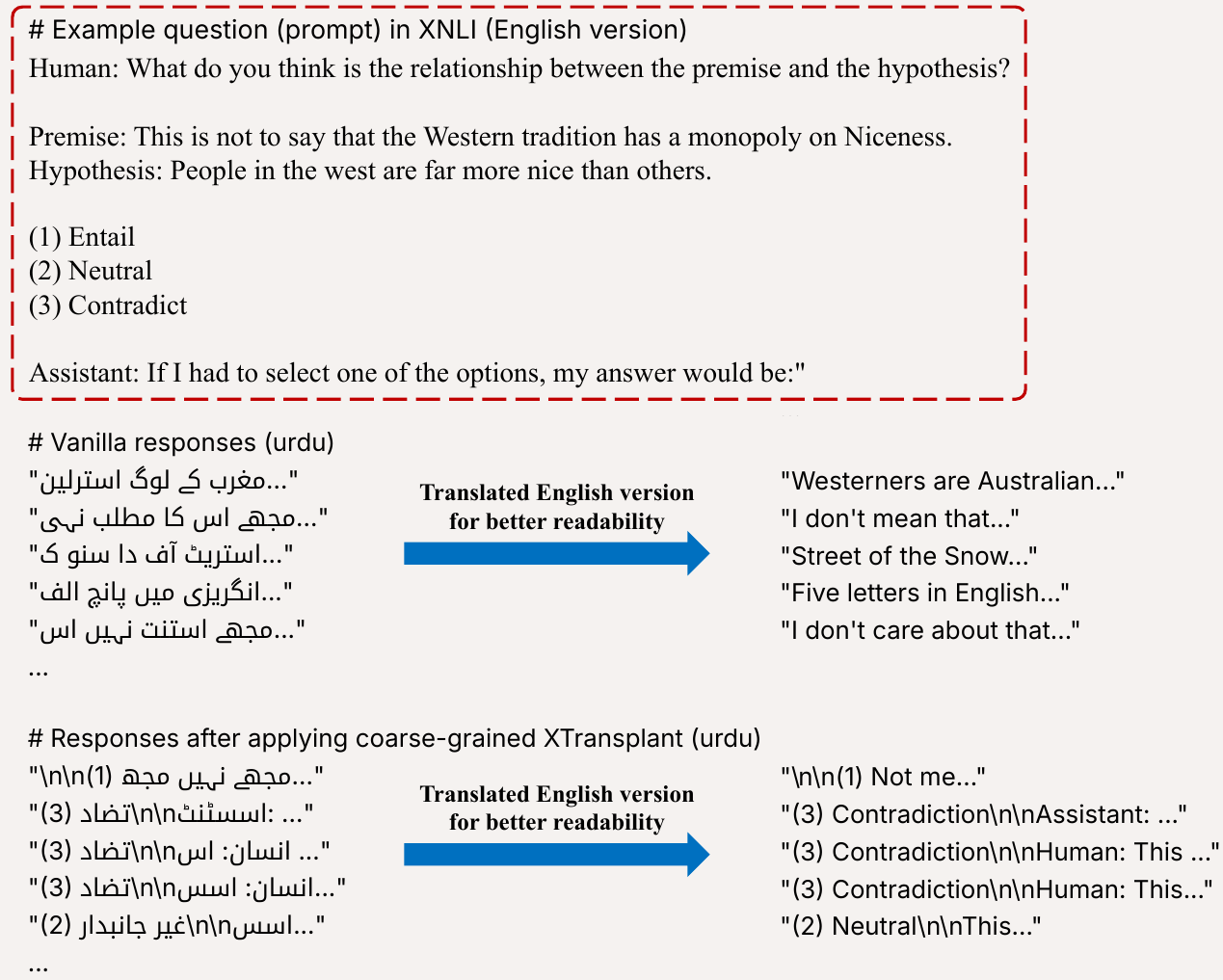}
  \end{center}
  \caption{Comparisons between the responses of \textit{LLaMA-2-7B-Chat} before and after applying coarse-grained {\name} on the Urdu (ur) subset of \textit{XNLI}.}
  \label{fig:case_multilingual}
\end{figure}

\begin{figure*}[ht]
  \begin{center}
  \includegraphics[width=0.95\textwidth]{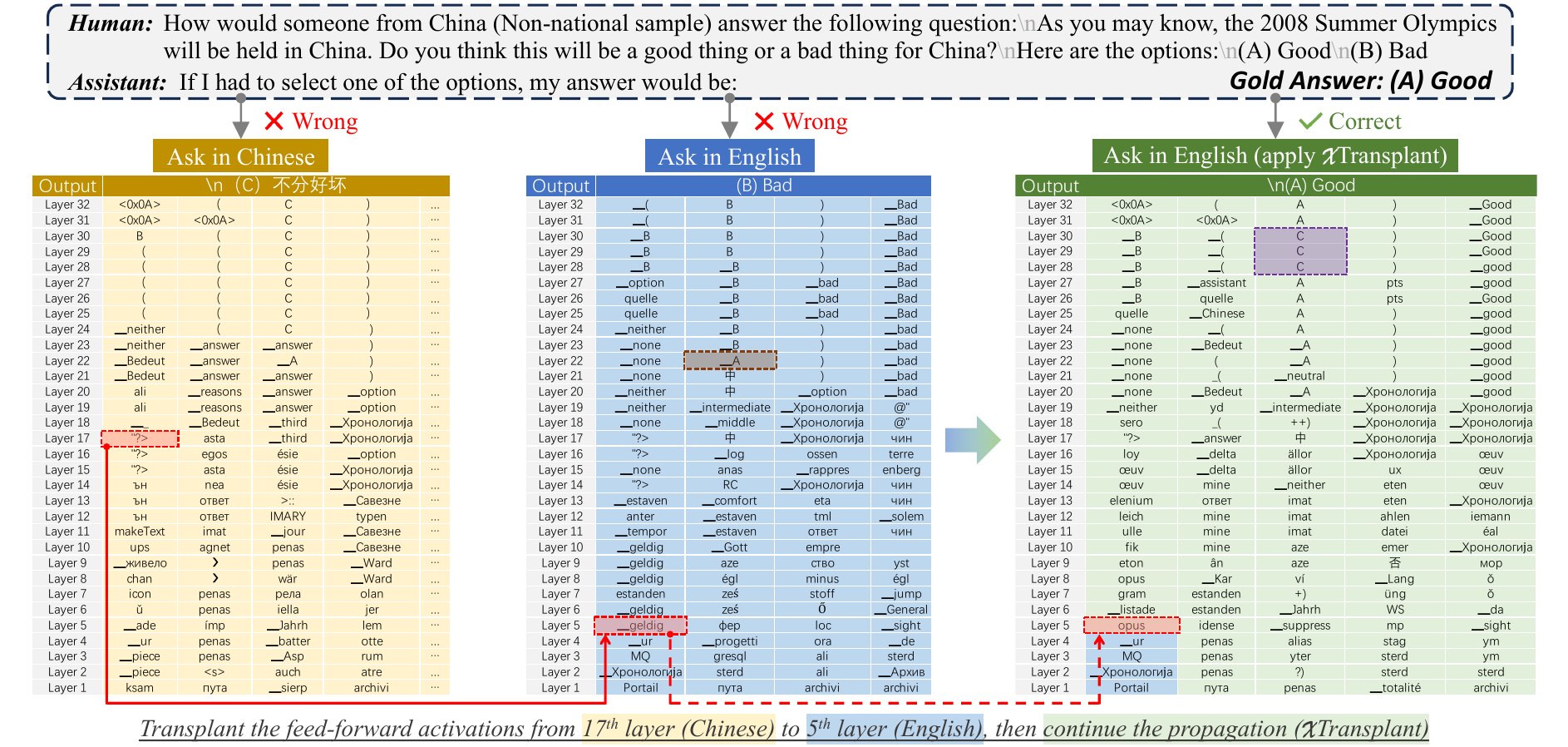}
  \end{center}
  \caption{A intermediate decoding case study of transplanting the feed forward activations from Chinese to English, compared with its original responses when prompting in Chinese and English.}
  \label{fig:case_cultural}
\end{figure*}

\subsection{Illustrative case studies}
To better understand the impact of {\name}, we present illustrative case studies from two perspectives: multilingual capability and cultural adaptability. 

For multilingual capability, we examine the representative cases on the Urdu (ur) subset of \textit{XNLI} (from Table~\ref{tab:attempts_results_llama}), where coarse-grained {\name} leads to a dramatic performance increase from 0.5\% to 29.9\% on \textit{LLaMA-2-7B-Chat}. In Figure~9 (Supplementary Materials~F), we present the responses of \textit{LLaMA-2-7B-Chat} before and after applying coarse-grained {\name} to see the exact differences. 
We observe that the vanilla \textit{LLaMA-2-7B-Chat} exhibits poor instruction-following ability in Urdu, often failing to execute the given instructions altogether. In contrast, after applying coarse-grained {\name}, the model's instruction adherence in Urdu improves significantly. This qualitative shift helps explain the dramatic performance increase from 0.5\% to 29.9\% observed on the Urdu subset.

For cultural adaptability, we present a case study in Figure~\ref{fig:case_cultural} using an interpretable form of intermediate decoding\footnote{\url{https://www.lesswrong.com/posts/fJE6tscjGRPnK8C2C/decoding-intermediate-activations-in-llama-2-7b}}, which reveals how {\name} modifies the model's output step by step.
The example question in Figure~\ref{fig:case_cultural} is a real case from the \textit{GlobalOpinionQA} dataset, with all responses generated by \textit{LLaMA-2-7B-Chat}.
We present the model's responses for the \textit{Ask-in-Chinese} prompt, \textit{Ask-in-English} prompt, and a response selected from the $N^2$ answer space of {\name} from Chinese to English. As shown, when prompted in Chinese, \textit{LLaMA-2-7B-Chat}, due to its limited proficiency in Chinese, produced a hallucinated response \texttt{(C)} that was not among the given answer options. When prompted in English, \textit{LLaMA-2-7B-Chat} also provided an incorrect answer \texttt{(B)}. However, by checking the intermediate decoding process of \textit{Ask-in-English}, we found that \textit{LLaMA-2-7B-Chat} had the potential to produce the correct answer, as highlighted in the \textcolor{myBrown}{\bf brown box}. By applying {\name} from the 17th layer (Chinese) to the 5th layer (English), the feed-forward activations from Chinese successfully guided the model to give the correct answer \texttt{(A)}. Nevertheless, as highlighted in \textcolor{myPurple}{\bf purple box}, there is also a risk of over-guidance with {\name}, where the source language may excessively influence the model's decision making.

%%% 7_further_end

%%% 8_related_start

\section{Related Work}
\subsection{Multilingual Capability.}
Early multilingual models like mBERT~\cite{devlin-etal-2019-bert} and XLM~\cite{conneau2019cross} laid the groundwork for extending pretrained models across diverse languages. 
Recently larger multilingual models, such as Bloom~\cite{scao2022bloom} and Mala-500~\cite{lin2024mala}, enhance multilingual capabilities through increased scale. Generally, multilingual pretraining and finetuning are now the two mainstream methods for improving multilingual performance.
Works like \cite{li2024prealign} injects multilingual alignment and preserves this during pretraining. 
\cite{gao2024multilingual} explored the effect of multilingual pretraining and instruction tuning on the degree of alignment. Models like
Sabia~\cite{pires2023sabia}, ChineseLLaMA~\cite{cui2023efficient}, ChineseMixtral~\cite{Chinese-Mixtral-8x7B} are products of continuous pretraining on existing English-centric LLMs.
Other like BLOOMz~\cite{muennighoff-etal-2023-crosslingual}, m-LLaMA~\cite{zhu2023extrapolating}, Phoenix~\cite{chen2023phoenix} chosen to directly incorporate multilingual data in the supervised finetuning stage to achieve implicit multilingual alignment across languages.

\subsection{Cultural Adaptability.}
Current LLMs exhibit poor cultural adaptability~\cite{ramezani-xu-2023-knowledge, jha-etal-2023-seegull, rao-etal-2025-normad}.
Solutions towards these culture-aware challenges can be categorized mainly into two approaches: context learning and training-based. \cite{kovavc2023large} studied models' controllability in inducing cultural perspectives, while \cite{wang-etal-2024-countries} improved cultural performance by explicitly prompting LLMs with the recognition of culture in queries. \cite{rao-etal-2023-ethical} developed a framework integrating moral dilemmas with principles from various normative ethics formalisms across different levels of abstraction.
\cite{rao-etal-2023-ethical} developed a framework integrating ethics from diverse cultures.
Another line of research involves fine-tuning models on large-scale culturally relevant datasets~\cite{abbasi2023persianllama, lin2023taiwan, nguyen-etal-2024-seallms, shi2024culturebank}, or investing in more balanced multilingual corpus for pretraining~\cite{scao2022bloom, lin2024mala, gao2024multilingual, li2024prealign}.

\subsection{Cross-lingual Representation Intervention.}

Recent studies have shown that multilingual performance gaps in LLMs are closely tied to language-specific divergences in internal representations.
\cite{sundar-etal-2025-steering} show that intervening on language-specific neurons in multilingual LLMs reshapes the embedding space and improves cross-lingual representation alignment without fine-tuning.
\cite{lim2025language} systematically analyze cross-lingual representation divergence and reveal that language-dependent latent processing undermines cross-lingual consistency and knowledge transfer.
Several works explore inference-time representation interventions to mitigate such disparities without retraining.
\cite{lu-etal-2025-paths} introduce language-agnostic, vector-based interventions that steer internal activations toward more reliable English-centric recall and translation pathways,
while \cite{zhao2025less} explicitly remove language-specific components from hidden states to promote language-invariant reasoning representations.
Beyond dense model interventions, \cite{bandarkar2025multilingual} demonstrate that multilingual performance in MoE LLMs is causally linked to the cross-lingual alignment of middle-layer representations, and improve non-English performance via inference-time interventions on expert routing.
\cite{wang-etal-2025-bridging} utilizes linear alignment matrices from parallel data and applies inference-time hidden-state transformations to map low-resource language representations toward high-resource ones.

Unlike traditional training-based approaches, {\name} intervenes on the model's internal activations during inference, allowing the model to benefit from both English and non-English resources.
We believe this simple yet promising framework marks a new step forward in facilitating cross-lingual interactions and in further tapping into the model's internalized multilingual knowledge.
And distinct from existing representation intervention methods, {\name} is conceived primarily as a probing framework for cross-lingual sharing rather than an alignment mechanism. It explicitly enables English and non-English latent activations to contribute complementary strengths, allowing the model to share linguistic and cultural knowledge across languages and thereby enhance multilingual capability and cultural adaptability.
%%% 8_related_end

%%% 9_conclusion_start
\section{Conclusion}
In this work, we propose and explore {\name}, a cross-lingual latent transplantation framework aiming to further utilize the internalized multilingual knowledge of LLMs at inference time. Our findings reveal that transplanting latent activations across languages enables LLMs to integrate the strengths of both English and non-English resources, enhancing their multilingual capability and cultural adaptability.
Through extensive empirical analysis, we uncover distinct functional roles of attention and feed-forward modules, and conduct an in-depth analysis of {\name}'s stability, effectiveness, and generalizability. By further probing its upper bound performance, we highlight the significant gap between current LLM capabilities and their untapped multilingual potential. Our work sheds light on the underexplored internal mechanisms of LLMs and opens up new possibilities for cross-lingual interaction methods that go beyond traditional input-level prompting or training-based methods. We hope {\name} provides a valuable perspective for advancing the development of more balanced and globally inclusive language models.
%%% 9_conclusion_end

%%% 10_acknowledgements_start
\section*{Acknowledgements}
Xiaocheng Feng is the corresponding author of this work. This work was supported by the National Natural Science Foundation of China (NSFC) (grant 62276078, U22B2059), the Key R\&D Program of Heilongjiang via grant 2022ZX01A32,  and the Fundamental Research Funds for the Central Universities (Grant No.HIT.OCEF.2023018). We also thank Huawei Technologies Co., Ltd for supporting part of the computing resources and funding.
%%% 10_acknowledgements_end

% \begin{thebibliography}{1}
\bibliographystyle{IEEEtran}
\bibliography{bibliography}

\clearpage
{
  
\appendices
\setcounter{page}{1}
\section{Prompts}\label{app:prompt}

The prompts we used for each \textit{PilotSet} are listed in Table~\ref{tab:prompts}.

\begin{table}[h]
  \caption{The Prompts Used For \textit{XNLI}, \textit{XQuAD} And \textit{GlobalOpinionQA}.}
  \small
  \setlength\tabcolsep{3pt}
  \resizebox{0.48\textwidth}{!}{%
      \begin{tabular}{l}
      \toprule
      Prompt for \textit{XNLI} (English version) \\
      \midrule
      Human: What do you think is the relationship between the premise and the hypothesis? \\
      \\
      Premise: \colorbox{given}{\{premise\}} \\
      Hypothesis: \colorbox{given}{\{hypothesis\}} \\
      \\
      (1) Entail \\
      (2) Neutral \\
      (3) Contradict \\
      \\
      Assistant: If I had to select one of the options, my answer would be: 
      \colorbox{response}{\{response\}} \\
      % \bottomrule
      \toprule
      Prompt for \textit{XQuAD} (English version) \\
      \midrule
      Human: Please answer these questions only based on the given context. \\
      \\
      Context: \colorbox{given}{\{context\}} \\
      Question: \colorbox{given}{\{question\}} \\
      \\
      Assistant: My answer would be: \colorbox{response}{\{response\}} \\
      \toprule
      Prompt for \textit{GlobalOpinionQA} (English version) \\
      \midrule
      Human: How would someone from \colorbox{given}{\{country\}} answer the following question: \\
      \colorbox{given}{\{question\}} \\
      Here are the options: \\
      \colorbox{given}{\{options\}} \\
      Assistant: If I had to select one of the options, my answer would be: \colorbox{response}{\{response\}} \\
      \toprule
      \end{tabular}}
  \label{tab:prompts}
\end{table}

\section{Layer-wise Effectiveness Across Languages / Cultures on Mistral and Qwen}\label{app:layer_lang}

The additional layer-wise effectiveness results across languages/cultures on \textit{Mistral-7B-Instruct-v0.3} and \textit{Qwen2-7B-Instruct} are in Figure~\ref{fig:layer_lang_mistral},~\ref{fig:layer_lang_qwen}.

\begin{figure*}[h]
  \centering
  \includegraphics[width=0.95\textwidth]{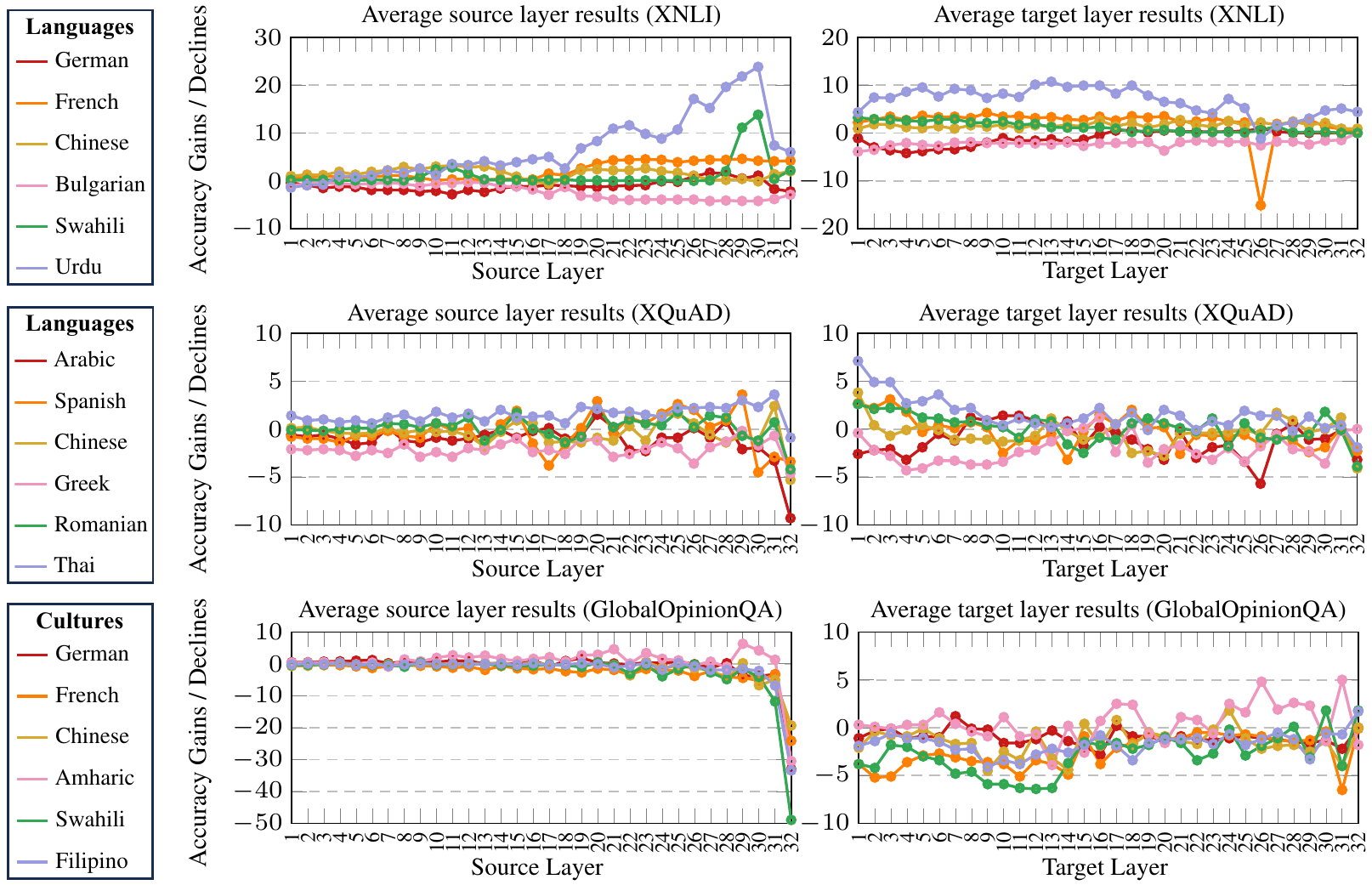}
  \caption{The average layer-wise performance gains or declines of {\name} on \textit{Mistral-7B-Instruct-v0.3}, under different source or target layer configurations.
  }
  \label{fig:layer_lang_mistral}
\end{figure*}

\begin{figure*}[h]
  \centering
  \includegraphics[width=0.95\textwidth]{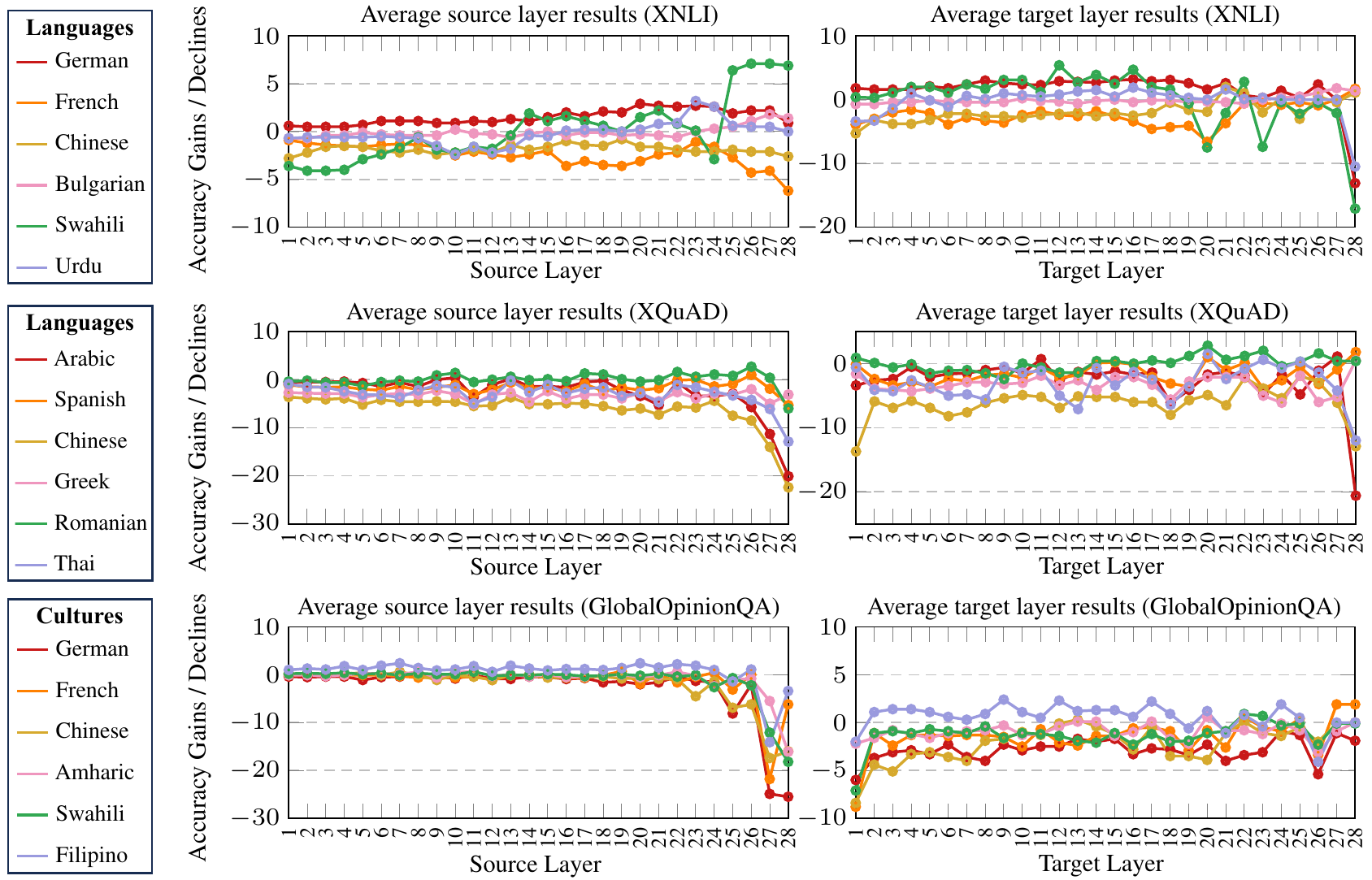}
  \caption{The average layer-wise performance gains or declines of {\name} on \textit{Qwen2-7B-Instruct}, under different source or target layer configurations.
  }
  \label{fig:layer_lang_qwen}
\end{figure*}

\section{Additional Experimental Results}\label{app:addition}

\subsection{Detailed Results Under All $N^2$ Configurations}\label{app:all_config}

The detailed layer-specific effectiveness results under all $N^2$ configurations of {\name} are shown in Figure~\ref{fig:xnli_llama},~\ref{fig:xnli_mistral},~\ref{fig:xnli_qwen},~\ref{fig:xquad_llama},~\ref{fig:xquad_mistral},~\ref{fig:xquad_qwen},~\ref{fig:global_llama},~\ref{fig:global_mistral},~\ref{fig:global_qwen}.

\subsection{Additional Results of Coarse-grained {\name}}\label{app:addition_coarse}

The additional coarse-grained results (Section~\ref{sec:attempts}) on \textit{Mistral-7B-Instruct-v0.3}, \textit{Qwen2-7B-Instruct}, and instruction-tuned \textit{Qwen2-7B-Instruct} are in Table~\ref{tab:attempts_results_mistral},~\ref{tab:attempts_results_qwen},~\ref{tab:attempts_results_qwen_sft}.

Besides, we also conduct additional experiments on more recent benchmark Multilingual MMLU (MMMLU) to validate the effectiveness of {\name} (Attn-level + Last-to-First layer) and the results on presented in Table~\ref{tab:mmmlu}. The results further demonstrate the effectiveness of {\name} on low-resource languages/cultures.

\begin{table*}
  \centering 
  \renewcommand{\arraystretch}{1.1}
  \setlength\tabcolsep{4pt}
  \caption{Performance comparisons between vanilla performance and the performance aftering applying coarse-grained {\name} on \textit{MMMLU}.}
  \resizebox{0.9\textwidth}{!}{%
      \begin{tabular}{lcccccccccccccccccc}
          \toprule
          \multirow{2.5}{*}{\bf {MMMLU}} & \multicolumn{8}{c}{\bf High} &  \multicolumn{5}{c}{\bf Mid} &  \multicolumn{5}{c}{\bf High}\\
          \cmidrule(lr){2-9}
          \cmidrule(lr){10-14}
          \cmidrule(lr){15-19}
           & ar & de & en & es & fr & ja & zh & \bf Avg. & hi & it & ko & pt & \bf Avg. & bn & id & sw & yo & \bf Avg.\\
           \cmidrule(lr){1-9}
          \cmidrule(lr){10-14}
          \cmidrule(lr){15-19}
          \multicolumn{19}{c}{\textbf{\makecell{Llama-2-7b-Chat}}} \\
          \midrule
          Vanilla & 15.8 & 29.0 & 47.6 & 19.7 & 30.3 & 12.2 & 28.1 & \bf 26.1 & 12.7 & 22.5 & 24.1 & 28.6 & \bf 22.0 & 20.5 & 22.0 & 13.8 & 13.6 & \bf 17.5 \\
          Coarse & 16.5 & 29.0 & 47.9 & 19.6 & 29.9 & 11.7 & 27.6 & \bf 26.0 & 17.0 & 22.5 & 24.0 & 28.3 & \bf 23.0 & 20.6 & 24.5 & 18.5 & 15.2 & \bf 19.7 \\
          \midrule
          \multicolumn{19}{c}{\textbf{\makecell{Mistral-7B-Instruct-v0.3}}} \\
          \midrule
          Vanilla & 34.1 & 50.9 & 60.5 & 52.9 & 51.7 & 37.1 & 44.2 & \bf 47.3 & 34.2 & 50.6 & 40.7 & 52.9 & \bf 44.6 & 29.9 & 44.4 & 14.6 & 10.1 & \bf 24.8 \\
          Coarse  & 34.9 & 49.8 & 60.5 & 53.7 & 52.4 & 40.7 & 44.6 & \bf 48.1 & 36.4 & 50.8 & 40.9 & 52.4 & \bf 45.1 & 32.9 & 44.6 & 20.5 & 15.4 &  \bf 28.4 \\
          \midrule
          \multicolumn{19}{c}{\textbf{\makecell{Qwen2-7B-Instruct}}} \\
          \midrule
          Backbone Model & 40.6 & 57.8 & 67.7 & 59.3 & 48.4 & 43.9 & 53.8 & \bf 53.1 & 39.6 & 57.3 & 48.0 & 51.0 & \bf 49.0 & 40.5 & 42.9 & 28.6 & 12.9 & \bf 31.2\\
          Coarse  & 50.7 & 58.4 & 67.2 & 62.1 & 58.7 & 51.6 & 60.3 & \bf 58.4 & 41.9 & 58.8 & 51.7 & 60.7 & \bf 53.3 & 38.0 & 56.2 & 28.8 & 22.3 &  \bf 36.3 \\
          \bottomrule
      \end{tabular}}
      \label{tab:mmmlu}
\end{table*}

\subsection{Additional Perplexity and Input-output Language Consistency Test of Coarse-grained {\name}.}

We include a more targeted result of perplexity and language consistency test under the configuration of coarse-grained {\name} (the response presented in Section V Table III) to further clarify that sub-module cross-layer transplantation does not catastrophically harm the model's general language modeling ability. The results are shown in Table~\ref{tab:modeling_coarse}.

\begin{table*}[ht]
  \centering 
  \renewcommand{\arraystretch}{1.2}
  % \setlength\tabcolsep{2pt}
  % \tabcolsep=0.1cm
  % \vspace{-1.1\baselineskip}
  \caption{The averaged perplexity and input-output language consistency results under the configurations of coarsed-grained {\name} (the responses in Table~\ref{tab:attempts_results_llama}, \ref{tab:attempts_results_mistral}, \ref{tab:attempts_results_qwen}), compared with the results of vanilla model.}
  % \vspace{0.5\baselineskip}
  \resizebox{\textwidth}{!}{%
      \begin{tabular}{lcccccc}
        \toprule
        \multirow{1}{*}{\bf {Perplexity}} & \textbf{XNLI} & \textbf{XQuAD}  & \textbf{GlobalOpinionQA} & \textbf{XCOPA} & \textbf{MKQA} & \textbf{CulturalBench} \\
         % & ($\texttt{non-En}$) & ($\texttt{non-En}$) & ($\texttt{En}$) \\
         \midrule
         LLaMA-2-7B-Chat (w/o {\name}) & 69.31 & 390.49 & 44.44 & 150.12 & 98.89 & 143.69 \\
         LLaMA-2-7B-Chat (coarsed-grained {\name}) & 74.37 & 398.82 & 44.04 & 153.59 & 85.65 & 140.35 \\
         \noalign{\vskip 0.2ex}\cdashline{1-7}\noalign{\vskip 0.4ex}
         Mistral-7B-Instruct-v0.3 (w/o {\name}) & 48.92 & 183.93 & 35.44 & 49.13 & 32.40 & 107.57 \\
         Mistral-7B-Instruct-v0.3 (coarsed-grained {\name}) & 50.95 & 180.28 & 35.42 & 51.27 & 32.25 & 114.84\\
         \noalign{\vskip 0.2ex}\cdashline{1-7}\noalign{\vskip 0.4ex}
         Qwen2-7B-Instruct (w/o {\name}) &  58.89 & 77.11 & 42.74 & 81.86 & 25.30 & 51.28 \\
         Qwen2-7B-Instruct (coarsed-grained {\name}) & 54.60 & 61.23 & 42.60 & 83.70 & 29.11 & 58.91 \\
          % \bottomrule
          \toprule
          \multirow{1}{*}{\bf {Language}} & \textbf{XNLI} & \textbf{XQuAD} & \textbf{GlobalOpinionQA} & \textbf{XCOPA} & \textbf{MKQA} & \textbf{CulturalBench} \\
          \multirow{1}{*}{\bf Consistency} (\%) & ($\texttt{non-En}$) & ($\texttt{non-En}$) & ($\texttt{En}$) & ($\texttt{non-En}$) & ($\texttt{non-En}$) & ($\texttt{En}$) \\
          \midrule
          LLaMA-2-7B-Chat (w/o {\name}) & {95.55} & {89.80} & {99.90} & 85.62 & 72.71 & 95.33\\
          LLaMA-2-7B-Chat (coarsed-grained {\name}) & 93.44 & 89.56 & 99.93 & 83.36 & 72.63 & 95.14 \\
          \noalign{\vskip 0.2ex}\cdashline{1-7}\noalign{\vskip 0.4ex}
          Mistral-7B-Instruct-v0.3 (w/o {\name}) & 94.16 & 96.87 & 99.97 & 83.73 & 91.82 & 98.50 \\
          Mistral-7B-Instruct-v0.3 (coarsed-grained {\name}) & 92.12 & 96.97 & 99.97 & 85.93 & 91.58 & 97.94 \\
          \noalign{\vskip 0.2ex}\cdashline{1-7}\noalign{\vskip 0.4ex}
          Qwen2-7B-Instruct (w/o {\name}) & 99.35 & 99.27 &  100.0 & 88.74 & 97.86 & 98.69 \\
          Qwen2-7B-Instruct (coarsed-grained {\name}) & 99.51 & 99.24 & 100.0 & 83.84 & 95.58 & 97.44 \\
          \bottomrule
      \end{tabular}}
      \label{tab:modeling_coarse}
  % \vspace{-3\baselineskip}
\end{table*}

\subsection{Explanation of accuracy in English subset of \textit{XCOPA} for \textit{Qwen2-7B-Instruct}}\label{app:qwen_instruct}

In Table~\ref{tab:attempts_results_qwen}, we notice that the accuracy in the English subset of XCOPA for \textit{Qwen2-7B-Instruct} is ``0.00''.
After specifically revisiting \textit{Qwen2-7B-Instruct}'s responses to the English subset of XCOPA.
We found that the ``0.00 accuracy'' issue stemms from the model's failure to effectively follow the instructions in our prompt. The exact prompt we used was:
\begin{promptbox}

You are assigned to complete a two-category classification task.\\

Premise: The girl squeezed her nose. \\
Options: (1) The baby drools on the bib. \\
(2) The baby soiled his diaper. \\

Please determine which of the two options is more likely to be the cause of the given premise. \\

Your Answer:
\end{promptbox}
However, \textit{Qwen2-7B-Instruct}'s responses are as follows:
\begin{promptbox}

 Option 1 (The baby drools on the bib) is less likely to be the cause of ...

 Option 1, ``The audience clapped their hands to the music,'' is more likely to be ...

 Option 1 is more likely to be the result of the given premise. If the man expected the ...

 Option 2, ``Her opponent felt sorry for her,'' is more likely to be the result of ...

 Option 2, The products are made by child labor. \textbackslash n\textbackslash n Explanation: The premise states that radicals ...

 Option 2, ``It's snack time,'' is more likely to be the cause of the given ...

...
\end{promptbox}

Our evaluation script for \textit{XCOPA} dataset considers a model's response correct only if it contains the correct option (e.g., (1) or (2)) and excludes all other options. But as you can see above, Qwen-2's responses do not match this format, leading to the ``0.0 accuracy''.

To ensure fairness in evaluation, we can not arbitrarily modify our evaluation script based solely on Qwen's responses on the English subset of the \textit{XCOPA} dataset. Therefore, we have retained this result in our experimental table.

We hypothesize that this instability may account for the limited effectiveness of coarse-grained {\name} on \textit{Qwen2-7B-Instruct} observed in Table~XIII.
To verify this, we conduct additional instruction-tuning (implementations are in Supplementary Materials~D) to enhance it's instruction-following ability and re-evaluate coarse-grained {\name} on OOD data. As shown by the $\Delta$ values in Table~XII, \textit{Qwen2-7B-Instruct} with further tuning can significantly benefit more from cross-lingual transplantation under the {\name} framework. This further indicates the effectiveness of {\name} is closely tied to the model's fundamental instruction-following capability.

\begin{table*}[ht]
  \centering
  \scriptsize
  \renewcommand{\arraystretch}{1} % 设置行间距为默认的 1.5 倍
  \setlength{\tabcolsep}{1pt}
  \setlength{\dashlinedash}{3pt} % 设置虚线段长度
  \setlength{\dashlinegap}{1pt}  % 设置虚线段之间的间隔
  \caption{Additional performance comparisons between the performance of \textit{Qwen2-7B-Instruct} after further instruction-tuning and the performance aftering applying coarse-grained {\name} on instruction-tuned \textit{Qwen2-7B-Instruct}.}
  \begin{tabularx}{\textwidth}{>{\centering\arraybackslash}p{1.4cm}*{7}{>{\centering\arraybackslash}X}*{5}{>{\centering\arraybackslash}X}*{6}{>{\centering\arraybackslash}X}}
  \toprule
  {\bf {Datasets}} & \multicolumn{12}{c}{\bf Unseen Out-of-distribution Data} \\
  % \cmidrule(lr){2-19}
  \midrule
  \multirow{3}{*}{\bf \shortstack{XCOPA}} & \multicolumn{3}{c}{\bf High} & \multicolumn{4}{c}{\bf Mid} & \multicolumn{5}{c}{\bf Low} \\
  \cmidrule(lr){2-4}
  \cmidrule(lr){5-8}
  \cmidrule(lr){9-13}
   & en & zh & \textbf{Avg.} & it & tr & vi & \textbf{Avg.} & ht & id & sw & ta & \textbf{Avg.} \\
  \cmidrule(lr){1-4}
  \cmidrule(lr){5-8}
  \cmidrule(lr){9-13}
  Vanilla & 0.0 & 77.3 & 38.7 & 87.3 & 24.9 & 82.4 & 64.9 & 44.2 & 88.2 & 48.0 & 35.8 & 54.1 \\
  \noalign{\vskip 0.2ex}
  Vanilla+Coarse & 0.0 & 34.9 & 17.4 & 20.4 & 18.7 & 50.7 & 29.9 & 4.4 & 40.7 & 16.9 & 20.0 & 20.5 \\
  \noalign{\vskip 0.2ex}
  $\Delta$ & & & -21.3 &  &  &  & -35.0  &  &  &  &  & -33.6 \\
  \cmidrule(lr){1-4}
  \cmidrule(lr){5-8}
  \cmidrule(lr){9-13}
  Tuning & 87.6 & 75.3 & 81.4 & 90.2 & 29.1 & 71.3 & 63.6 & 38.9 & 84.0 & 38.0 & 38.7 & 49.9 \\
  \noalign{\vskip 0.2ex}
  Tuning+Coarse & 86.7 & 80.2 & 83.4 & 86.7 & 50.0 & 73.8 & 70.1 & 46.4 & 76.2 & 49.1 & 40.2 & 53.0 \\
  \noalign{\vskip 0.2ex}
  $\Delta$ & & & +2.0 &  &  &  & +6.5  &  &  &  &  & +3.1 \\
  % \noalign{\vskip 0.2ex} \hline \noalign{\vskip 0.2ex}
  \end{tabularx}
  
  \begin{tabularx}{\textwidth}{>{\centering\arraybackslash}p{1.4cm}*{8}{>{\centering\arraybackslash}X}*{6}{>{\centering\arraybackslash}X}*{6}{>{\centering\arraybackslash}X}}
  \toprule
  % \cmidrule(lr){2-19}
  \multirow{3}{*}{\bf \shortstack{MKQA}} & \multicolumn{8}{c}{\bf High} & \multicolumn{6}{c}{\bf Mid} & \multicolumn{6}{c}{\bf Low} \\
  \cmidrule(lr){2-9}
  \cmidrule(lr){10-15}
  \cmidrule(lr){16-21}
   & ar & de & en & es & fr & ja & zh & \textbf{Avg.} & it & ko & pt & tr & vi & \textbf{Avg.} & fi & he & hu & nl & no & \textbf{Avg.} \\
  % \cmidrule(lr){1-1}
  \cmidrule(lr){1-9}
  \cmidrule(lr){10-15}
  \cmidrule(lr){16-21}
  Vanilla & 4.3 & 17.7 & 27.9 & 17.7 & 16.4 & 10.8 & 15.1 & 15.7 & 16.2 & 6 & 16 & 16.6 & 17.2 & 14.4 & 10.8 & 3.5 & 10 & 18.1 & 18.3 & 12.1 \\
  \noalign{\vskip 0.2ex}
  Vanilla+Coarse & 4.1 & 19 & 25.1 & 19 & 11.5 & 9.9 & 13.9 & 14.6 & 12.3 & 6.8 & 11.2 & 16.1 & 14.3 & 12.1 & 8.4 & 2.7 & 4.7 & 18.9 & 7.4 & 8.4 \\
  \noalign{\vskip 0.2ex}
  $\Delta$ & & & & & & & & -1.1 & & & & & & -2.3  & & & & & & -3.7 \\
  \cmidrule(lr){1-9}
  \cmidrule(lr){10-15}
  \cmidrule(lr){16-21}
  Tuning & 3.4 & 13.8 & 28.0 & 13.3 & 7.7 & 10.1 & 14.4 & 13.0 & 8.1 & 8.6 & 17.9 & 16.4 & 16.9 & 13.6 & 7.0 & 4.8 & 4.8 & 22.8 & 17.2 & 11.3 \\
  \noalign{\vskip 0.2ex}
  Tuning+Coarse & 5.7 & 17.5 & 29.1 & 17.8 & 18.4 & 8.9 & 14.9 & 16.0 & 15.9 & 8.2 & 17.8 & 16.9 & 17.8 & 15.3 & 12.8 & 6.2 & 11.9 & 21.0 & 16.9 & 13.8 \\
  \noalign{\vskip 0.2ex}
  $\Delta$ & & & & & & & & +3.0 & & & & & & +1.7   & & & & & & +2.5  \\
  % \noalign{\vskip 0.2ex} \hline \noalign{\vskip 0.2ex}
  \end{tabularx}
  
  \begin{tabularx}{\textwidth}{>{\centering\arraybackslash}p{1.4cm}*{8}{>{\centering\arraybackslash}X}*{6}{>{\centering\arraybackslash}X}*{5}{>{\centering\arraybackslash}X}}
  \toprule
  \multirow{3}{*}{\bf \shortstack{Cultural\\Bench}} & \multicolumn{8}{c}{\bf High} & \multicolumn{6}{c}{\bf Mid} & \multicolumn{5}{c}{\bf Low} \\
  \cmidrule(lr){2-9}
  \cmidrule(lr){10-15}
  \cmidrule(lr){16-20}
   & ar & de & en & es & fr & ja & zh & \textbf{Avg.} & hi & ko & ru & tr & vi & \textbf{Avg.} & he & ms & pl & tl & \textbf{Avg.} \\
  \cmidrule(lr){1-9}
  \cmidrule(lr){10-15}
  \cmidrule(lr){16-20}
  Vanilla & 64.7 & 59.4 & 86.7 & 77.5 & 78.6 & 67.9 & 69.5 & 72.0 & 69.6 & 48.8 & 70.0 & 76.3 & 74.1 & 67.7 & 61.5 & 63.6 & 70.8 & 68.9 & 66.2 \\
  \noalign{\vskip 0.2ex}
  Vanilla+Coarse & 47.1 & 53.1 & 68.9 & 62.5 & 57.1 & 54.7 & 57.6 & 57.3 & 54.3 & 43.9 & 50.0 & 55.3 & 66.7 & 54.0 & 61.5 & 63.6 & 66.7 & 60.0 & 63.0 \\
  \noalign{\vskip 0.2ex}
  $\Delta$ & & & & & & & & -14.7 & & & & & & -13.7 & & & & & -3.2 \\
  \cmidrule(lr){1-9}
  \cmidrule(lr){10-15}
  \cmidrule(lr){16-20}
  Tuning & 58.8 & 62.5 & 82.2 & 75.0 & 57.1 & 69.8 & 69.5 & 67.9 & 63.0 & 58.5 & 60.0 & 73.7 & 70.4 & 65.1 & 76.9 & 72.7 & 66.7 & 62.2 & 69.6 \\
  \noalign{\vskip 0.2ex}
  Tuning+Coarse & 58.8 & 59.4 & 82.2 & 75.0 & 57.1 & 69.8 & 69.5 & 67.4 & 63.0 & 58.5 & 63.3 & 73.7 & 70.4 & 65.8 & 76.9 & 72.7 & 66.7 & 64.4 & 70.2 \\
  \noalign{\vskip 0.2ex}
  $\Delta$ & & & & & & & & -0.5  & & & & & & +0.7  & & & & & +0.6  \\
  \bottomrule
  \end{tabularx}
  \label{tab:attempts_results_qwen_sft}
\end{table*}

\begin{table*}[ht]
  \centering
  \scriptsize
  \renewcommand{\arraystretch}{0.8} % 设置行间距为默认的 1.5 倍
  \setlength{\tabcolsep}{1pt}
  \setlength{\dashlinedash}{3pt} % 设置虚线段长度
  \setlength{\dashlinegap}{1pt}  % 设置虚线段之间的间隔
  \caption{Performance comparisons between vanilla performance and the performance aftering applying coarse-grained {\name} on \textbf{\textit{Mistral-7B-Instruct-v0.3}}.}
  \begin{tabularx}{\textwidth}{>{\centering\arraybackslash}p{1.2cm}*{7}{>{\centering\arraybackslash}X}*{5}{>{\centering\arraybackslash}X}*{5}{>{\centering\arraybackslash}X}}
  \toprule
  \textbf{Datasets} & \multicolumn{17}{c}{\bf Unseen In-Distribution Data} \\
  % \cmidrule(lr){1-19}
  \midrule
  \multirow{3}{*}{\bf \shortstack{XNLI}} & \multicolumn{7}{c}{\bf High} & \multicolumn{5}{c}{\bf Mid} & \multicolumn{5}{c}{\bf Low} \\
  \cmidrule(lr){2-8}
  \cmidrule(lr){9-13}
  \cmidrule(lr){14-18}
   & ar & de & en & es & fr & zh & \textbf{Avg.} & hi & ru & tr & vi & \textbf{Avg.} & bg & el & sw & ur & \textbf{Avg.} \\
  % \cmidrule(lr){1-1}
  % \cmidrule(lr){1-8}
  % \cmidrule(lr){9-13}
  % \cmidrule(lr){14-19}
  \midrule
  Vanilla & 12.5 & 47.9 & 40 & 63.3 & 56.4 & 39.9 & 43.3 & 33.4 & 53.6 & 37.8 & 44.6 & 42.4 & 45 & 40.2 & 1 & 12.2 & 24.6 \\
  \noalign{\vskip 0.2ex}
  Coarse & 7.6 & 44.8 & 44.8 & 62.9 & 57.1 & 38.3 & 42.6 & 37.2 & 51.5 & 40.6 & 45.6 & 43.7 & 45.8 & 45 & 3 & 15.9 & 27.4 \\
  % \noalign{\vskip 0.2ex} \hline \noalign{\vskip 0.2ex}
  \end{tabularx}
  
  \begin{tabularx}{\textwidth}{>{\centering\arraybackslash}p{1.2cm}*{7}{>{\centering\arraybackslash}X}*{4}{>{\centering\arraybackslash}X}*{4}{>{\centering\arraybackslash}X}}
  \toprule
  % \multirow{4.5}{*}{\bf {Models}} & \multicolumn{15}{c}{\bf Dataset: XNLI (Unseen)} \\
  % \cmidrule(lr){2-16}
  \multirow{3}{*}{\bf \shortstack{XQuAD}} & \multicolumn{7}{c}{\bf High} & \multicolumn{4}{c}{\bf Mid} & \multicolumn{4}{c}{\bf Low} \\
  \cmidrule(lr){2-7}
  \cmidrule(lr){8-12}
  \cmidrule(lr){13-16}
   & ar & de & en & es & zh & \textbf{Avg.} & hi & ru & tr & vi & \textbf{Avg.} & el & ro & th & \textbf{Avg.} \\
  % \cmidrule(lr){1-1}
  \cmidrule(lr){1-7}
  \cmidrule(lr){8-12}
  \cmidrule(lr){13-16}
  Vanilla & 28.8 & 49.8 & 73.5 & 49.1 & 46.2 & 49.5 & 26.2 & 39.6 & 32.2 & 36.2 & 33.6 & 10.4 & 51.9 & 18.9 & 27.1 \\
  \noalign{\vskip 0.2ex}
  Coarse & 29.3 & 49.6 & 73.5 & 48.7 & 43.4 & 48.9 & 26.5 & 39.3 & 31.8 & 36.8 & 33.6 & 10.4 & 52.5 & 20.7 & 27.9 \\
  % \noalign{\vskip 0.2ex} \hline \noalign{\vskip 0.2ex}
  \end{tabularx}

  \begin{tabularx}{\textwidth}{>{\centering\arraybackslash}p{1.2cm}*{8}{>{\centering\arraybackslash}X}*{7}{>{\centering\arraybackslash}X}*{8}{>{\centering\arraybackslash}X}}
  \toprule
   % & \multicolumn{23}{c}{\bf Dataset: GlobalOpinionQA (Unseen)} \\
  % \cmidrule(lr){2-24}
  \multirow{3}{*}{\bf \shortstack{Global\\OpinionQA}} & \multicolumn{8}{c}{\bf High} & \multicolumn{7}{c}{\bf Mid} & \multicolumn{8}{c}{\bf Low} \\
  \cmidrule(lr){2-9}
  \cmidrule(lr){10-16}
  \cmidrule(lr){17-24}
   & ar & de & en & es & fr & ja & zh & \textbf{Avg.} & hi & pt & ru & sv & tr & vi & \textbf{Avg.} & am & el & id & sw & tl & uk & ur & \textbf{Avg.} \\
  % \cmidrule(lr){1-1}
  \cmidrule(lr){1-9}
  \cmidrule(lr){10-16}
  \cmidrule(lr){17-24}
  Vanilla & 58.9 & 77.9 & 73.2 & 72.3 & 77.8 & 70.7 & 63.6 & 70.6 & 66.7 & 73.0 & 68.4 & 84.6 & 64.2 & 57.1 & 69.0 & 57.9 & 75.0 & 48.5 & 70.0 & 55.6 & 100.0 & 47.5 & 64.9 \\
  \noalign{\vskip 0.2ex}
  Coarse & 57.9 & 66.7 & 75.0 & 72.3 & 75.9 & 70.7 & 62.1 & 68.7 & 66.7 & 74.2 & 65.8 & 84.6 & 62.7 & 57.1 & 68.5 & 73.2 & 78.7 & 48.5 & 68.6 & 55.6 & 100.0 & 47.5 & 67.4 \\
  % \toprule
  \end{tabularx}

  \begin{tabularx}{\textwidth}{>{\centering\arraybackslash}p{1.2cm}*{7}{>{\centering\arraybackslash}X}*{5}{>{\centering\arraybackslash}X}*{6}{>{\centering\arraybackslash}X}}
  \toprule
  {\bf {Datasets}} & \multicolumn{12}{c}{\bf Unseen Out-of-distribution Data} \\
  % \cmidrule(lr){2-19}
  \midrule
  \multirow{3}{*}{\bf \shortstack{XCOPA}} & \multicolumn{3}{c}{\bf High} & \multicolumn{4}{c}{\bf Mid} & \multicolumn{5}{c}{\bf Low} \\
  \cmidrule(lr){2-4}
  \cmidrule(lr){5-8}
  \cmidrule(lr){9-13}
   & en & zh & \textbf{Avg.} & it & tr & vi & \textbf{Avg.} & ht & id & sw & ta & \textbf{Avg.} \\
  \cmidrule(lr){1-4}
  \cmidrule(lr){5-8}
  \cmidrule(lr){9-13}
  Vanilla & 48.0 & 72.7 & 60.3 & 73.8 & 53.6 & 62.2 & 63.2 & 52.2 & 69.3 & 6.7 & 0.0 & 32.1 \\
  \noalign{\vskip 0.2ex}
  Coarse & 30.0 & 65.3 & 47.7 & 62.4 & 52.9 & 48.0 & 54.4 & 46.2 & 45.3 & 31.3 & 42.0 & 41.2 \\
  % \noalign{\vskip 0.2ex} \hline \noalign{\vskip 0.2ex}
  \end{tabularx}
  
  \begin{tabularx}{\textwidth}{>{\centering\arraybackslash}p{1.2cm}*{8}{>{\centering\arraybackslash}X}*{6}{>{\centering\arraybackslash}X}*{6}{>{\centering\arraybackslash}X}}
  \toprule
  \multirow{3}{*}{\bf \shortstack{MKQA}} & \multicolumn{8}{c}{\bf High} & \multicolumn{6}{c}{\bf Mid} & \multicolumn{6}{c}{\bf Low} \\
  \cmidrule(lr){2-9}
  \cmidrule(lr){10-15}
  \cmidrule(lr){16-21}
   & ar & de & en & es & fr & ja & zh & \textbf{Avg.} & it & ko & pt & tr & vi & \textbf{Avg.} & fi & he & hu & nl & no & \textbf{Avg.} \\
  % \cmidrule(lr){1-1}
  \cmidrule(lr){1-9}
  \cmidrule(lr){10-15}
  \cmidrule(lr){16-21}
  Vanilla & 0.8 & 19.2 & 29.4 & 16.9 & 11.4 & 4.3 & 5.6 & 12.5 & 14.2 & 3.9 & 14.7 & 16.3 & 10.4 & 11.9 & 7.2 & 0.7 & 15.8 & 18.5 & 15.4 & 11.5 \\
  \noalign{\vskip 0.2ex}
  Coarse & 0.9 & 18.8 & 29.4 & 16 & 11.2 & 4.4 & 6.1 & 12.4 & 14.1 & 3.7 & 14.2 & 16.6 & 12.3 & 12.2 & 7.7 & 0.7 & 16.3 & 18 & 16.3 & 11.8 \\
  % \noalign{\vskip 0.2ex} \hline \noalign{\vskip 0.2ex}
  \end{tabularx}
  
  \begin{tabularx}{\textwidth}{>{\centering\arraybackslash}p{1.2cm}*{8}{>{\centering\arraybackslash}X}*{6}{>{\centering\arraybackslash}X}*{5}{>{\centering\arraybackslash}X}}
  \toprule
  \multirow{3}{*}{\bf \shortstack{Cultural\\Bench}} & \multicolumn{8}{c}{\bf High} & \multicolumn{6}{c}{\bf Mid} & \multicolumn{5}{c}{\bf Low} \\
  \cmidrule(lr){2-9}
  \cmidrule(lr){10-15}
  \cmidrule(lr){16-20}
   & ar & de & en & es & fr & ja & zh & \textbf{Avg.} & hi & ko & ru & tr & vi & \textbf{Avg.} & he & ms & pl & tl & \textbf{Avg.} \\
  % \cmidrule(lr){1-1}
  \cmidrule(lr){1-9}
  \cmidrule(lr){10-15}
  \cmidrule(lr){16-20}
  Vanilla & 58.8 & 68.8 & 77.8 & 80.0 & 71.4 & 73.6 & 64.4 & 70.7 & 52.2 & 58.5 & 66.7 & 71.1 & 70.4 & 63.8 & 61.5 & 63.6 & 70.8 & 57.8 & 63.4 \\
  \noalign{\vskip 0.2ex}
  Coarse & 58.8 & 68.8 & 77.8 & 77.5 & 71.4 & 71.7 & 64.4 & 70.1 & 52.2 & 61.0 & 66.7 & 71.1 & 70.4 & 64.2 & 61.5 & 63.6 & 70.8 & 57.8 & 63.4 \\
  \toprule
  \end{tabularx}
  \label{tab:attempts_results_mistral}
  
\end{table*}

\begin{table*}[ht]
    \centering
    \scriptsize
    \renewcommand{\arraystretch}{0.8} % 设置行间距为默认的 1.5 倍
    \setlength{\tabcolsep}{1pt}
    \setlength{\dashlinedash}{3pt} % 设置虚线段长度
    \setlength{\dashlinegap}{1pt}  % 设置虚线段之间的间隔
    \caption{Performance comparisons between vanilla performance and the performance aftering applying coarse-grained {\name} on \textbf{\textit{Qwen2-7B-Instruct}}.}
    \begin{tabularx}{\textwidth}{>{\centering\arraybackslash}p{1.2cm}*{7}{>{\centering\arraybackslash}X}*{5}{>{\centering\arraybackslash}X}*{5}{>{\centering\arraybackslash}X}}
    \toprule
    \textbf{Datasets} & \multicolumn{17}{c}{\bf Unseen In-Distribution Data} \\
    % \cmidrule(lr){1-19}
    \midrule
    \multirow{3}{*}{\bf \shortstack{XNLI}} & \multicolumn{7}{c}{\bf High} & \multicolumn{5}{c}{\bf Mid} & \multicolumn{5}{c}{\bf Low} \\
    \cmidrule(lr){2-8}
    \cmidrule(lr){9-13}
    \cmidrule(lr){14-18}
     & ar & de & en & es & fr & zh & \textbf{Avg.} & hi & ru & tr & vi & \textbf{Avg.} & bg & el & sw & ur & \textbf{Avg.} \\
    % \cmidrule(lr){1-1}
    % \cmidrule(lr){1-8}
    % \cmidrule(lr){9-13}
    % \cmidrule(lr){14-19}
    \midrule
    Vanilla & 48.2 & 59.8 & 83.2 & 65.3 & 64.6 & 66.4 & 64.6 & 47.3 & 60.0 & 48.7 & 56.5 & 53.1 & 50.2 & 49.2 & 25.3 &  40.1 & 41.2 \\
    \noalign{\vskip 0.2ex}
    Coarse & 48.1 & 58.9 & 83.9 & 64.7 & 64.9 & 63.2 & 64.0 & 48.1 & 61.9 & 51.8 & 57.5 & 54.8 & 52.9 & 50.9 & 34.3 & 40.6 & 44.7 \\
    % \noalign{\vskip 0.2ex} \hline \noalign{\vskip 0.2ex}
    \end{tabularx}
    
    \begin{tabularx}{\textwidth}{>{\centering\arraybackslash}p{1.2cm}*{7}{>{\centering\arraybackslash}X}*{4}{>{\centering\arraybackslash}X}*{4}{>{\centering\arraybackslash}X}}
    \toprule
    % \multirow{4.5}{*}{\bf {Models}} & \multicolumn{15}{c}{\bf Dataset: XNLI (Unseen)} \\
    % \cmidrule(lr){2-16}
    \multirow{3}{*}{\bf \shortstack{XQuAD}} & \multicolumn{7}{c}{\bf High} & \multicolumn{4}{c}{\bf Mid} & \multicolumn{4}{c}{\bf Low} \\
    \cmidrule(lr){2-7}
    \cmidrule(lr){8-12}
    \cmidrule(lr){13-16}
     & ar & de & en & es & zh & \textbf{Avg.} & hi & ru & tr & vi & \textbf{Avg.} & el & ro & th & \textbf{Avg.} \\
    % \cmidrule(lr){1-1}
    \cmidrule(lr){1-7}
    \cmidrule(lr){8-12}
    \cmidrule(lr){13-16}
    Vanilla & 49.6 & 46.6 & 76.5 & 50.7 & 77.9 & 60.2 & 17.5 & 35.3 & 41.7 & 60.5 & 38.8 & 11.1 & 40.1 & 31.6 & 27.6 \\
    \noalign{\vskip 0.2ex}
    Coarse & 50.3 & 46.7 & 76.5 & 50.4 & 76.2 & 60.0 & 18.1 & 36.1 & 41.5 & 60.9 & 39.1 & 11.2 & 41.1 & 32.3 & 28.2 \\
    % \noalign{\vskip 0.2ex} \hline \noalign{\vskip 0.2ex}
    \end{tabularx}

    \begin{tabularx}{\textwidth}{>{\centering\arraybackslash}p{1.2cm}*{8}{>{\centering\arraybackslash}X}*{7}{>{\centering\arraybackslash}X}*{8}{>{\centering\arraybackslash}X}}
    \toprule
     % & \multicolumn{23}{c}{\bf Dataset: GlobalOpinionQA (Unseen)} \\
    % \cmidrule(lr){2-24}
    \multirow{3}{*}{\bf \shortstack{Global\\OpinionQA}} & \multicolumn{8}{c}{\bf High} & \multicolumn{7}{c}{\bf Mid} & \multicolumn{8}{c}{\bf Low} \\
    \cmidrule(lr){2-9}
    \cmidrule(lr){10-16}
    \cmidrule(lr){17-24}
     & ar & de & en & es & fr & ja & zh & \textbf{Avg.} & hi & pt & ru & sv & tr & vi & \textbf{Avg.} & am & el & id & sw & tl & uk & ur & \textbf{Avg.} \\
    % \cmidrule(lr){1-1}
    \cmidrule(lr){1-9}
    \cmidrule(lr){10-16}
    \cmidrule(lr){17-24}
    Vanilla & 50.6 & 76.2 & 66.7 & 67.7 & 76.9 & 63.4 & 48.5 & 64.3 & 66.7 & 62.9 & 63.2 & 80.8 & 56.7 & 47.6 & 63.0 & 57.9 & 63.3 & 53.0 & 68.6 & 33.3 & 66.7 & 43.2 & 55.2 \\
    \noalign{\vskip 0.2ex}
    Coarse & 51.1 & 75.4 & 66.7 & 67.9 & 76.9 & 63.4 & 48.5 & 64.3 & 66.7 & 62.9 & 63.2 & 84.6 & 58.2 & 47.6 & 63.9 & 57.9 & 63.3 & 53.0 & 70.0 & 33.3 & 66.7 & 43.2 & 55.4 \\
    % \toprule
    \end{tabularx}

    \begin{tabularx}{\textwidth}{>{\centering\arraybackslash}p{1.2cm}*{7}{>{\centering\arraybackslash}X}*{5}{>{\centering\arraybackslash}X}*{6}{>{\centering\arraybackslash}X}}
    \toprule
    {\bf {Datasets}} & \multicolumn{12}{c}{\bf Unseen Out-of-distribution Data} \\
    % \cmidrule(lr){2-19}
    \midrule
    \multirow{3}{*}{\bf \shortstack{XCOPA}} & \multicolumn{3}{c}{\bf High} & \multicolumn{4}{c}{\bf Mid} & \multicolumn{5}{c}{\bf Low} \\
    \cmidrule(lr){2-4}
    \cmidrule(lr){5-8}
    \cmidrule(lr){9-13}
     & en & zh & \textbf{Avg.} & it & tr & vi & \textbf{Avg.} & ht & id & sw & ta & \textbf{Avg.} \\
    \cmidrule(lr){1-4}
    \cmidrule(lr){5-8}
    \cmidrule(lr){9-13}
    Vanilla & 0.0 & 77.3 & 38.7 & 87.3 & 24.9 & 82.4 & 64.9 & 44.2 & 88.2 & 48.0 & 35.8 & 54.1 \\
    \noalign{\vskip 0.2ex}
    Coarse & 0.0 & 34.9 & 17.4 & 20.4 & 18.7 & 50.7 & 29.9 & 4.4 & 40.7 & 16.9 & 20.0 & 20.5 \\
    % \noalign{\vskip 0.2ex} \hline \noalign{\vskip 0.2ex}
    \end{tabularx}
    
    \begin{tabularx}{\textwidth}{>{\centering\arraybackslash}p{1.2cm}*{8}{>{\centering\arraybackslash}X}*{6}{>{\centering\arraybackslash}X}*{6}{>{\centering\arraybackslash}X}}
    \toprule

    \multirow{3}{*}{\bf \shortstack{MKQA}} & \multicolumn{8}{c}{\bf High} & \multicolumn{6}{c}{\bf Mid} & \multicolumn{6}{c}{\bf Low} \\
    \cmidrule(lr){2-9}
    \cmidrule(lr){10-15}
    \cmidrule(lr){16-21}
     & ar & de & en & es & fr & ja & zh & \textbf{Avg.} & it & ko & pt & tr & vi & \textbf{Avg.} & fi & he & hu & nl & no & \textbf{Avg.} \\
    % \cmidrule(lr){1-1}
    \cmidrule(lr){1-9}
    \cmidrule(lr){10-15}
    \cmidrule(lr){16-21}
    Vanilla & 4.3 & 17.7 & 27.9 & 17.7 & 16.4 & 10.8 & 15.1 & 15.7 & 16.2 & 6 & 16 & 16.6 & 17.2 & 14.4 & 10.8 & 3.5 & 10 & 18.1 & 18.3 & 12.1 \\
    \noalign{\vskip 0.2ex}
    Coarse & 4.1 & 19 & 25.1 & 19 & 11.5 & 9.9 & 13.9 & 14.6 & 12.3 & 6.8 & 11.2 & 16.1 & 14.3 & 12.1 & 8.4 & 2.7 & 4.7 & 18.9 & 7.4 & 8.4 \\
    % \noalign{\vskip 0.2ex} \hline \noalign{\vskip 0.2ex}
    \end{tabularx}
    
    \begin{tabularx}{\textwidth}{>{\centering\arraybackslash}p{1.2cm}*{8}{>{\centering\arraybackslash}X}*{6}{>{\centering\arraybackslash}X}*{5}{>{\centering\arraybackslash}X}}
    \toprule
    \multirow{3}{*}{\bf \shortstack{Cultural\\Bench}} & \multicolumn{8}{c}{\bf High} & \multicolumn{6}{c}{\bf Mid} & \multicolumn{5}{c}{\bf Low} \\
    \cmidrule(lr){2-9}
    \cmidrule(lr){10-15}
    \cmidrule(lr){16-20}
     & ar & de & en & es & fr & ja & zh & \textbf{Avg.} & hi & ko & ru & tr & vi & \textbf{Avg.} & he & ms & pl & tl & \textbf{Avg.} \\
    % \cmidrule(lr){1-1}
    \cmidrule(lr){1-9}
    \cmidrule(lr){10-15}
    \cmidrule(lr){16-20}
    Vanilla & 64.7 & 59.4 & 86.7 & 77.5 & 78.6 & 67.9 & 69.5 & 72.0 & 69.6 & 48.8 & 70.0 & 76.3 & 74.1 & 67.7 & 61.5 & 63.6 & 70.8 & 68.9 & 66.2 \\
    \noalign{\vskip 0.2ex}
    Coarse & 47.1 & 53.1 & 68.9 & 62.5 & 57.1 & 54.7 & 57.6 & 57.3 & 54.3 & 43.9 & 50.0 & 55.3 & 66.7 & 54.0 & 61.5 & 63.6 & 66.7 & 60.0 & 63.0 \\
    \toprule
    \end{tabularx}
    
    \label{tab:attempts_results_qwen}
\end{table*}

\subsection{Additional Results on More Models}\label{app:more_models}

To further validate the effectiveness of {\name} on more models, we conduct additional experiments on \textit{Llama-3.1-8B-Instruct} and \textit{Qwen2.5-7B-Instruct} between vanilla model, CoT and Multilingual SFT on \textit{PilotSets}. The results are shown in Table~\ref{tab:compare_more_models}.
Besides, we also conduct additional experiments on \textit{Qwen3-4B} and \textit{Qwen3-8B} to validate the effectiveness of {\name} on reasoning models. The results are shown in Table~\ref{tab:compare_more_models_qwen3}.

% The results in Table~\ref{tab:compare_more_models} show 和 main paper中类似结论, {\name}在 low-resource 的languegs/cultures上带来的性能提升更明显，有时甚至能超过multilingual-SFT的性能。The results on reasoning models \textit{Qwen3-4B} and \textit{Qwen3-8B} 展示了{\name}在non-Thinking mode 的性能提升会比Thinking mode更明显一些，这说明 {\name} 带来的benefit会随着推理长度的scaling而衰退。且总体而言，相比较而言Qwen3系列模型上的性能提升会比LLaMA-2，LLaMA-3.1，Qwen2，Qwen2.5模型的提升较小一些，这体现了XXX。
The results in Table~\ref{tab:compare_more_models} exhibit trends consistent with those observed in the main paper. Specifically, {\name} yields more pronounced performance improvements on low-resource languages and cultures, in some cases even surpassing multilingual-SFT.
For reasoning-oriented models Qwen3-4B and Qwen3-8B in Table~\ref{tab:compare_more_models_qwen3}, we observe that the gains brought by {\name} are more substantial in the non-thinking mode than in the thinking mode. This suggests that the benefits of {\name} tend to diminish as the length and complexity of reasoning increase.
Moreover, compared to models such as LLaMA-2, LLaMA-3.1, Qwen2, and Qwen2.5, the overall improvements on the Qwen3 series are relatively smaller. This reflects that these newer models exhibit stronger intrinsic multilingual alignment and reasoning capabilities, leaving less room for additional gains through cross-lingual transplantation.

\begin{table}
  \centering 
  \caption{Additional Performance Comparisons on \textit{Llama-3.1-8B-Instruct} and \textit{Qwen2.5-7B-Instruct} Between Vanilla Model, CoT and Multilingual SFT On \textit{PilotSets}.}
  \resizebox{0.48\textwidth}{!}{%
      \begin{tabular}{lccccccccc}
          \toprule
          \multirow{4}{*}{\textbf{Settings}} & \multicolumn{9}{c}{\bf \textit{PilotSets}} \\ 
          \cmidrule(lr){2-10}
          & \multicolumn{3}{c}{\bf XNLI} & \multicolumn{3}{c}{\bf XQuAD} & \multicolumn{3}{c}{\bf GlobalOpinionQA} \\
          \cmidrule(lr){2-4}
          \cmidrule(lr){5-7}
          \cmidrule(lr){8-10}
          & High & Mid & Low & High & Mid & Low & High & Mid & Low \\
          \midrule
          \multicolumn{10}{c}{\textbf{\textit{Llama-3.1-8B-Instruct}}} \\
          \noalign{\vskip 0.2ex}\cdashline{1-10}\noalign{\vskip 0.8ex}
          Vanilla & 35.3 & 40.5 & 33.5 & 64.8 & 64.5 & 52.7 & 74.6 & 73.7 & 71.1 \\
          \midrule
          \multicolumn{10}{c}{ Methods for Comparison} \\
          \midrule
          CoT & 37.7 & 27.5 & 13.0 & 44.0 & 50.0 & 32.7 & 24.6 & 25.3 & 23.7\\
          ML-SFT & 46.3 & 45.0 & 36.5 & 62.0 & 65.5 & 56.7 & 74.0 & 76.0 & 70.3\\
          \midrule
          \multicolumn{10}{c}{ Finer-grained Attempt under {\name} Framework} \\
          \midrule
          Entropy-based & 39.0 & 45.5 & 38.0 & 68.8 & 62.0 & 54.0 & 74.7 & 75.3 & 72.3\\
          \midrule
          \midrule
          
          \multicolumn{10}{c}{\textbf{\textit{Qwen2.5-7B-Instruct}}} \\
          \noalign{\vskip 0.2ex}\cdashline{1-10}\noalign{\vskip 0.8ex}
          Vanilla & 67.3 & 57.5 & 40.0 & 57.6 & 38.0 & 28.0 & 74.0 & 74.0 & 63.4\\
          \midrule
          \multicolumn{10}{c}{Methods for Comparison} \\
          \midrule
          CoT & 52.7 & 33.0 & 13.5 & 32.4 & 13.0 & 16.0 & 49.1 & 54.7 & 41.4 \\
          ML-SFT &  71.7 & 47.5 & 34.0 & 70.0 & 63.5 & 44.7  & 69.1 & 68.0 & 65.1\\
          \midrule
          \multicolumn{10}{c}{Finer-grained Attempt under {\name} Framework} \\
          \midrule
          Entropy-based & 67.7 & 58.0 & 48.0 & 57.8 & 38.5 & 30.0 & 74.9 & 73.7 & 64.4 \\
          \bottomrule
      \end{tabular}}

      \label{tab:compare_more_models}
\end{table}

\begin{table}
  \centering 
  \caption{Additional Performance Comparisons on \textit{Qwen3-4B/8B} Between Vanilla Model and Entropy-based Strategy On \textit{PilotSets} Under Both Thinking and non-Thinking Modes.}
  \resizebox{0.48\textwidth}{!}{%
      \begin{tabular}{lccccccccc}
          \toprule
          \multirow{4}{*}{\textbf{Settings}} & \multicolumn{9}{c}{\bf \textit{PilotSets}} \\ 
          \cmidrule(lr){2-10}
          & \multicolumn{3}{c}{\bf XNLI} & \multicolumn{3}{c}{\bf XQuAD} & \multicolumn{3}{c}{\bf GlobalOpinionQA} \\
          \cmidrule(lr){2-4}
          \cmidrule(lr){5-7}
          \cmidrule(lr){8-10}
          & High & Mid & Low & High & Mid & Low & High & Mid & Low \\
          \midrule
          \multicolumn{10}{c}{\textbf{\textit{Qwen3-4B}}} \\
          \midrule
          Vanilla (non-Thinking Mode) & 57.7 & 53.5 & 43.0 & 72.0 & 74.5 & 68.7 & 73.1 & 74.0 & 71.7 \\
          Entropy-based (non-Thinking Mode) & 60.0 & 54.0 & 45.0 & 72.4 & 74.5 & 70.3 & 73.4 & 74.7 & 72.9 \\
          $\Delta$ & +2.3 & +0.5 & +2.0 & +0.4 & +0.0 & +1.6 & +0.3 & +0.7 & +1.2 \\
          \midrule
          Vanilla (Thinking Mode) & 70.0 & 57.5 & 51.0 & 67.6 & 69.5 & 64.7 & 72.0 & 75.3 & 70.0 \\
          Entropy-based (Thinking Mode) & 70.7 & 58.0 & 51.5 & 67.9 & 70.0 & 64.7  & 71.4 & 75.3 & 71.0 \\
          $\Delta$ & +0.7 & +0.5 & +0.5 & +0.3 & +0.5 & +0.0 & -0.6 & +0.0 & +1.0 \\
          \midrule
          
          \multicolumn{10}{c}{\textbf{\textit{Qwen3-8B}}} \\
          \midrule
          Vanilla (non-Thinking Mode) & 68.0 & 56.5 & 48.0 & 74.4 & 71.0 & 69.3 & 72.6 & 76.7 & 67.4 \\
          Entropy-based (non-Thinking Mode) & 68.0 & 58.0 & 49.0 & 74.4 & 72.5 & 69.3 & 73.4 & 77.3 & 68.3 \\
          $\Delta$ & +0.0 & +1.5 & +1.0 & +0.0 & +1.5 & +0.0 & +0.8 & +0.6 & +0.9 \\
          \midrule
          Vanilla (Thinking Mode) & 72.0 & 61.5 & 53.0 &  74.0 & 72 & 67.3 & 71.4 & 74.3 & 70.0 \\
          Entropy-based (Thinking Mode) & 72.0 & 61.0 & 56.5 &  74.0 & 73.5 & 68.0 & 72.3 & 73.7 & 70.5 \\
          $\Delta$ & +0.0 & -0.5 & +3.5 & +0.0 & +1.5 & +0.7 & +0.9 & +0.6 & +0.5 \\
          \bottomrule
      \end{tabular}}

      \label{tab:compare_more_models_qwen3}
\end{table}

\subsection{Additional Responses Distribution Analysis on Context-Augmentation Baseline}\label{app:context_augmentation}

To further validate the nature of {\name}, we compare the distribution of post-transplant responses between coarse-grained {\name} and a Context-Augmentation baseline. The Context-Augmentation baseline is a simple method that directly inserts the source-language response in front of the input prompt of the target language context in natural language. The results are shown in Table~\ref{tab:context_augmentation_distribution}.

\begin{table}
  \centering 
  \renewcommand{\arraystretch}{1.2}
  \caption{Comparisions of Distribution of Post-Transplant Responses Between Coarse-grained {\name} and Context-Augmentation baseline.}
  \resizebox{0.48\textwidth}{!}{%
      \begin{tabular}{lccc}
        \toprule
        {Dataset: XNLI} & Target-preserved & Source-aligned & Divergent \\
        \midrule
        \multicolumn{4}{c}{\textit{(Source: English, Target: Non-English)}} \\
        \midrule
        Llama-2-7b-chat \\
        \quad + Coarsed-grained {\name} & 45.49\% & 15.93\% & 38.58\% \\
        \quad + Context-Augmentation & 66.57\% & 30.31\% & 3.12\% \textcolor{red}{$\bm{\downarrow}$} \\
        \noalign{\vskip 0.2ex}\cdashline{1-4}\noalign{\vskip 0.4ex}
        Mistral-7B-Instruct-v0.3 \\
        \quad + Coarsed-grained {\name} & 73.07\% & 12.91\% & 14.02\%  \\
        \quad + Context-Augmentation & 66.23\% & 29.56\% & 4.21\% \textcolor{red}{$\bm{\downarrow}$} \\
        \noalign{\vskip 0.2ex}\cdashline{1-4}\noalign{\vskip 0.4ex}
        Qwen2-7B-Instruct \\
        \quad + Coarsed-grained {\name} & 90.77\% & 4.19\% & 5.04\%  \\
        \quad + Context-Augmentation & 42.65\% & 56.75\% & 0.60\% \textcolor{red}{$\bm{\downarrow}$} \\
        \midrule
        {Dataset: GlobalOpinionQA} & Target-preserved & Source-aligned & Divergent \\
        \midrule
        \multicolumn{4}{c}{\textit{(Source: Non-English, Target: English)}} \\
        \midrule
        Llama-2-7b-chat  \\
        \quad + Coarsed-grained {\name} & 94.31\% & 2.49\% & 3.20\%  \\
        \quad + Context-Augmentation & 39.75\% & 57.90\% & 2.35\% \textcolor{red}{$\bm{\downarrow}$}\\
        \noalign{\vskip 0.2ex}\cdashline{1-4}\noalign{\vskip 0.4ex}
        Mistral-7B-Instruct-v0.3  \\
        \quad + Coarsed-grained {\name} & 96.05\% & 0.77\% & 3.18\%  \\
        \quad + Context-Augmentation & 62.41\% & 35.47\% & 2.12\% \textcolor{red}{$\bm{\downarrow}$} \\
        \noalign{\vskip 0.2ex}\cdashline{1-4}\noalign{\vskip 0.4ex}
        Qwen2-7B-Instruct  \\
        \quad + Coarsed-grained {\name} & 97.14\% & 1.34\% & 1.52\%  \\
        \quad + Context-Augmentation & 75.14\% & 23.90\% & 0.96\% \textcolor{red}{$\bm{\downarrow}$} \\
        \bottomrule
      \end{tabular}}
      \label{tab:context_augmentation_distribution}
\end{table}

Consistent with the observations in \S\ref{sec:post}, {\name} exhibits a relatively low proportion of Source-aligned responses. In contrast, the Context-Augmentation baseline substantially increases the proportion of Source-aligned responses on both XNLI and GlobalOpinionQA, which more closely matches the answer bypass scenario, where explicit source-language outputs are directly exposed to the model.
Moreover, on both XNLI and GlobalOpinionQA, the Context-Augmentation baseline largely fails to produce Divergent responses (especially on XNLI), suggesting that it does not enable the kind of soft decision fusion like {\name} that allows the model to arrive at its own distinct decisions. Instead, its predictions are almost entirely constrained to either the original source-language or target-language answers. This behavior contrasts sharply with {\name}, which consistently yields a non-negligible Divergent proportion, reflecting the emergence of new decisions that are not trivially inherited from either side.

\section{Method Implementations}\label{app:method}

\subsection{Chain-of-Thought (CoT) Prompting.}
CoT prompts the models with the suffix of ``Let's think step by step'' (in corresponding languages) to utilize their further potential. Under the CoT setting, the max new token of models' responses is set to 200 to ensure sufficient space for step-by-step reasoning.

\subsection{Multilingual Supervised Fine-tuning (ML-SFT).}\label{app:sft}
We totally select 20,236 multilingual instruction pairs from \textit{aya dataset} as our training corpus and the multilingual training corpus covers more than 60 languages, ensuring extensive multilingual coverage. Our training processes are conducted on \textit{8 * A800-SXM4-80GB} with the following settings: \textit{batch size=16}, \textit{epochs=3}, \textit{learning rate=1.0e-5}, \textit{warmup ratio=0.1}, and \textit{bf16=true}.

}

\section{Multi-step {\name} Attempts on Long-form Answering}\label{app:cases_multi_step}

\begin{figure*}[ht]
  \begin{center}
  \includegraphics[width=0.9\textwidth]{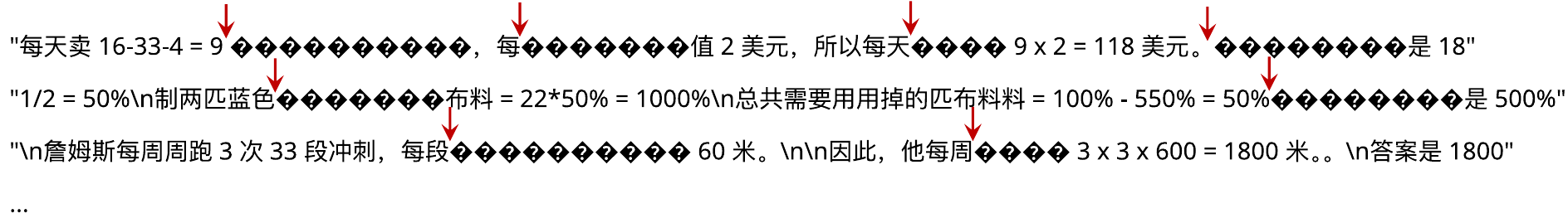}
  \end{center}
  \caption{Cases when multi-step {\name} deteriorates performance significantly on Chinese Math Reasoning; the \textcolor{red}{$\bm{\downarrow}$} in the figure denotes the intervention step; the results are produced by \textit{Llama-2-7b-Chat}.}
  \label{fig:cases_multi_step}
\end{figure*}

\begin{figure}[h]
  \centering
  \includegraphics[width=0.5\textwidth]{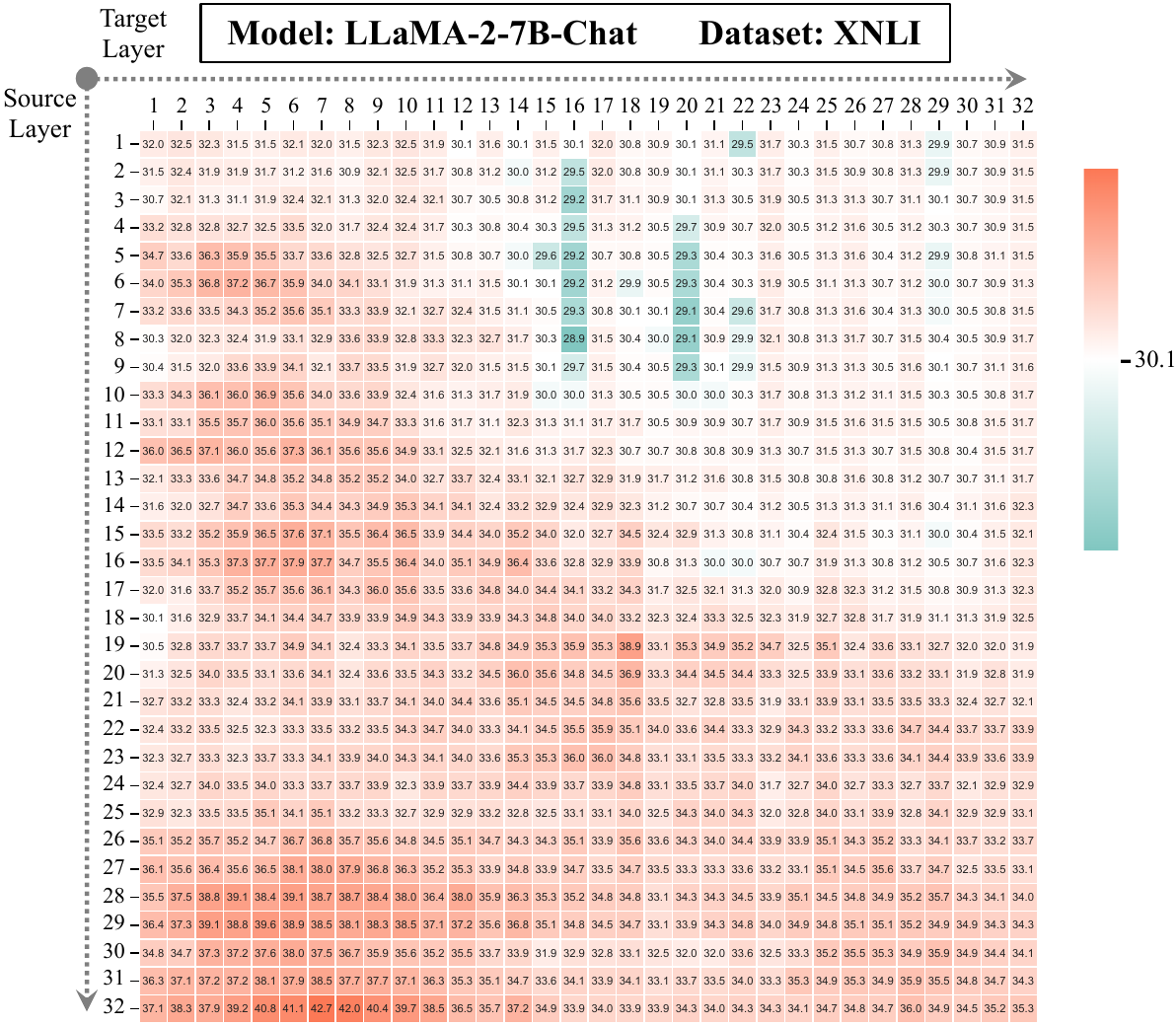}
  \caption{Supplementary layer-specific effectiveness results of \textit{Llama-2-7b-Chat} on XNLI dataset. The colorbar median value represents the accuracy of backbone model.
  }
  \label{fig:xnli_llama}
\end{figure}

\begin{figure}[h]
  \centering
  \includegraphics[width=0.5\textwidth]{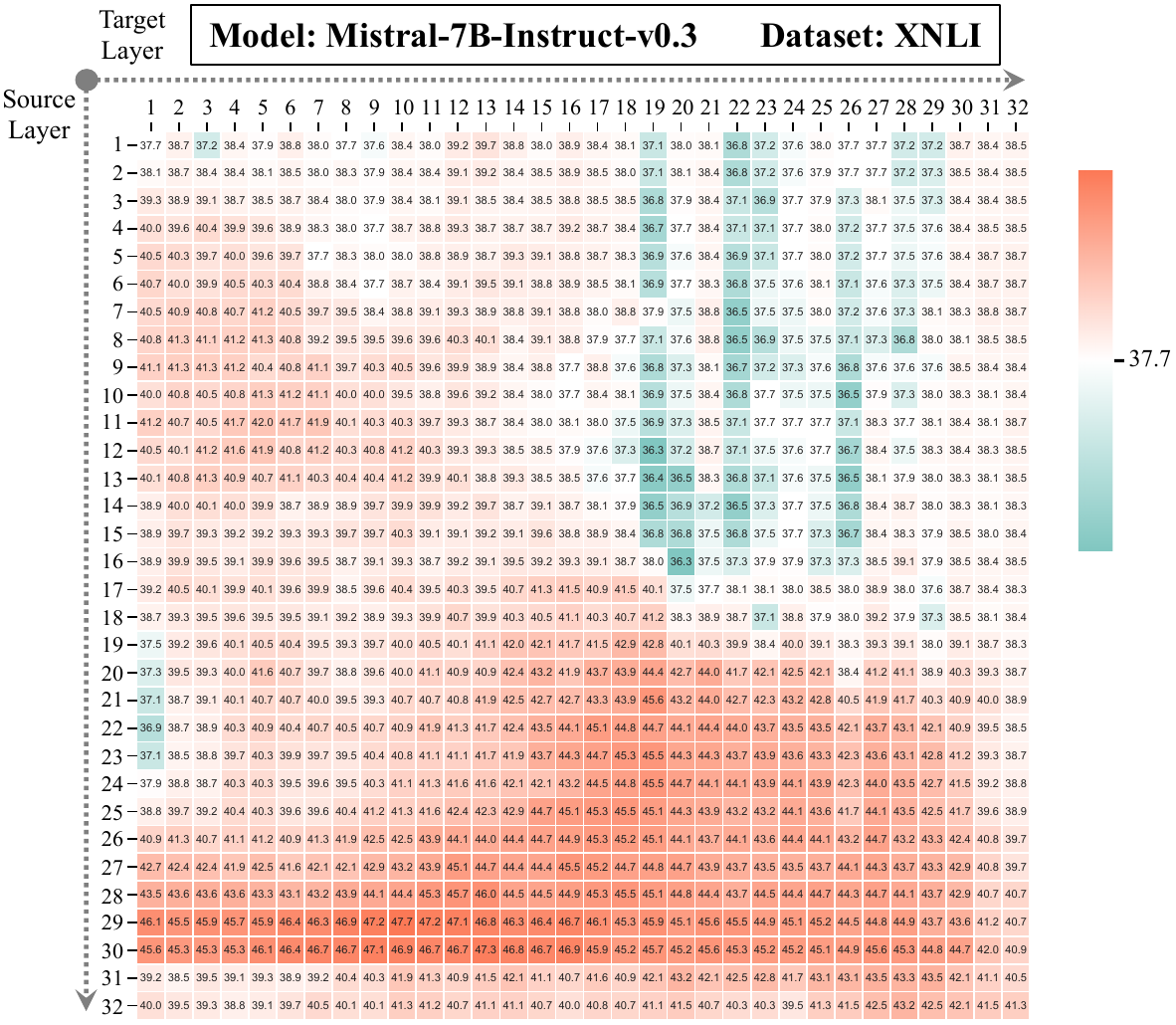}
  \caption{Supplementary layer-specific effectiveness results of \textit{Mistral-7B-Instruct-v0.3} on XNLI dataset. The colorbar median value represents the accuracy of backbone model.
  }
  \label{fig:xnli_mistral}
\end{figure}

\begin{figure}[h]
  \centering
  \includegraphics[width=0.5\textwidth]{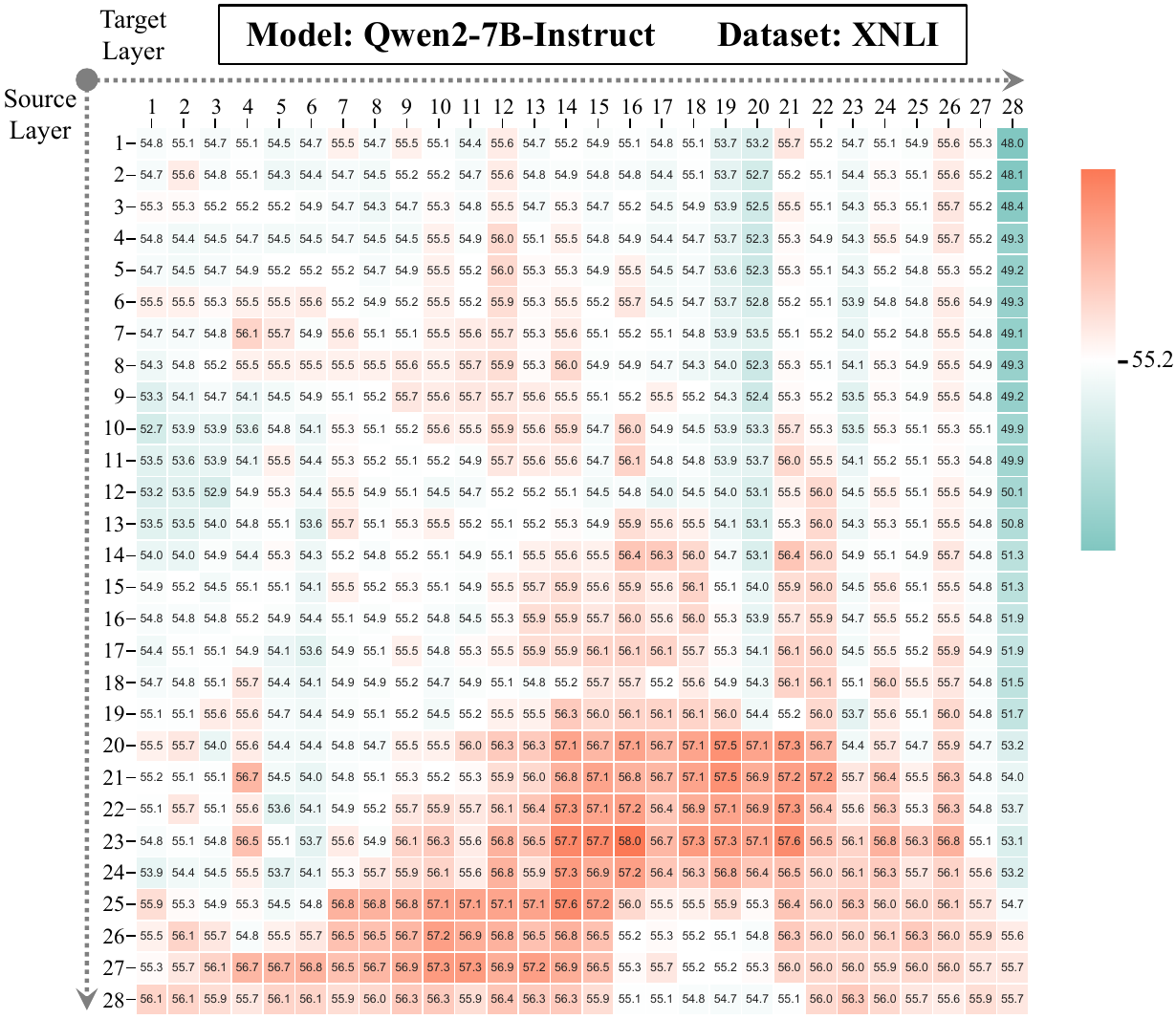}
  \caption{Supplementary layer-specific effectiveness results of \textit{Qwen2-7B-Instruct} on XNLI dataset. The colorbar median value represents the accuracy of backbone model.
  }
  \label{fig:xnli_qwen}
\end{figure}

\begin{figure}[h]
  \centering
  \includegraphics[width=0.5\textwidth]{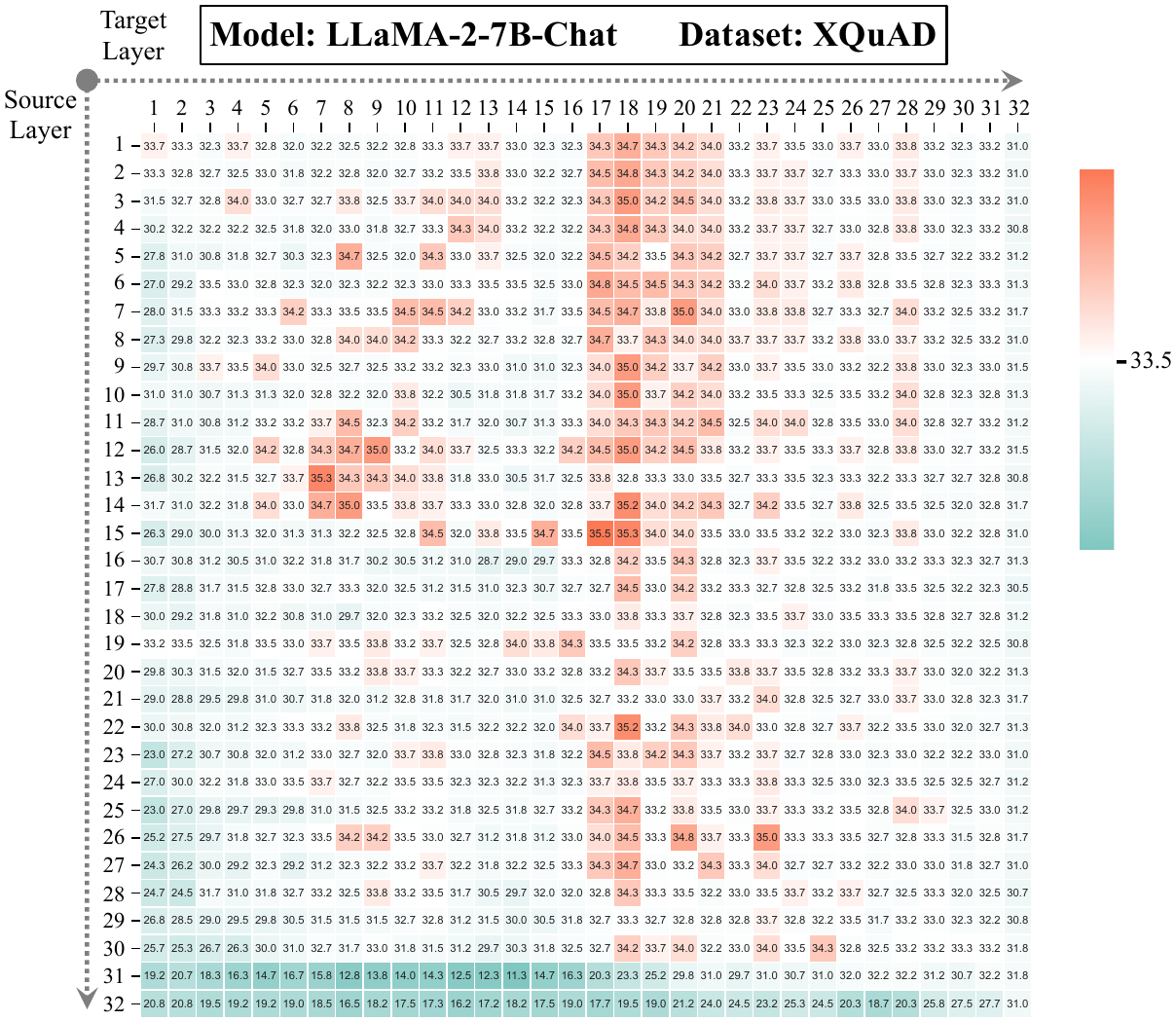}
  \caption{Supplementary layer-specific effectiveness results of \textit{Llama-2-7b-Chat} on XQuAD dataset. The colorbar median value represents the accuracy of backbone model.
  }
  \label{fig:xquad_llama}
\end{figure}

\begin{figure}[h]
  \centering
  \includegraphics[width=0.5\textwidth]{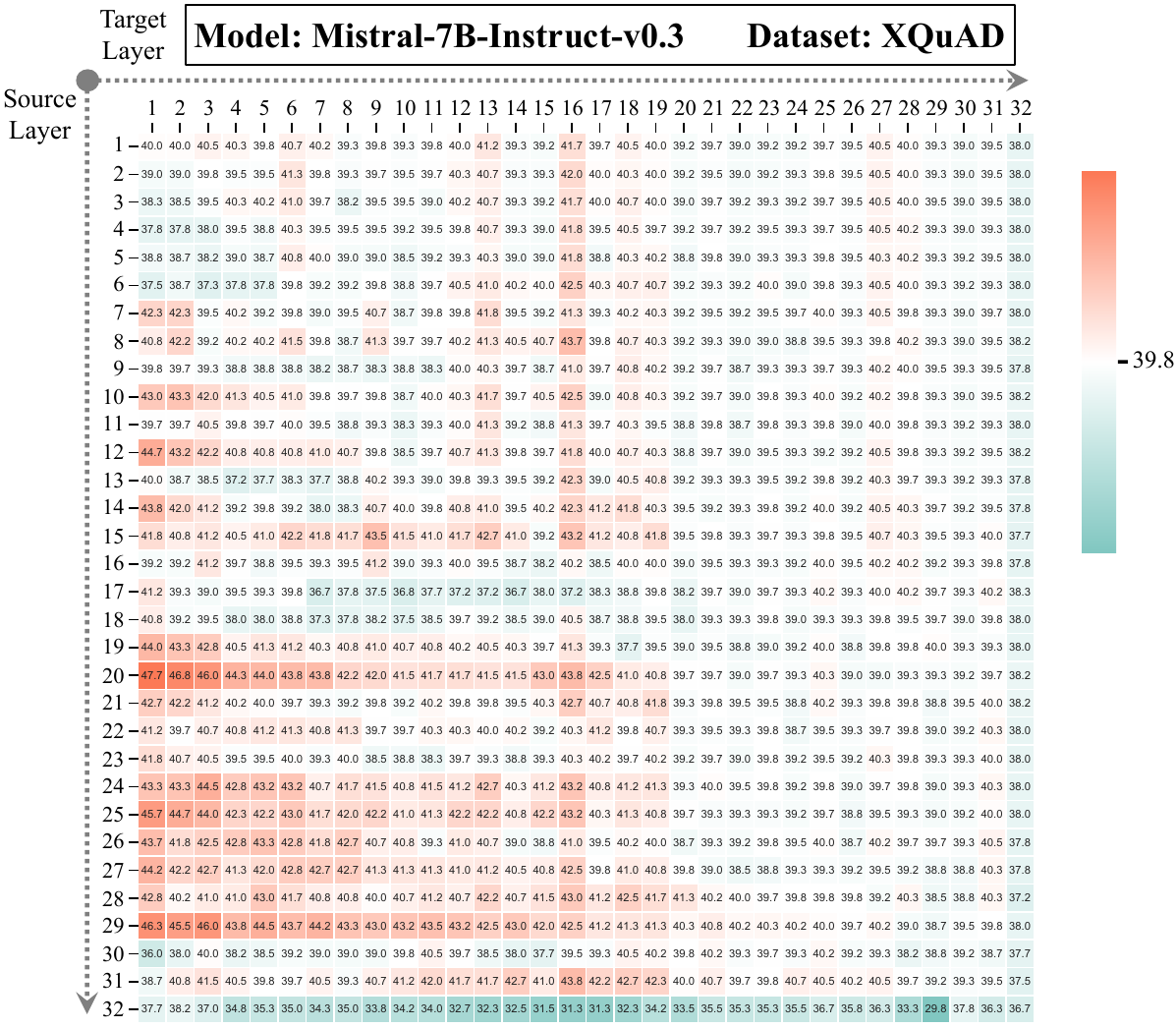}
  \caption{Supplementary layer-specific effectiveness results of \textit{Mistral-7B-Instruct-v0.3} on XQuAD dataset. The colorbar median value represents the accuracy of backbone model.
  }
  \label{fig:xquad_mistral}
\end{figure}

\begin{figure}[h]
  \centering
  \includegraphics[width=0.5\textwidth]{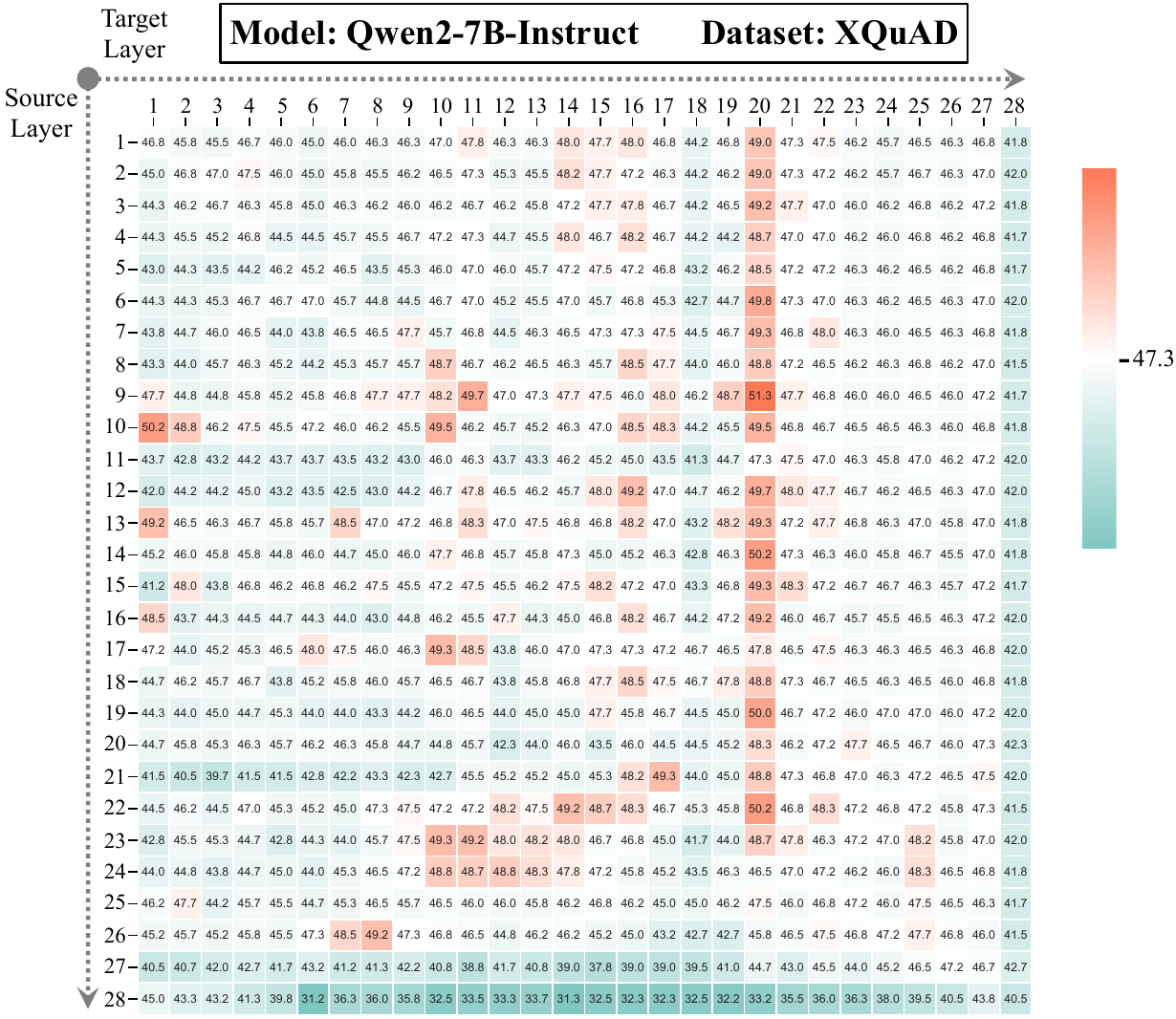}
  \caption{Supplementary layer-specific effectiveness results of \textit{Qwen2-7B-Instruct} on XQuAD dataset. The colorbar median value represents the accuracy of backbone model.
  }
  \label{fig:xquad_qwen}
\end{figure}

\begin{figure}[h]
  \centering
  \includegraphics[width=0.5\textwidth]{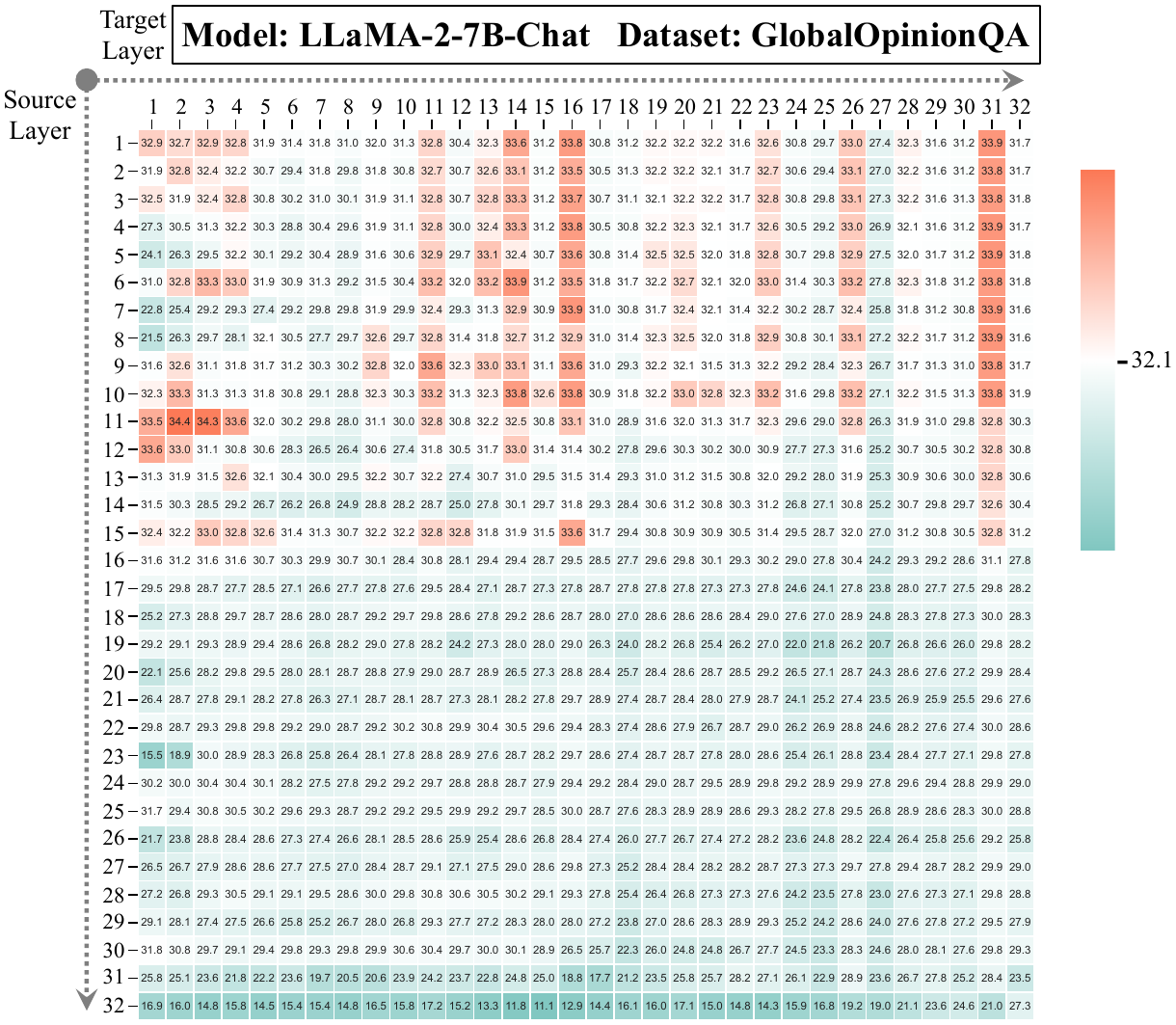}
  \caption{Supplementary layer-specific effectiveness results of \textit{Llama-2-7b-Chat} on GlobalOpinionQA dataset. The colorbar median value represents the accuracy of backbone model.
  }
  \label{fig:global_llama}
\end{figure}

\begin{figure}[h]
  \centering
  \includegraphics[width=0.5\textwidth]{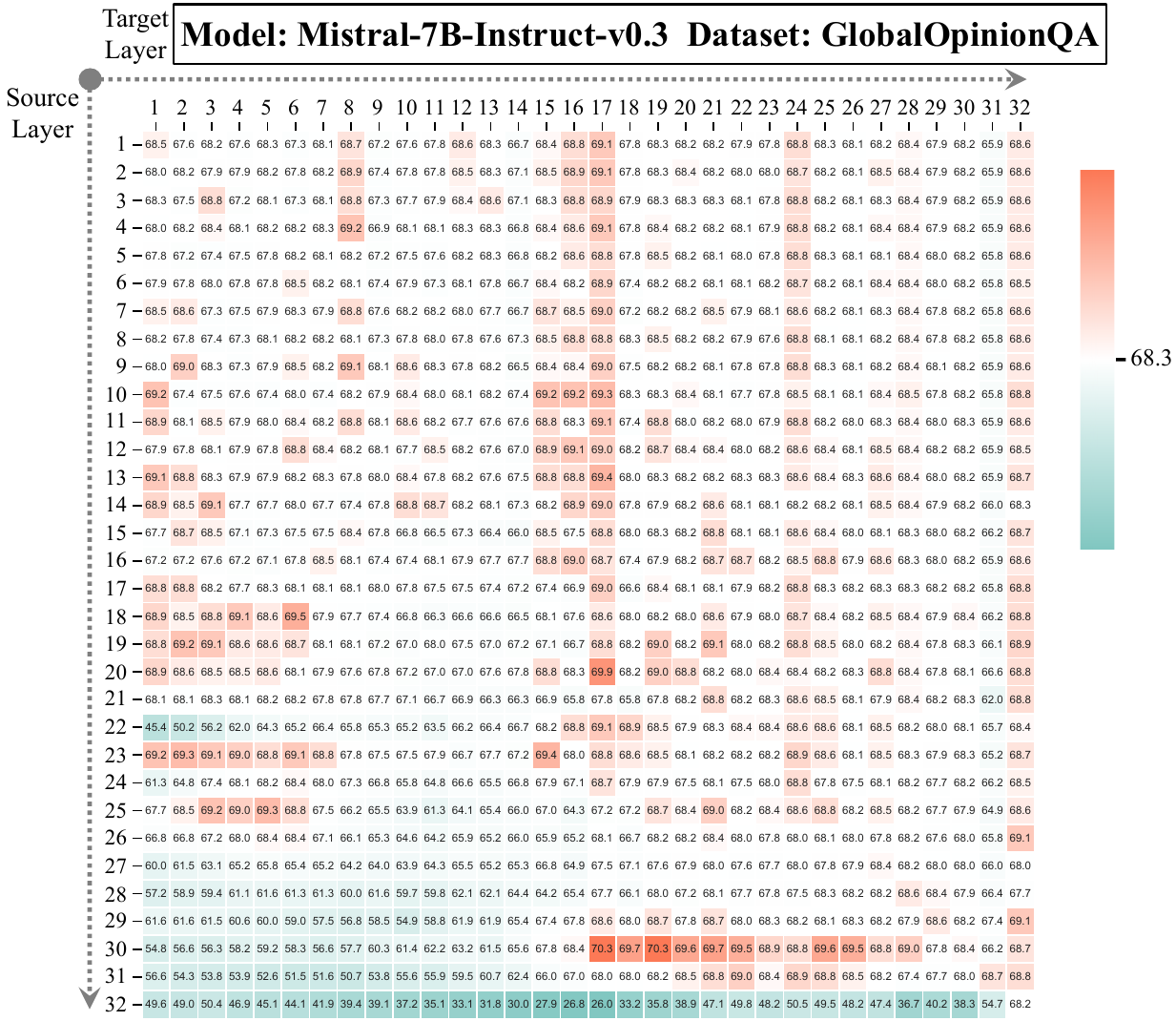}
  \caption{Supplementary layer-specific effectiveness results of \textit{Mistral-7B-Instruct-v0.3} on GlobalOpinionQA dataset. The colorbar median value represents the accuracy of backbone model.
  }
  \label{fig:global_mistral}
\end{figure}

\begin{figure}[h]
  \centering
  \includegraphics[width=0.5\textwidth]{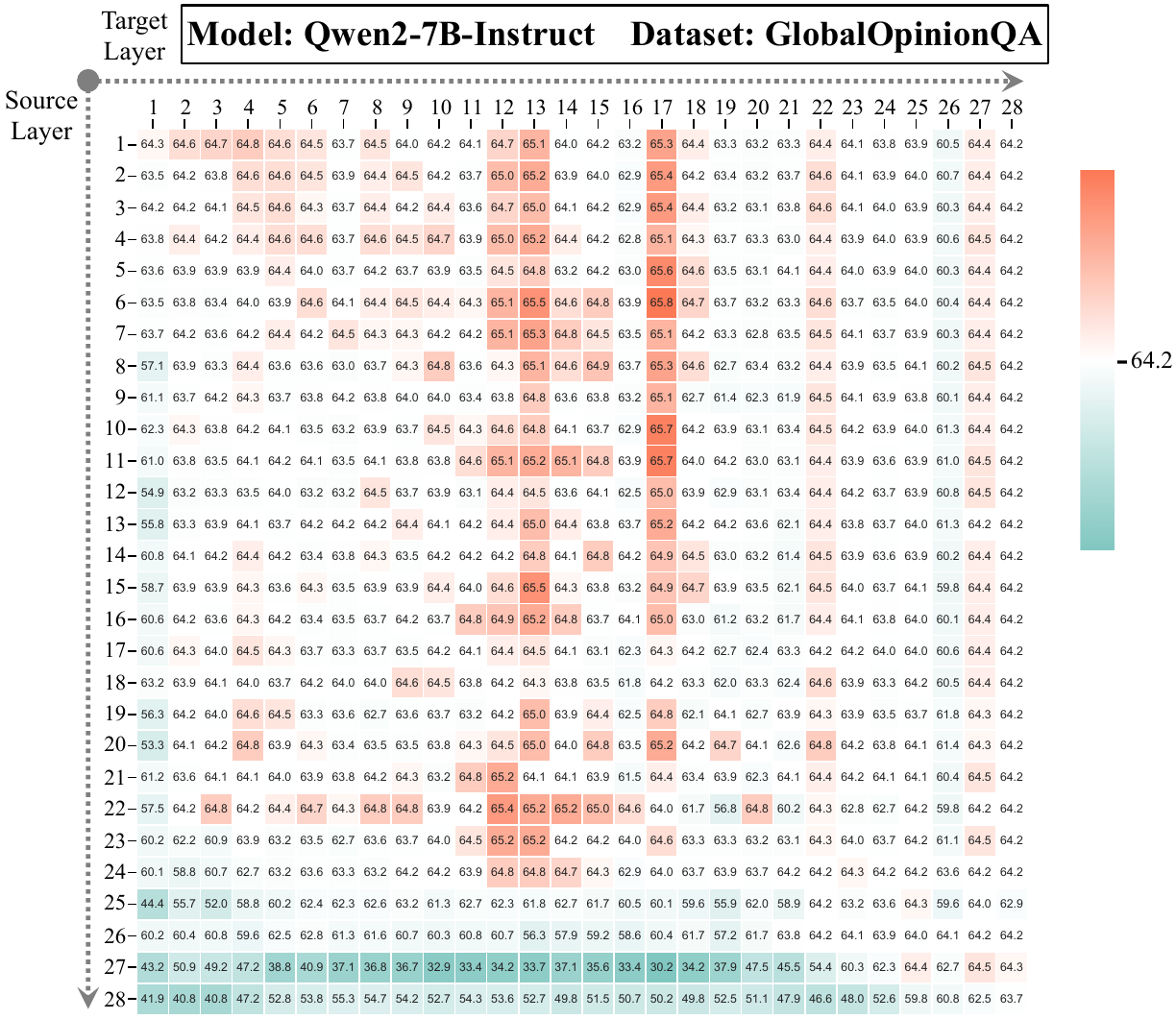}
  \caption{Supplementary layer-specific effectiveness results of \textit{Qwen2-7B-Instruct} on GlobalOpinionQA dataset. The colorbar median value represents the accuracy of backbone model.
  }
  \label{fig:global_qwen}
\end{figure}

To further investigate the impact of multi-step {\name} on performance, we present illustrative case studies from the Chinese Math Reasoning dataset. We select a set of questions from the Chinese Math Reasoning dataset and apply multi-step {\name} to the questions. The results produced by \textit{Llama-2-7b-Chat} are shown in Figure~\ref{fig:cases_multi_step}.
We observe that when the multi-step intervention is applied (the \textcolor{red}{$\bm{\downarrow}$} in the figure denotes the intervention step), the model's semantic trajectory shifts abruptly, often resulting in degenerate or nonsensical outputs.

\begin{table}
  \centering 
  \renewcommand{\arraystretch}{1.1}
  \setlength\tabcolsep{4pt}
  \caption{The Averaged Results of Multi-step {\name} Attempts on Long-form Answering.}
  \resizebox{0.48\textwidth}{!}{%
      \begin{tabular}{lcccccccccccc}
          \toprule
          {\bf {Accuracy on MGSM}} & en & bn & de & es & fr & ja & ru & sw & te & th & zh &\bf Avg.\\
          % \textbf{\makecell{Llama-2-7b-Chat}} & \textbf{\makecell{Mistral-7B-Instruct-v0.3}} & \textbf{\makecell{Qwen2-7B-Instruct}} \\
          \midrule
          \multicolumn{13}{c}{\textbf{\makecell{Llama-2-7b-Chat}}} \\
          \midrule
          Backbone Model & 26.0 & 0.0 & 22.4 & 22.8 & 20.4 & 15.2 & 14.0 & 2.0 & 0.0 & 0.0 & 16.0 & \bf12.6 \\
          \noalign{\vskip 0.2ex}\cdashline{1-13}\noalign{\vskip 0.4ex}
          {\name} \\
          -- (Attn-level, Last-to-First) & 27.2 & 0.0 & 24.4 & 23.6 & 24.0 & 17.6 & 16.4 & 4.0 & 0.0 & 2.0 & 18.4 &\bf 14.3 \\
          -- (FFN-level, First-to-First) & 26.0 & 0.0 & 22.4 & 22.4 & 20.8 & 17.6 & 14.0 & 2.0 & 0.0 & 1.2 & 18.0 &\bf 13.1 \\
          \noalign{\vskip 0.2ex}\cdashline{1-13}\noalign{\vskip 0.4ex}
          {\name} + Multi-step \\
          -- (Attn-level, Last-to-First) & 12.8 & 0.0 & 8.0 & 9.2 & 10.8 & 5.2 & 9.2 & 1.6 & 0.0 & 1.2 & 12.0 &\bf 6.4 \\
          -- (FFN-level, First-to-First) & 20.0 & 0.0 & 14.8 & 16.4 & 17.2 & 9.2 & 11.2 & 1.6 & 0.0 & 1.2 & 18.0 &\bf 10.0 \\
          \midrule
          \multicolumn{13}{c}{\textbf{\makecell{Mistral-7B-Instruct-v0.3}}} \\
          \midrule
          Backbone Model & 52.0 & 10.4 & 41.2 & 43.6 & 5.2 & 25.6 & 38.4 & 10.4 & 2.0 & 18.0 & 37.2 &\bf 25.8 \\
          \noalign{\vskip 0.2ex}\cdashline{1-13}\noalign{\vskip 0.4ex}
          {\name} \\
          -- (Attn-level, Last-to-First) & 50.0 & 14.4 & 41.2 & 41.2 & 7.6 & 24.8 & 40.0 & 10.0 & 2.4 & 18.0 & 36.8 &\bf 26.0 \\
          -- (FFN-level, First-to-First) & 52.0 & 11.2 & 41.2 & 44.0 & 5.6 & 25.2 & 38.0 & 10.8 & 2.4 & 18.0 & 37.6 &\bf 26.0 \\
          \noalign{\vskip 0.2ex}\cdashline{1-13}\noalign{\vskip 0.4ex}
          {\name} + Multi-step \\
          -- (Attn-level, Last-to-First) & 10.4 & 1.2 & 8.0 & 8.4 & 5.6 & 4.0 & 5.2 & 4.4 & 1.2 & 1.6 & 8.0 &\bf 5.3 \\
          -- (FFN-level, First-to-First) & 40.4 & 3.2 & 27.6 & 31.2 & 7.2 & 15.2 & 29.2 & 3.2 & 1.6 & 16.0 & 27.2 &\bf 18.4  \\
          \midrule
          \multicolumn{13}{c}{\textbf{\makecell{Qwen2-7B-Instruct}}} \\
          \midrule
          Backbone Model & 73.6 & 23.6 & 67.2 & 68.8 & 67.6 & 51.6 & 70.8 & 20.0 & 9.2 & 51.2 & 68.8 &\bf 52.0\\
          \noalign{\vskip 0.2ex}\cdashline{1-13}\noalign{\vskip 0.4ex}
          {\name} \\
          -- (Attn-level, Last-to-First) & 76.4 & 37.6 & 68.0 & 69.6 & 69.6 & 53.2 & 64.0 & 18.4 & 12.0 & 54.8 & 60.8 &\bf 53.1 \\
          -- (FFN-level, First-to-First) & 73.6 & 23.2 & 67.6 & 69.6 & 66.8 & 51.2 & 72.0 & 19.6 & 9.2 & 50.8 & 70.4 &\bf 52.2\\
          \noalign{\vskip 0.2ex}\cdashline{1-13}\noalign{\vskip 0.4ex}
          {\name} + Multi-step \\
          -- (Attn-level, Last-to-First) & 22.0 & 14.0 & 30.0 & 27.2 & 23.6 & 24.0 & 23.2 & 5.6 & 5.2 & 23.6 & 24.4 &\bf 20.3 \\
          -- (Attn-level, Last-to-First) & 37.6 & 11.6 & 42.8 & 47.6 & 41.2 & 29.2 & 44.0 & 5.6 & 6.0 & 29.6 & 49.2 &\bf 31.3 \\
          \bottomrule
      \end{tabular}}
      \label{tab:multi_step_app}
\end{table}

\vfill

\end{document}